\documentclass[letterpaper, 10 pt, journal, twoside]{IEEEtran}
\usepackage{graphicx}
\usepackage{verbatim}
\usepackage{amsfonts}
\usepackage{xcolor,soul,framed} 
\usepackage{lineno}
\usepackage{longtable}
\usepackage{caption}
\usepackage{rotating}  
\colorlet{shadecolor}{yellow}
\usepackage{booktabs} 
\usepackage{multirow}
\graphicspath{{../pdf/}{../jpeg/}}
\DeclareGraphicsExtensions{.pdf,.jpeg,.png}

\usepackage[cmex10]{amsmath}
\usepackage{array}
\usepackage{mdwmath}
\usepackage{mdwtab}
\usepackage{eqparbox}
\usepackage{url}

\usepackage{caption}
\usepackage{subcaption}
\usepackage[export]{adjustbox}
\usepackage{authblk}
\usepackage{gensymb}
\IEEEoverridecommandlockouts
\usepackage{titlesec}

\titlespacing*{\section}{0pt}{1.7em}{0.5em} 
\titlespacing*{\subsection}{0pt}{1.7em}{0.5em} 

\AtBeginDocument{
  \setlength{\intextsep}{-1pt plus 1pt minus 1pt}

  \setlength{\textfloatsep}{-1pt plus 1pt minus 1pt}

  \setlength{\floatsep}{-1pt plus 1pt minus 1pt}
}

\begin{document}
\title{TransForSeg: A Multitask Stereo ViT for Joint Stereo Segmentation and 3D Force Estimation in Catheterization}
 
\author{Pedram Fekri$^{1}$,
            Mehrdad Zadeh$^{2}$,
            Javad Dargahi$^{1}$
\thanks{}
\thanks{
} 
\thanks{$^{1}$Pedram Fekri and Javad Dargahi are with the Mechanical, Industrial and Aerospace Engineering Department, Concordia University, Montréal, Quebec, Canada
        {\tt\footnotesize      p$\_$fekri@encs.concordia.ca, dargahi@encs.concordia.ca}}%

\thanks{$^{2} $Mehrdad Zadeh is with the Electrical and Computer Engineering Department, Kettering University, Flint, Michigan, USA
        {\tt\footnotesize mzadeh@kettering.edu}}%
\thanks{This work has not been peer-reviewed. It is a preprint submitted to arXiv for early dissemination.\\
\copyright \ 2025 by the authors. This work is licensed under a non-exclusive license to arXiv.
}
}

\markboth{}
{Fekri \MakeLowercase{\textit{et al.}}: TransForceg: A Multi-modal Encoder-Decoder Architecture for Stereo Segmentation and 3D Force Estimation}

\maketitle

\begin{abstract}
Recently, the emergence of multitask deep learning models has enhanced catheterization procedures by providing tactile and visual perception data through an end-to-end architecture. This information is derived from a segmentation and force estimation head, which localizes the catheter in X-ray images and estimates the applied pressure based on its deflection within the image. These stereo vision architectures incorporate a CNN-based encoder-decoder that captures the dependencies between X-ray images from two viewpoints, enabling simultaneous 3D force estimation and stereo segmentation of the catheter. With these tasks in mind, this work approaches the problem from a new perspective. We propose a novel encoder-decoder Vision Transformer model that processes two input X-ray images as separate sequences. Given sequences of X-ray patches from two perspectives, the transformer captures long-range dependencies without the need to gradually expand the receptive field for either image. The embeddings generated by both the encoder and decoder are fed into two shared segmentation heads, while a regression head employs the fused information from the decoder for 3D force estimation. The proposed model is a stereo Vision Transformer capable of simultaneously segmenting the catheter from two angles while estimating the generated forces at its tip in 3D. This model has undergone extensive experiments on synthetic X-ray images with various noise levels and has been compared against state-of-the-art pure segmentation models, vision-based catheter force estimation methods, and a multitask catheter segmentation and force estimation approach. It outperforms existing models, setting a new state-of-the-art in both catheter segmentation and force estimation.  
\end{abstract}


%
\IEEEpeerreviewmaketitle


\section{Introduction}
A catheter is a flexible, intravascular tube used in cardiac catheterization to access and navigate the cardiovascular system with precision. Surgeons or robotic platforms employ this technique for diagnostic angiography and therapeutic interventions, including hemodynamic assessment, intravascular biopsy, percutaneous coronary intervention (PCI), and targeted drug delivery. To ensure accurate placement, the catheter is introduced through the femoral, jugular, or subclavian access points and guided in real time using X-ray fluoroscopic imaging \cite{complic:30285356, complic3}. 
\par
Surgeons, semi-autonomous, or autonomous robotic platforms require both visual and tactile sensory feedback to safely perform the procedure. However, the majority of these interventions are conducted using standard catheters without integrated force sensors or micro-cameras at the tip, primarily due to cost constraints. In practice, a surgeon relies on haptic perception to avoid applying excessive pressure on the vessel lumen, reducing the risk of complications. Simultaneously, visual feedback, such as catheter localization and segmentation, is essential for precise navigation through the vascular pathway \cite{roshanfar2023autonomous, autonomous, fekri1}.

To address the challenges associated with tactile and visual information, learning-based deep learning models have been introduced to extract these features directly from X-ray images. These data-driven approaches mitigate the need for physical force sensors by inferring contact forces and catheter positioning from imaging data. Deep learning-based methods for contact force estimation can be broadly classified into 2D and 3D force estimators, while visual information is typically obtained through semantic segmentation models, enabling precise catheter localization and tracking \cite{y-net, segmed}. A more recent class of methods integrates both catheter segmentation and force estimation into a unified framework, leveraging a multimodal end-to-end network to extract both visual and tactile information directly from X-ray images. This model simultaneously localizes the catheter and infers contact forces, eliminating the need for additional hardware sensors or two-stage processing with separate networks for each task, thereby improving efficiency and real-time applicability \cite{fekri_h-net}. 
\par
The majority of existing catheter force estimation models rely on CNN-based (Convolutional Neural Network) encoders to capture the relationship between catheter deflections observed from two different angles. These stereo encoders transform spatial pixel information into a latent feature space by progressively expanding the receptive field through image downsampling. Once semantically rich embeddings are extracted, the feature vectors corresponding to the two input views are fused and fed into an MLP regression head to estimate contact forces along the $x$, $y$, and $z$ axes. Catheter segmentation models on the other hand, are predominantly monocular, employing either CNN-based encoder-decoder structures or ViT-based encoder (Vision Transformer) architectures for pixel-wise classification \cite{vit}. To date, the only existing multimodal architecture that performs both stereo catheter segmentation and force estimation simultaneously is based on a CNN-based encoder-decoder framework \cite{fekri_h-net}. Moreover, there is currently no ViT-based architecture specifically designed for stereo segmentation for this application; the related use of ViTs in stereo vision involves depth estimation, where a ViT module is employed to estimate disparity from paired images \cite{instr-stereo, vit-stereo-depth}. Consequently, the potential of a multimodal, patch-based model leveraging stereo ViT capable of simultaneously segmenting the catheter from two viewpoints and estimating contact forces at the catheter tip remains unexplored.
\par
In this work, we propose a novel multitask encoder-decoder ViT architecture featuring two segmentation heads and a regression head, enabling the simultaneous segmentation of the catheter from two viewpoints and the estimation of the applied contact force at the catheter tip. The model incorporates a transformer encoder and decoder, each receiving patch sequences from two X-ray images concurrently as input. The patches are projected into semantically rich embeddings capturing the global context of the images from both the encoder and decoder. The extracted embeddings from both views are progressively upsampled through a CNN-based decoder to generate the segmentation maps. Given that the contact forces along the $x$, $y$, and $z$ axes can be inferred from the catheter's deflection across two viewpoints, a cross self-attention mechanism is applied atop the ViT decoder to capture the dependencies between the encoder and decoder embeddings. Finally, the [CLS] token from the decoder output, post self-attention, is passed to an MLP regression head to predict the 3D contact force. To ensure an efficient and lightweight architecture, the ViT decoder shares its weights with the ViT encoder, effectively mirroring its graph. Additionally, the CNN-based upsampling head used to reconstruct the segmentation maps is shared between the encoder and decoder, reducing model complexity and overall parameter count. The main \textbf{contributions} of this work are as follows:
\begin{itemize}
    \item We propose a novel multitask, patch-based encoder-decoder architecture built on Vision Transformers (ViT) for simultaneous stereo catheter segmentation and 3D contact force estimation from X-ray images in catheterization procedures.
    \item The model can estimate contact forces directly from X-ray images, as the segmentation task guides the network to focus on the catheter's deflection shape rather than background variations.
    \item The ViT encoder and decoder share weights to enhance computational efficiency, while a cross-attention mechanism fuses tokens from the X-ray images at both angles to enable accurate 3D contact force prediction.
    \item The proposed ViT-based stereo vision model processes two input X-ray images and produces three outputs across two modalities: two segmentation maps and a force vector predicting contact forces along the $x$, $y$, and $z$ axes.
    \item The embedding upsampler module, referred to as the segmentation head, is shared between the ViT encoder and decoder to reduce redundancy and enhance efficiency. 
    \item The proposed architecture has undergone extensive evaluation and ablation studies, achieving a new state-of-the-art in sensor-free, learning-based 3D catheter force estimation and segmentation.  
\end{itemize}
As the proposed model is based on a Transformer architecture for both catheter segmentation and force estimation, we refer to it as TransForSeg for brevity. 
\section{Related Work}
\textbf{Vision-based force estimator}: These types of solutions estimate the contact forces applied at the catheter tip directly from X-ray images \cite{khodashenas2021, roshanfar2023deeplearning}. Specifically, they project the input images into a latent feature space, where the catheter’s deflections are mapped under supervised learning into a vector representing the applied forces. In \cite{fekri2021deep}, a ResNet-based encoder is used as a feature extractor to map single-view catheter images into a latent feature space, followed by an MLP regression head that estimates 2D forces. In \cite{y-net}, a Y-shaped architecture is proposed to process a pair of clean images simultaneously, capturing the catheter from two different angles. This stereo processing enhances the model’s ability to predict forces not only along the $x$ and $y$ axes but also along the $z$ axis.
Although both models successfully project images into the force space, they rely on a separate segmentation model during the data preprocessing phase to extract the catheter shape and generate clean input images. This additional step increases computational cost during training, as it requires running an extra model in parallel. Moreover, it affects the overall pipeline efficiency at test time.    
\begin{figure*}[t]
	\centering
	\includegraphics[trim={0mm 0mm 0 0},clip, width=0.92\textwidth]{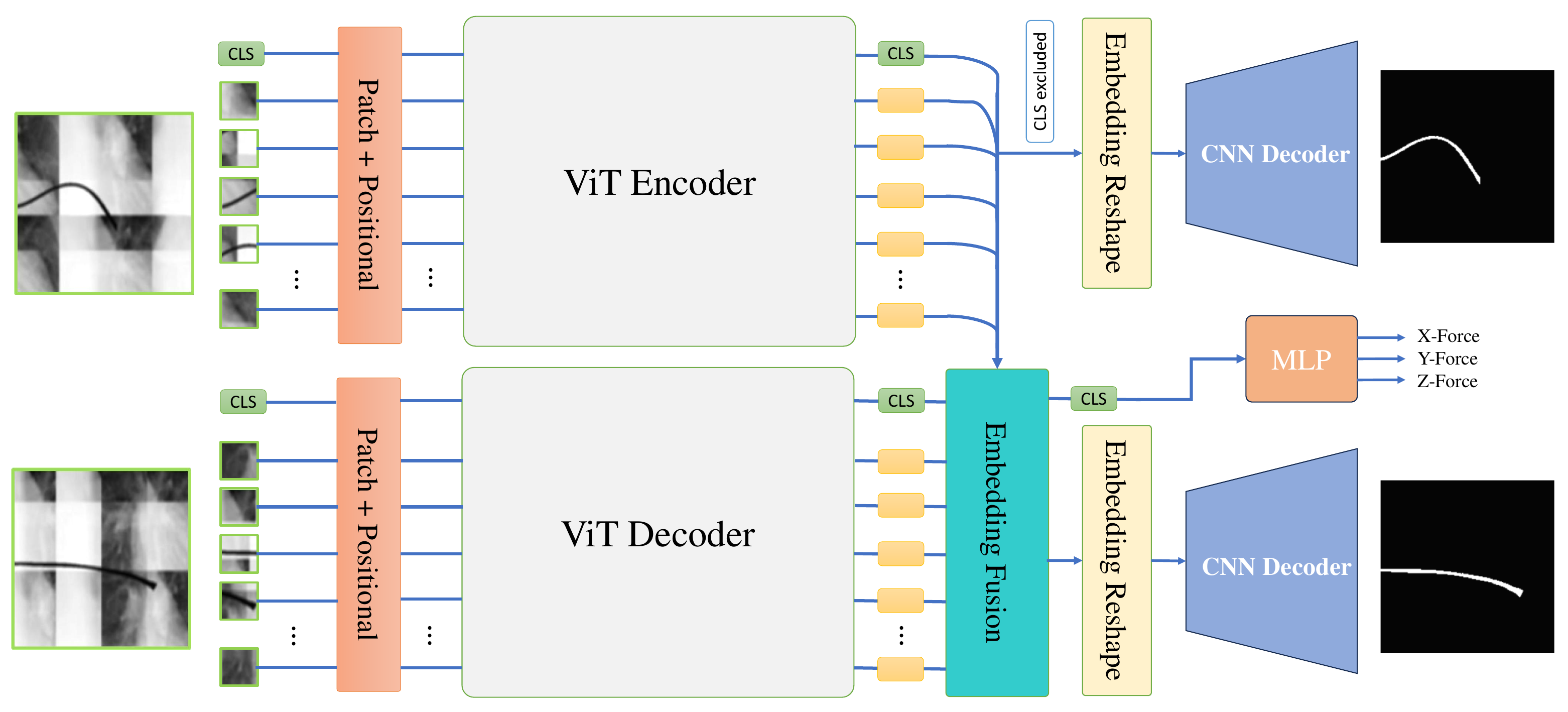}
	\caption{The diagram shows TransForSeg's architecture. It is equipped with a Vision Transformer (ViT)-based encoder and decoder that share weights. The model receives two X-ray images; one for the encoder and one for the decoder. Each image is divided into patches, a [CLS] token is added, and the resulting sequence passes through a positional embedding layer. Both the encoder and decoder consist of $T$ stacked Transformer blocks, with the final block generating the output tokens. The decoder’s output token fuses with the encoder’s tokens through an embedding fusion layer. The [CLS] token from the decoder is then passed to an MLP head for force estimation. The token embeddings from both the encoder and decoder (excluding the [CLS] token) are reshaped and upsampled by a shared CNN-based decoder for segmentation.}
	\label{fig:net}
\end{figure*}
\par
\noindent \textbf{Catheter segmentation}: Localizing the catheter in X-ray images is essential for enabling surgeons to accurately navigate it through the vascular system. This task also serves as the perception module in semi-autonomous catheterization robotic platforms \cite{aut2, aut3, aut4, aut5}. The problem is commonly addressed using semantic segmentation techniques, with CNN-based encoder-decoder architectures being the most widely adopted in the literature. In these models, the encoder compresses the input image and extracts semantically rich feature maps, while the decoder progressively upsamples these features to reconstruct the spatial resolution and generate the final segmentation map \cite{unet, unet-2024, unet-2024_cat, fcn, FCN-cat, hrnet, hrnet_cat, deeplab}. On the other hand, transformer-based semantic segmentation models have recently gained traction, inspired by the success of similar architectures in Natural Language Processing (NLP) \cite{attentionallyouneed}. These models, often referred to as patch-based architectures, receive image patches as input token sequences and process them in parallel. The ViT encoder captures long-range dependencies across different regions of the image through self-attention mechanisms. The output embeddings from a Vision Transformer can take the form of patch embeddings or mask/object queries, depending on the model design. These embeddings are then upsampled; typically using a CNN-based decoder or interpolation techniques; to reconstruct the spatial resolution and generate the final feature maps for segmentation \cite{setr, segmenter, segment-anything, maskformer}. While the original ViT-based segmentation models process images at a single patch scale without inherent multi-scale feature extraction, recent architectures have been proposed that extract multi-scale embeddings to better capture features at different spatial resolutions \cite{segformer, swin}.
\par \noindent \textbf{Multitask architecture}: As previously discussed, catheter segmentation serves dual purposes: as a preprocessing module for vision-based force estimation and as a perception module for catheter navigation. In \cite{fekri_h-net, fs-net}, a CNN-based encoder-decoder architecture is proposed to address both tasks within a single end-to-end framework. Specifically, the model receives a pair of X-ray images as input and processes them through a CNN-based encoder. The resulting bottleneck feature maps are then passed to both an MLP regression head for 3D force estimation and a convolutional decoder for segmentation map generation. Our work is inspired by this multitask approach and shares the same objectives. However, we tackle the problem using a novel encoder-decoder design based on Vision Transformers (ViT), offering improved performance and representation capabilities through global attention mechanisms. 

\section{Methodology}
TransForSeg is a stereo vision model that processes two X-ray images of a catheter captured from different perspectives. By fusing visual information from both views, the model enhances its ability to capture 3D catheter deflections, leading to more accurate estimation of contact forces along the $x$, $y$, and $z$ axes. Furthermore, TransForSeg functions as a multitask transformer, simultaneously segmenting the catheter and producing two segmentation maps corresponding to the input views, alongside a 3D force estimation output. This end-to-end, multi-input multi-output network is built upon an encoder-decoder ViT architecture. Fig. 1 shows the architecture of TransForSeg. 
\par \noindent \textbf{Patch Embeddings:} During a single forward pass, the top-view X-ray image is divided into non-overlapping patches of size $(n \times n)$. This patch projection is implemented using a 2D convolution with $k$ kernels of size $(n \times n)$ and a stride of $n$, consistent with the original Vision Transformer (ViT) design \cite{vit}. Since TransForSeg does not utilize overlapping patches, the stride matches the patch size. The projection yields a sequence of $p \times p$ patches, where each $n\times n$ patch is mapped to an embedding of dimension $k$. Unlike some segmentation and detection model, a learnable  [CLS] token, with the same dimensionality as the patch embeddings, is then appended to the sequence, resulting in a total of $ + 1$ tokens. This [CLS] token is later used by the force estimation head, as will be explained in the following section. Next, positional embeddings of shape $(p^2 + 1, k)$ are added to inject spatial information into the sequence. The resulting input, a sequence of $p^2 + 1$ embeddings of dimension $k$, is then passed to the transformer encoder. 
\par
\noindent \textbf{ViT Encoder and Decoder:} The encoder consists of a stack of $T$ Transformer blocks. Each block begins with element-wise layer normalization applied to the input embeddings, followed by a Multi-Head Self-Attention (MHSA) mechanism with $H$ attention heads. A residual connection is added between the input tokens and the MHSA output. The result is then passed through a second layer normalization, followed by a Feed-Forward Network (FFN), typically implemented as a two-layer MLP. A second residual connection is applied between the MHSA output and the FFN output, preserving gradient flow and improving stability during training. A more detailed technical explanation will be provided in the following section.
\par Unlike ViT-based object detection and segmentation models, the decoder in TransForSeg is fed with the side-view X-ray image. In models such as OWL-ViT, DETR, and SAM, the decoder typically receives object queries as input \cite{detr, owl-vit, segment-anything}. In contrast, the decoder in the proposed architecture directly processes patched X-ray images, similar to the encoder. The decoder shares the same architectural structure as the encoder, and for computational efficiency, the weights are shared between them. The encoder output consists of embeddings corresponding to each input patch, which are then split into two branches. One branch is directed to the segmentation upsampler, while the other is passed to the decoder. The encoder generates embeddings denoted as $x_t \in \mathbb{R}^{B, p^2 + 1, k}$, where $B$ is the batch size, $p^2 + 1$ represents the number of tokens (including the [CLS] token), and $k$ is the embedding dimension. Similarly, the decoder produces embeddings from the side-view X-ray patches, denoted as $x_s$, which have the same dimensionality as $x_t$. In TransForSeg, the decoder attends to the encoder’s output embeddings using Multi-Head Cross-Attention (MHCA) to capture the long-range dependencies between the two X-ray views. In essence, this mechanism enables the model to focus on corresponding regions across both images where variations contribute to differences in the resulting contact force values along the $x$, $y$, and $z$ axes. To be more specific, firstly, the input $x$  (either in the encoder or decoder) is normalized by a layer norm:
\begin{equation}
\label{7}
\hat{x} = \frac{x - \mu}{\sqrt{\sigma^2 + \epsilon}}
\end{equation}

\begin{equation}
\label{7}
norm(x) = \gamma \cdot \hat{x} + \beta
\end{equation}

\noindent where \( \sigma^2 \) and \( \mu \) are standard deviation and mean across the embedding dimension. \( \epsilon \) is a small constant (e.g., \( \epsilon = 10^{-6} \)) for numerical stability and (2) is the affine transformation with learnable parameters \( \gamma \) and \( \beta \). The attention inputs form the key ($K$), query ($Q$) and value ($V$), while $H$ denotes the number of attention heads. Each each attends to a part of the embedding dimension. This is done through a projection of $K$, $V$, and $Q$ for each head (h) as follows:

\begin{equation}
\label{7}
\begin{aligned}
Q_h &= x \cdot W_q^{h} + b_q^{h} \\
V_h &= x \cdot W_v^{h} + b_v^{h} \\
K_h &= x \cdot W_k^{h} + b_k^{h}
\end{aligned}
\end{equation}
\noindent where all weight matrices $W \in \mathbb{R}^{k\times k_h}$ and $k_h = k / H$. Given the projected key, query and value, the attention score in each head is calculated as follows:
\begin{equation}
\text{A}_h = softmax(\frac{Q_h \cdot K_h^T}{\sqrt{D_h}})\cdot V_h
\end{equation}

\noindent Next, the attention ($A_h$) from all heads are concatenated to form the full attention $A$ to go through the final projection followed by a residual connection and layer norm:

\begin{equation}
\text{A} = norm(x + (A \cdot W_o + b_k))
\end{equation}

\noindent  The embedding $A$ feeds a two-layer MLP in which the first layer expands the embedding dimension by factor $F$ while the second layer reduces it back to the original dimension. Finally, a residual connection is applied between the input embedding and the FFN's output followed by a layer norm. This structure is generalizable to all transformer blocks and TransForSeg uses it as MHSA heads in the encoder and decoder where $Q, K, V = x_s$ and $Q, K, V = x_t$ while it employs it as MHCA where where $Q = emb_s$ and $k,V = emb_t$. 
\par \noindent \textbf{Force Estimation Head:} The force estimation head receives the [CLS] token from the decoder, which encodes the mutual information between the X-ray images from both views. The remaining embeddings are discarded, as the regression head does not require visual information. For the sake of a fair comparison in benchmarking, the force estimation head adopts the same structure as in H-Net; a 3-layer MLP with 64, 32, and 3 units in each layer, respectively. The loss for this head is the MSE as follows:
\begin{equation}
\mathcal{L}_{\text{force}} = \frac{1}{FD} \| \hat{y} - y \|_2^2
\end{equation}
where $FD$ denotes the force dimension equals 3 (i.e., along $x$, $y$ and $z$). 
\par \noindent \textbf{Segmentation head:} Both the ViT encoder and decoder output embeddings $emb \in R^{B, {p^2}+1,k}$, representing the patch tokens and a [CLS] token for each input image. In the segmentation upsampling subnetwork, the [CLS] tokens are discarded, and the remaining patch embeddings are reshaped into feature maps $f(t,s) \in R^{p\times p \times k}$, where $t$ and $s$ denote the top and side views, respectively. Several methods have been proposed in the literature for upsampling embedding feature maps, including bilinear interpolation, progressive upsampling using deconvolution, and direct patch-size mapping \cite{setr, segformer, maskformer}. In TransForSeg, we adopt a deconvolution-based decoder, similar to the one used in H-Net, to ensure consistency in benchmarking \cite{fekri_h-net}. The stacked embeddings are upsampled to the original input image size through four stages. First, the embeddings pass through two successive 2D convolutional layers, which reduce the feature dimensionality by a factor of four while preserving the height and width of the feature maps. Next, four upsampling blocks, in contrast to H-Net, each consisting of a deconvolution layer followed by a convolution layer, progressively increase the spatial resolution by a factor of 2 and reduce the number of channels. Finally, a sigmoid activation function is applied to generate a single-channel segmentation map. The segmentation upsampler is shared between both the ViT encoder and decoder and is trained using the Binary Cross Entropy (BCE) loss, denoted as$\mathcal{L}_{\text{seg}}$. 
\par TransForSeg is an end-to-end multitask model, optimized by minimizing the following total loss function:

\begin{equation}
\mathcal{L}_{\text{total}} = \lambda_1\mathcal{L}_{\text{force}} + \lambda_2\mathcal{L}_{\text{seg}}^{\text{enc}} + \lambda_3\mathcal{L}_{\text{seg}}^{\text{dec}}
\end{equation}

\noindent where the coefficients \( \lambda_1, \lambda_2, \lambda_3 \) weight each component of the loss: mean squared error (MSE) for force estimation, and binary cross entropy (BCE) for the segmentation outputs from the encoder and decoder, respectively.

\section{Experimental Results}

Our proposed model is evaluated using three previously compiled datasets. These datasets are based on the Y-Net paper, which introduced an experimental setup designed to replicate a biplanar fluoroscopy system in operation \cite{y-net}. In this setup, a standard catheter was pressed against a force sensor while two cameras captured its deflection from different angles. This setup produced a total of 19,500 samples, each consisting of two RGB images (from different viewpoints) and a corresponding 3D force vector ($x$, $y$, $z$) measured by the force sensor.
However, the captured RGB images were not representative of real-world X-ray imagery. To address this, the H-Net work proposed a synthetic X-ray generator that transformed RGB images into synthetic X-ray images with two levels of background complexity \cite{fekri_h-net}. Additionally, it produced two segmentation maps per sample. As a result, each sample in the extended dataset includes: two synthetic X-ray images (XRay1 and XRay2), the original RGB images, a 3D force vector, and two corresponding segmentation maps.
In our experiments, we used the same datasets and preserved the original data split for a fair benchmark comparison: 80\% of the data (13,650 samples) was used for training. The remaining 20\% was split equally into validation and test sets, with 2,925 samples in each (please see Fig. 4).     

\subsection{Implementation Details}
The input to TransForSeg consists of two images of a catheter deflection captured simultaneously from two different angles. For the synthetic X-ray dataset, the images are denoted as $(x_s, x_t) \in R^{224\times224\times1}$, while for the RGB dataset, they are  $(x_s, x_t) \in R^{224\times224\times3}$. TransForSeg is implemented in two sizes: ViT-Tiny and ViT-Small. In the Tiny version, both the encoder and decoder use the ViT-Tiny architecture. The input images are split into patches using a 2D convolution with 192 kernels of size $16 \times 16$, resulting in 197 tokens (including the [CLS] token), each with a dimensionality of 192. The transformer consists of 12 blocks, each with 3 attention heads, and the FFN expands the hidden dimension to 768. In the Small version, each patch is projected into an embedding of size 384. Each transformer block contains 6 attention heads, and the FFN expands the dimensionality to 1536. 
\par
The embedding fusion module is implemented as a transformer block with 8 cross-attention heads and a feed-forward network (FFN) expansion size of 2048. The [CLS] token is excluded from both the encoder and decoder embeddings to form the fused feature map $f(t,s) \in R^{14 \times 14 \times n}$ where $n\in \{192, 384\}$ which serves as the input to the segmentation head. This shared segmentation head first reduces the feature map dimensionality to 96 using two successive convolutional layers with kernel size $3\times3$, applied in two stages. Then, it progressively upsamples the feature maps to the original image resolution of $224 \times 224$ using 4 deconvolution blocks in order to generate the segmentation maps. Each block contains a 2D convolutional layer with  $3\times3$ kernels that reduces the channel dimensionality by a factor of 2, followed by a transposed convolution layer with $4\times4$ kernel and stride 2, which doubles the spatial dimensions (height and width).
\par 
The force estimation head is implemented as a two-layer MLP, which receives the [CLS] token from the ViT decoder as input. Depending on the model variant, the input embedding has a dimensionality of either 192 (ViT-Tiny) or 384 (ViT-Small). This input is projected to intermediate dimensions of 64 and 32, respectively. The output layer predicts the force vector as a three-dimensional vector. Both ViT variants are pretrained on ImageNet. TransForSeg, in both configurations, is trained using the RGB, X-Ray1, and X-Ray2 datasets with a fixed learning rate of $1 \times {10}^{-4}$ (without any scheduling),   over 250 epochs, and with a batch size of 32 using Adam optimizer. The entire network is kept unfrozen during training. The loss weight coefficients (7) were empirically set to 1, assigning equal contribution to each loss term. For each dataset and model variant, the training set is randomly shuffled, and training is repeated three times. The best checkpoint from each run is selected based on the lowest MSE on the unshuffled validation set. MSE is used as the checkpointing criterion due to the slower convergence of the force estimation head compared to the segmentation heads. All reported results in the following section represent the average performance of the three best checkpoints for each model-dataset combination, evaluated on the corresponding test set. All experiments are conducted on an NVIDIA RTX 2080 GPU running Ubuntu 20.04.
\begin{table}[b]
\centering
\caption{Performance of Force Estimation Across Datasets}
\label{tab:performance}
\begin{tabular}{l|l|l|c|c|c|c}
\toprule
\textbf{Data} & \textbf{type} & \textbf{Model} & \textbf{MSE} & \textbf{MAE} & \textbf{RMSE} & \textbf{$R^2$} \\ 
\midrule
\multirow{4}{*}{Seg} & NN           & MLP \cite{khodashenas2021}
& 4.4e-05 
& 0.0040 
& - 
& 0.98 
\\ 
                           & \multirow{2}{*}{CNN} 
                           & ResNet \cite{fekri2021deep}
                           & - 
                           & - 
                           & 0.110 
                           & - \\
                           &                  & Y-Net \cite{y-net}
                           & 2.8e-05
                           & 0.0039 
                           & 0.005 
                           & 0.98 \\ 
\midrule
\multirow{3}{*}{RGB} 
                           & \multirow{1}{*}
                           {CNN} 
                           & H-Net \cite{fekri_h-net}
                           & 3.6e-05 
                           & 0.0039
                           & 0.005
                           & 0.98 \\
                           & \multirow{2}{*}{ViT}    & Ours (t) 
                           & \textbf{1.77e-05} 
                           & \textbf{0.0027} 
                           & \textbf{0.0042} 
                           & \textbf{0.99} \\ 
                           &
                           & Ours (s) 
                           & 1.81e-05 
                           & 0.0028 
                           & 0.0042 
                           & 0.99 \\ 
                           
\midrule
\multirow{4}{*}{Xray1} 
                           & \multirow{1}{*}{CNN} & H-Net \cite{fekri_h-net}
                           & 3.3e-05 
                           & 0.0039 
                           & 0.0049 
                           & 0.98 \\
                           
                           & \multirow{2}{*} {ViT} &
                           Ours (t) 
                           & \textbf{2.04e-05} 
                           & \textbf{0.0029} 
                           & \textbf{0.0045} 
                           & 0.98 \\ 
                           &
                           & Ours (s) 
                           & 2.79e-05 
                           & 0.0034 
                           & 0.0053 
                           & 0.98 \\
\midrule
\multirow{4}{*}{Xray2} 
                           & \multirow{1}{*}{CNN} 
                           & H-Net \cite{fekri_h-net}
                           & 4.3e-05 
                           & 0.0046 
                           & 0.0058 
                           & 0.97 \\
                           & \multirow{2}{*}{ViT}          
                           & Ours (t) 
                           & 3.43e-05 
                           & 0.0038 
                           & 0.0059 
                           & 0.98 \\ 
                           &
                           & Ours (s) 
                           & \textbf{3.34e-05} 
                           & \textbf{0.0038} 
                           & \textbf{0.0058} 
                           & 0.98 \\ 
\bottomrule
\end{tabular}
\end{table}

\begin{table*}[ht]
\centering
\caption{Performance of Catheter Segmentation Across Datasets}
\label{tab:performance}
\begin{tabular}{l|l|c|c|ccc|ccc|ccc}
\toprule
\textbf{Type} & \textbf{Model} & \textbf{\#Params} & \textbf{GFLOPs} & \multicolumn{3}{c|}{\textbf{RGB}} & \multicolumn{3}{c|}{\textbf{X-Ray-1}} & \multicolumn{3}{c}{\textbf{X-Ray-2}} \\ 
\cmidrule(lr){5-7} \cmidrule(lr){8-10} \cmidrule(lr){11-13}
& & & & \textbf{Acc} & \textbf{mIoU} & \textbf{mDice} & \textbf{Acc} & \textbf{mIoU} & \textbf{mDice} & \textbf{Acc} & \textbf{mIoU} & \textbf{mDice} \\ 
\midrule
CNN 
& FCN \cite{fcn} 
& 134
& 42.4 
& 99.2 
& 94.0 
& 96.8 
& 99.2 
& 94.1 
& 96.8 
& 99.1 
& 93.8 
& 96.7 \\ 
   
& U-Net \cite{unet}
& 34
& 77.5 
& 99.8 
& 95.7 
& 98.0 
& 99.7 
& 95.7 
& 98.0 
& 99.7 
& 95.4 
& 97.8 \\ 
    
& Hr-Net \cite{hrnet} 
& 9.6
& 3.5
& 98.8 
& 96.1 
& 98.4 
& 98.8 
& 96.1 
& 98.5 
& 98.8 
& 95.8 
& 98.0 \\ 
    
& H-Net \cite{fekri_h-net}
& 0.46
& 11.4 
& 98.8 
& 95.7 
& 98.0 
& 98.8 
& 95.7 
& 98.0 
& 98.8 
& 95.5 
& 97.8 \\ 

& DeepLabV3 \cite{deeplab}
& 13.8 
& 11.5 
& 99.9 
& 97.0 
& 98.4
& 99.9 
& 97.2 
& 98.5 
& 99.9 
& 96.9 
& 98.4 \\ 
\midrule
Transformer 

& SegFormer \cite{segformer}
& 3.7
& 1.3 
& 99.9 
& \textbf{98.6}
& \textbf{98.9} 
& 98.8 
& 97.6 
& 98.7 
& 99.9 
& 97.5 
& 98.7 \\ 
            
& MaskFormer \cite{maskformer}
& 41.2
& 106.6
& 99.9 
& 96.7 
& 98.3 
& 99.9 
& 96.7 
& 98.3 
& 98.8 
& 96.9 
& 98.3 \\ 
            
& Ours (Tiny) 
& 6.9 
& 2.8 
& 99.9
& \textbf{98.6}
& \textbf{99.3}
& 99.9 
& \textbf{98.7} 
& \textbf{99.3} 
& 99.9 
& \textbf{98.4} 
& \textbf{99.2} \\ 

& Ours (Base) 
& 25.1 
& 10.2 
& 99.9 
& \textbf{98.6} 
& \textbf{99.3}
& 99.9 
& \textbf{98.6}
& \textbf{99.3}
& 99.9 
& \textbf{98.5} 
& \textbf{99.2} \\
\bottomrule
\end{tabular}
\end{table*}

\subsection{Quantitative Results}
Since TransForSeg is a multitask Transformer, each task is evaluated separately by comparing against state-of-the-art models in the literature. The force estimation head is assessed using Mean Squared Error (MSE), Root Mean Squared Error (RMSE), Mean Absolute Error (MAE), and the coefficient of determination ($R^2$). The segmentation heads are evaluated using Accuracy (Acc), mean Intersection over Union (mIoU), and mean Dice coefficient (mDice). All benchmarked models from the literature, including those used for segmentation and force estimation, were trained, validated, and tested on the three aforementioned datasets: RGB, X-Ray1, and X-Ray2. It is worth noting that all reported results in this paper represent the average of three independent runs to ensure consistency and reliability.  
\par
\noindent \textbf{Force Estimation} Table 1 presents a benchmark comparison where TransForSeg-small (s) and TransForSeg-tiny (t) are evaluated against four learning-based catheter force estimation models from the literature. The experiments were conducted across four datasets: Segmented, RGB, Synthetic X-Ray1, and Synthetic X-Ray2. Each row in the table corresponds to the performance of the models on the test set of the respective dataset. All models were trained and validated on the training and validation splits of the same dataset to ensure a fair comparison.
In the segmented dataset, the models are provided with pre-processed images in which the catheter has been segmented and the background removed. The first two models (MLP and ResNet) estimate the applied force at the catheter tip only along the $x$ and $y$ axes, whereas Y-Net performs 3D force estimation. This dataset is considered the least challenging, as the foreground (i.e., the catheter) is already extracted using a thresholding method. However, this setup assumes the availability of a separate segmentation model as a preprocessor, which may reduce computational efficiency in real-world applications.
\par The second to fourth rows of Table 1 compare our proposed model with the only available multitask model in the literature; H-Net, which performs both segmentation and force estimation. Unlike the models tested on the segmented dataset, these comparisons are conducted on unsegmented inputs: RGB, Synthetic X-Ray1, and Synthetic X-Ray2, listed in order of increasing complexity for both segmentation and force estimation. H-Net is a CNN-based encoder-decoder architecture, whereas TransForSeg is built on a Vision Transformer (ViT) backbone. Both models estimate 3D contact forces. As shown in the results, TransForSeg consistently outperforms H-Net across all datasets, achieving MSE improvements of $51\%$ on RGB, $38\%$ on X-Ray1, and $22\%$ on X-Ray2. Even when compared to Y-Net, which is evaluated on the segmented dataset, TransForSeg achieves superior performance on the RGB and X-Ray1 datasets, with improvements of $36\%$ and $27\%$ in MSE, respectively. Regarding the performance of the different TransForSeg variants, the tiny version demonstrated better results on the easier datasets, while the small version slightly outperformed it on the more challenging X-Ray2 dataset. This may be attributed to the higher parameter count in the small model, which allows it to better handle complex scenarios but also makes it more prone to overfitting on simpler datasets. Overall, TransForSeg sets a new state-of-the-art in 3D catheter force estimation, significantly pushing the boundaries of performance in this domain.
\par \noindent \textbf{Segmentation:} Table II presents a comparison between TransForSeg and seven state-of-the-art segmentation models from the literature. Each model in the benchmark was independently trained on the three datasets (RGB, X-Ray1, and X-Ray2), and the results reflect performance on the test set of each respective dataset. In addition to segmentation, the table includes the number of parameters and FLOPs to highlight each model's computational complexity.
All models included in the comparison are single-task, except for H-Net and TransForSeg, which are multitask architectures featuring two segmentation heads and one regression head. For these two models, the reported segmentation scores are averaged across both heads. Unlike general-purpose multi-class segmentation, catheter segmentation is a relatively simple task, and most models handle it effectively. However, based on mIoU and mDice across all datasets, the FCN model demonstrated the weakest performance among both CNN-based and Transformer-based methods. In contrast, DeepLabV3 showed competitive performance among Transformer-based models and generally outperformed other CNN-based models.
\begin{table}[b]
\centering
\caption{Ablation: Model with and without segmentation }
\label{tab:single_column_performance}
\begin{tabular}{l|l|c|c|c|c}
\toprule
\textbf{Dataset} & \textbf{Model} & \textbf{MSE} & \textbf{MAE} & \textbf{RMSE} & \textbf{$R^2$} \\ 
\midrule
\multirow{4}{*}{RGB}    & Tiny (with)     & 1.77e-05 & 0.0027 & 0.0042 & 0.99 \\
                        & Tiny (without)  & \textbf{7.75e-06} & \textbf{0.0018} & \textbf{0.0028} & 0.99 \\
                        & Base (with)     & 1.81e-05 & 0.0028 & 0.0042 & 0.99 \\
                        & Base (without)  & \textbf{7.09e-06} & \textbf{0.0017} & \textbf{0.0027} & 0.99 \\
\midrule
\multirow{4}{*}{X-Ray1} & Tiny (with)     & \textbf{2.04e-05} & \textbf{0.0029} & \textbf{0.0045} & 0.98 \\
                        & Tiny (without)  & 2.86e-05 & 0.0033 & 0.0054 & 0.98 \\
                        & Base (with)     & \textbf{2.79e-05} & \textbf{0.0034} & \textbf{0.0053} & 0.98 \\
                        & Base (without)  & 3.06e-05 & 0.0034 & 0.0055 & 0.98 \\
\midrule
\multirow{4}{*}{X-Ray2} & Tiny (with)     & \textbf{3.43e-05} & \textbf{0.0038} & \textbf{0.0059} & 0.98 \\
                        & Tiny (without)  & 4.08e-05 & 0.0042 & 0.0064 & 0.97 \\
                        & Base (with)     & \textbf{3.34e-05} & \textbf{0.0038} & \textbf{0.0058} & 0.98 \\
                        & Base (without)  & 4.21e-05 & 0.0040 & 0.0065 & 0.97 \\
\bottomrule
\end{tabular}
\end{table}

\begin{figure*}[t]
    \centering

    \begin{subfigure}{.16\textwidth}
        \centering
        \caption{RGB-t (w)} 
        \includegraphics[width=\linewidth]{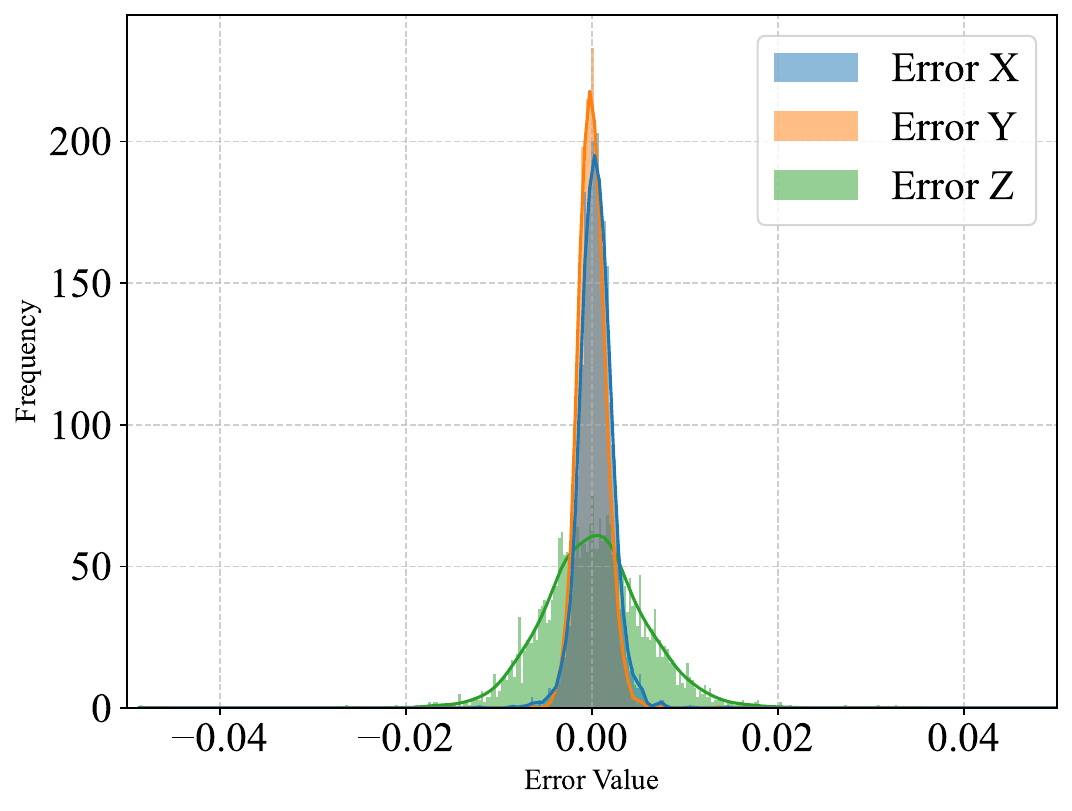}
        \label{fig:sub1}
    \end{subfigure}
    \hspace{-2mm}
    \begin{subfigure}{.16\textwidth}
        \centering
        \caption{XRay1-t (w)} 
        \includegraphics[width=\linewidth]{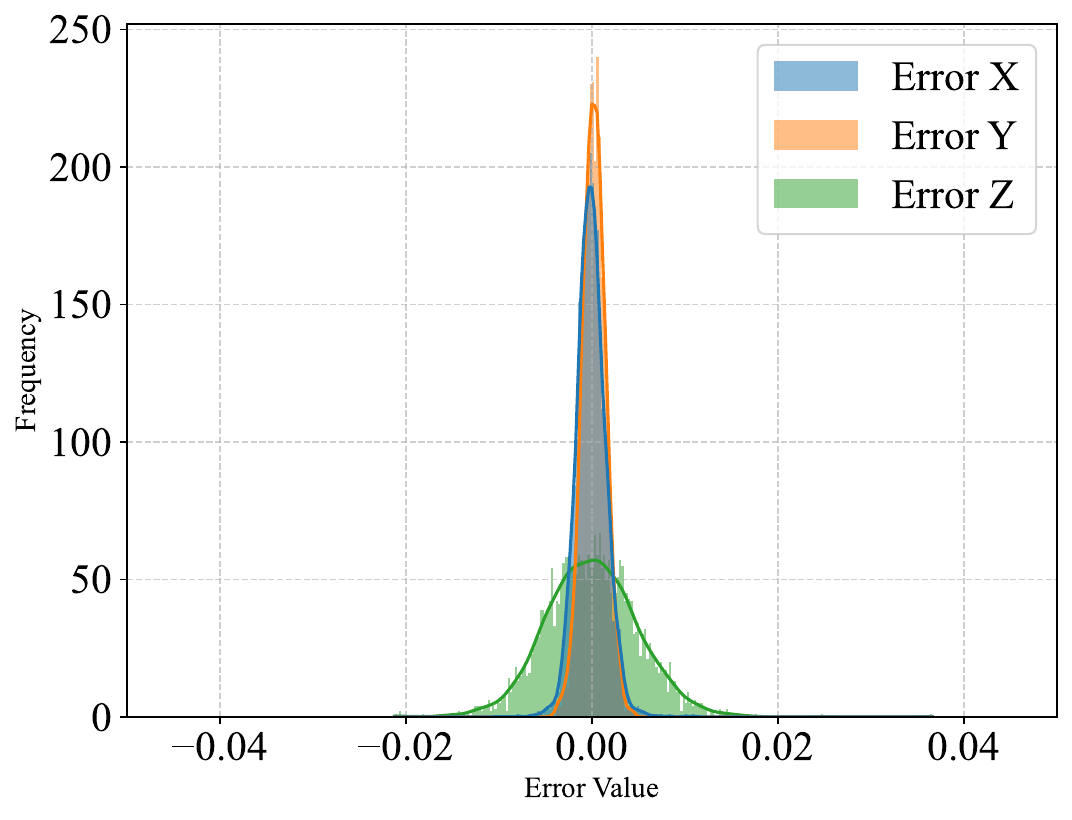}
        \label{fig:sub2}
    \end{subfigure}
    \hspace{-2mm}
    \begin{subfigure}{.16\textwidth}
        \centering
        \caption{XRay2-t (w)} 
        \includegraphics[width=\linewidth]{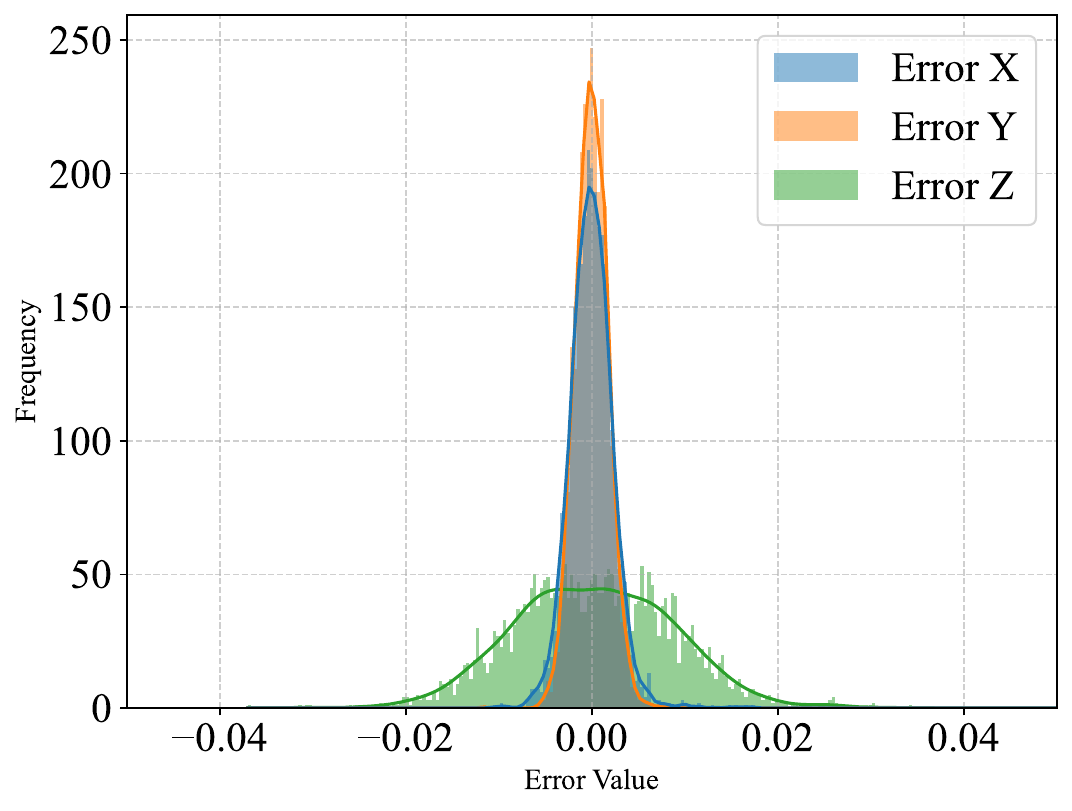}
        \label{fig:sub3}
    \end{subfigure}
    \hspace{-2mm}
    \begin{subfigure}{.16\textwidth}
        \centering
        \caption{RGB-s (w)} 
        \includegraphics[width=\linewidth]{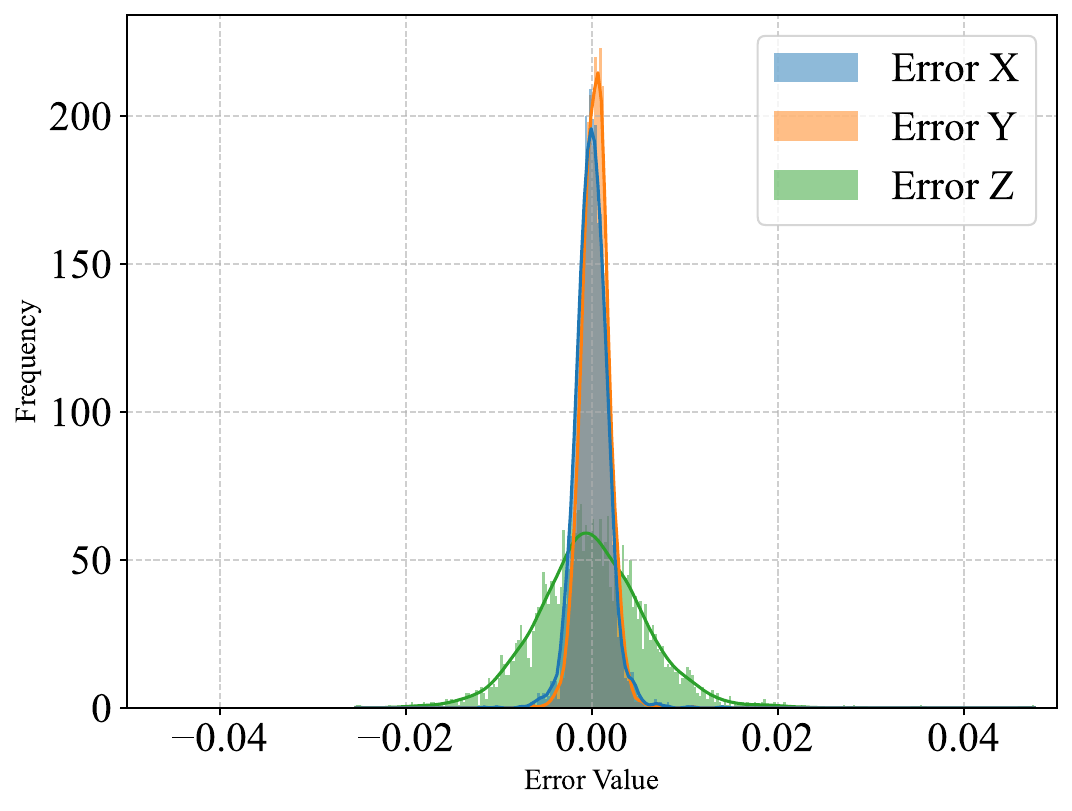}
        \label{fig:sub4}
    \end{subfigure}
    \hspace{-2mm}
    \begin{subfigure}{.16\textwidth}
        \centering
        \caption{XRay1-s (w)} 
        \includegraphics[width=\linewidth]{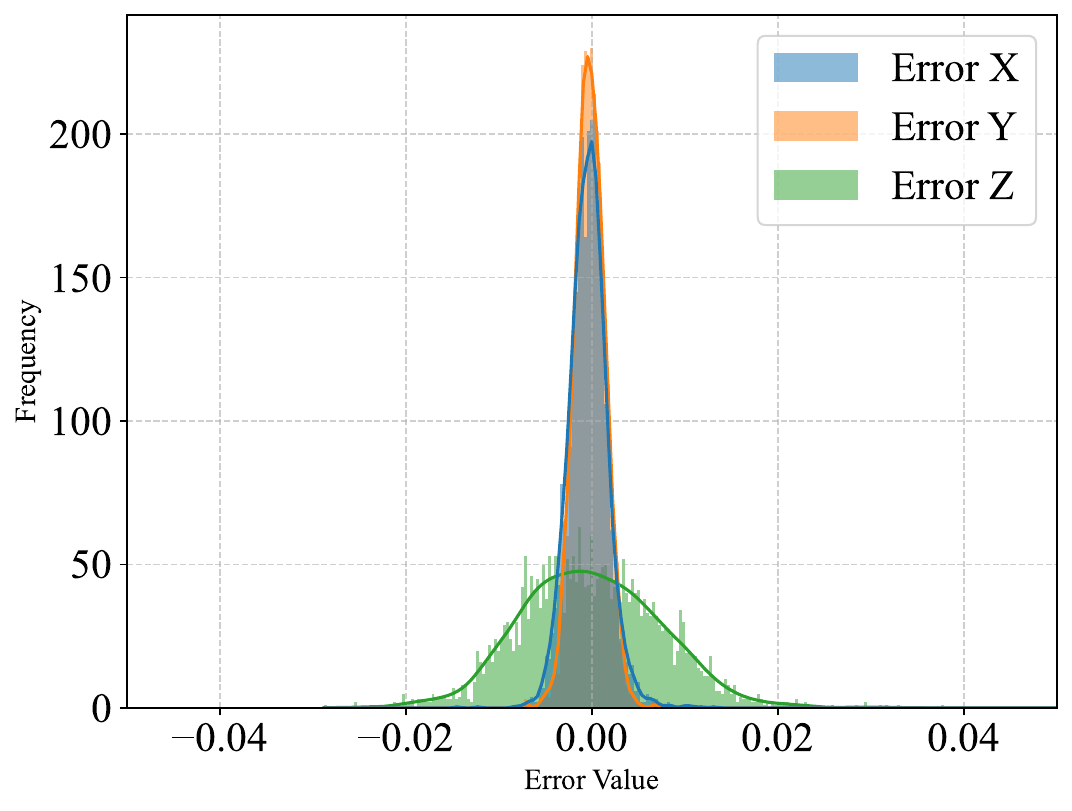}
        \label{fig:sub5}
    \end{subfigure}
    \hspace{-2mm}
    \begin{subfigure}{.16\textwidth}
        \centering
        \caption{XRay2-s (w)} 
        \includegraphics[width=\linewidth]{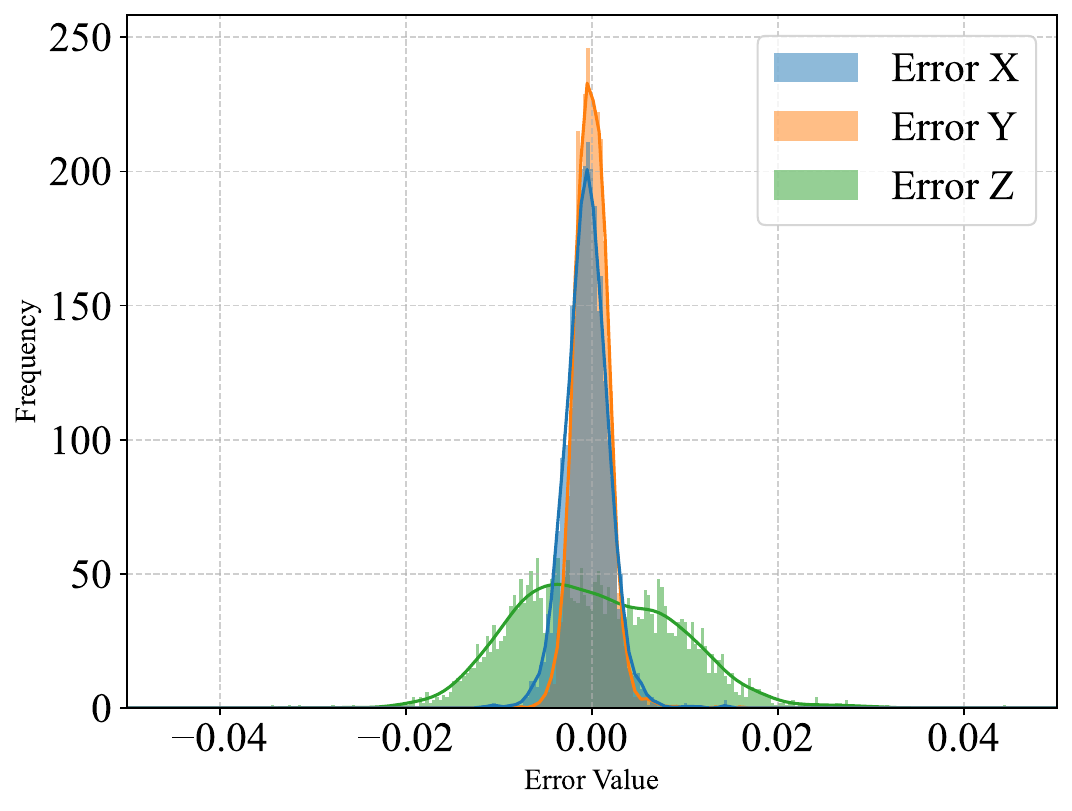}
        \label{fig:sub6}
    \end{subfigure}

    \vspace{-4mm} 

    \begin{subfigure}{.16\textwidth}
        \centering
        \includegraphics[width=\linewidth]{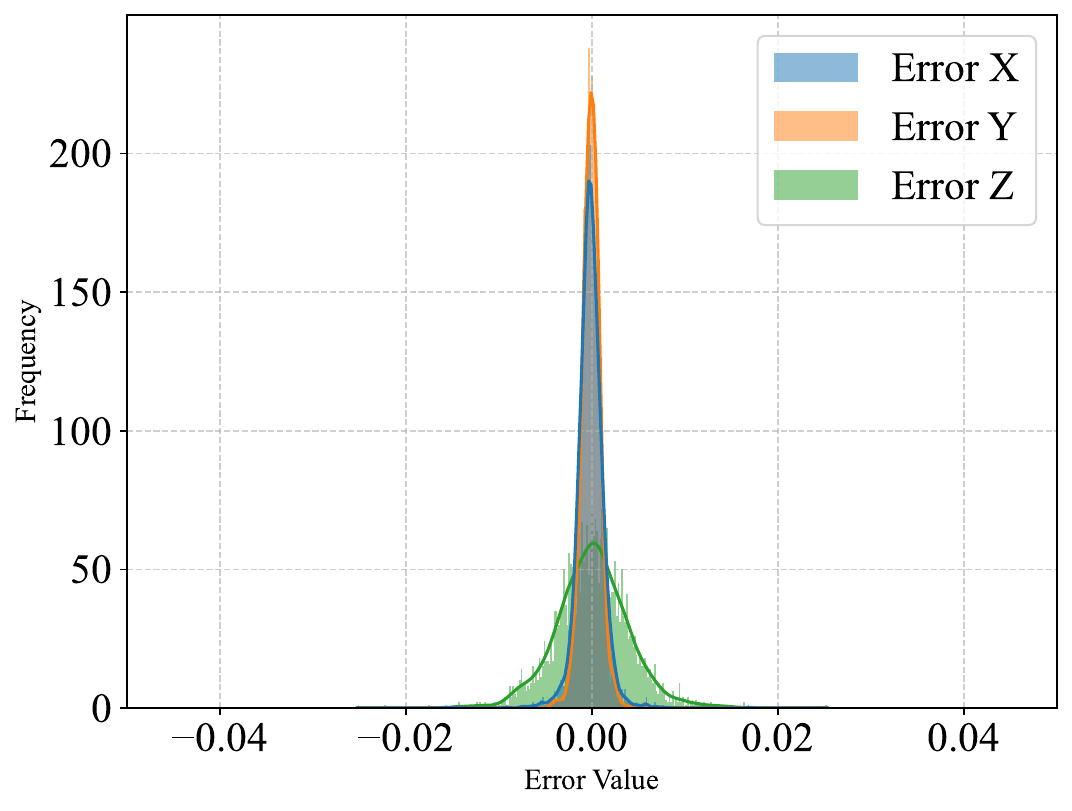}
        \caption{RGB-t (w/o)} 
        \label{fig:sub7}
    \end{subfigure}
    \hspace{-2mm}
    \begin{subfigure}{.16\textwidth}
        \centering
        \includegraphics[width=\linewidth]{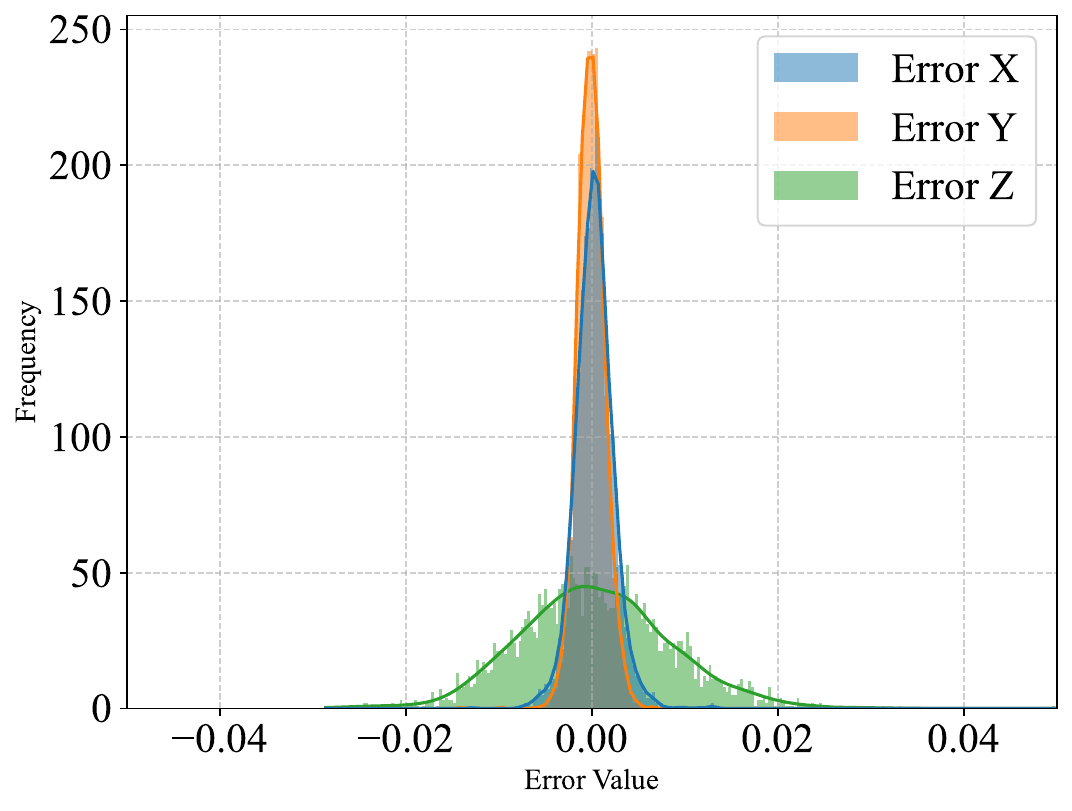}
        \caption{XRay1-t (w/o)} 
        \label{fig:sub8}
    \end{subfigure}
    \hspace{-2mm}
    \begin{subfigure}{.16\textwidth}
        \centering
        \includegraphics[width=\linewidth]{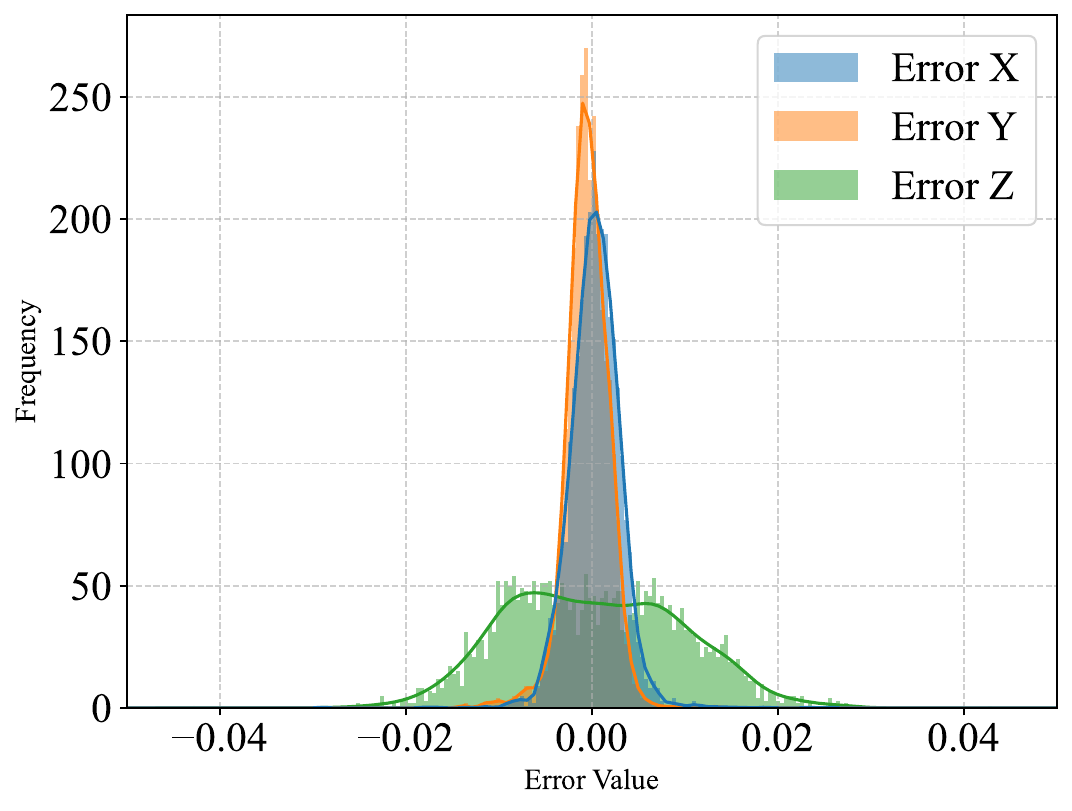}
        \caption{XRay2-t (w/o)} 
        \label{fig:sub9}
    \end{subfigure}
    \hspace{-2mm}
    \begin{subfigure}{.16\textwidth}
        \centering
        \includegraphics[width=\linewidth]{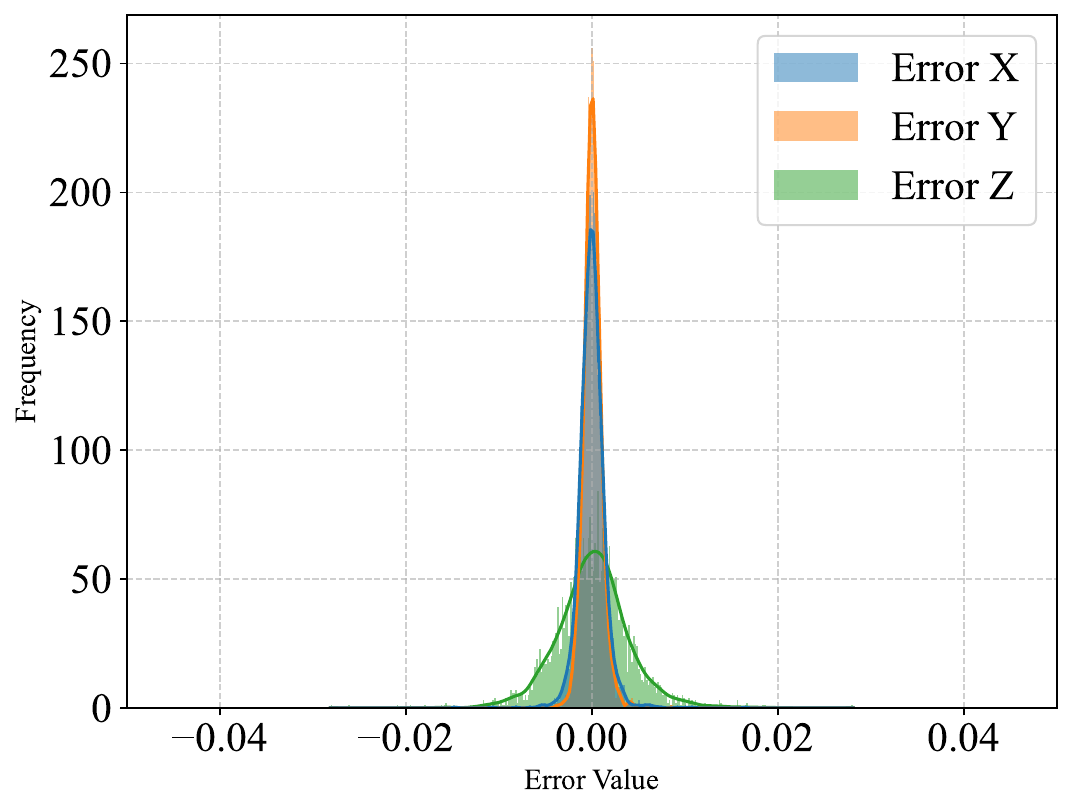}
        \caption{RGB-s (w/o)} 
        \label{fig:sub10}
    \end{subfigure}
    \hspace{-2mm}
    \begin{subfigure}{.16\textwidth}
        \centering
        \includegraphics[width=\linewidth]{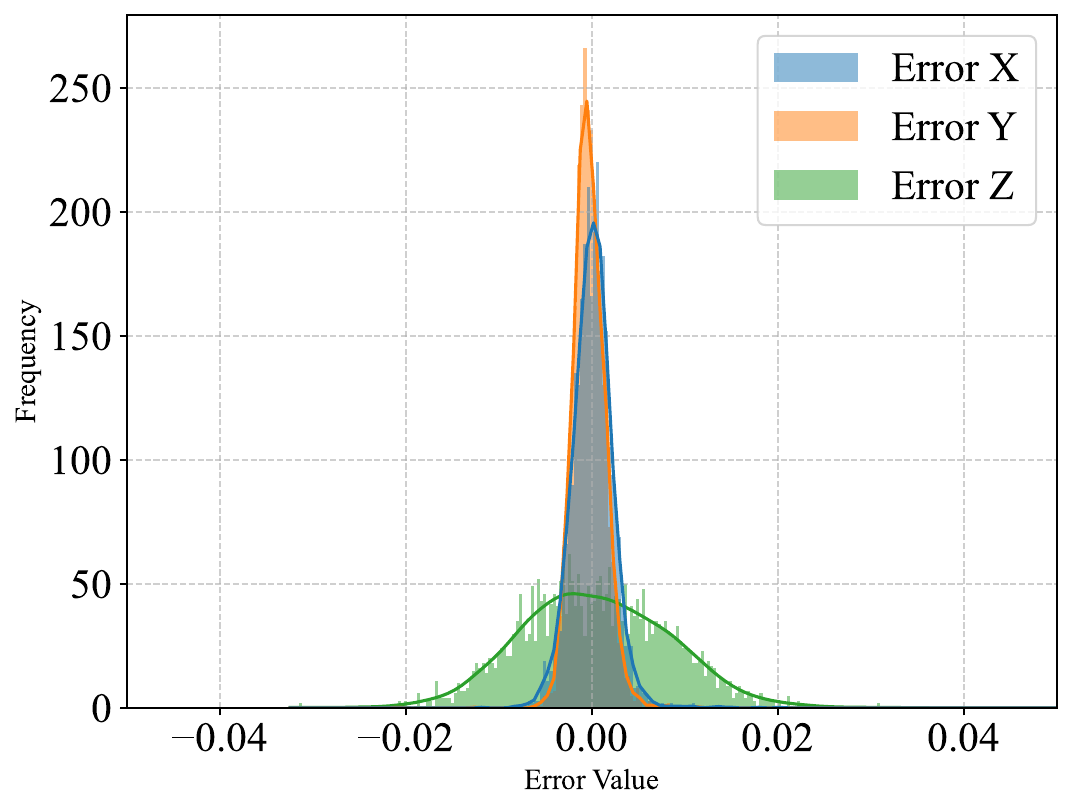}
        \caption{XRay1-s (w/o)} 
        \label{fig:sub11}
    \end{subfigure}
    \hspace{-2mm}
    \begin{subfigure}{.16\textwidth}
        \centering
        \includegraphics[width=\linewidth]{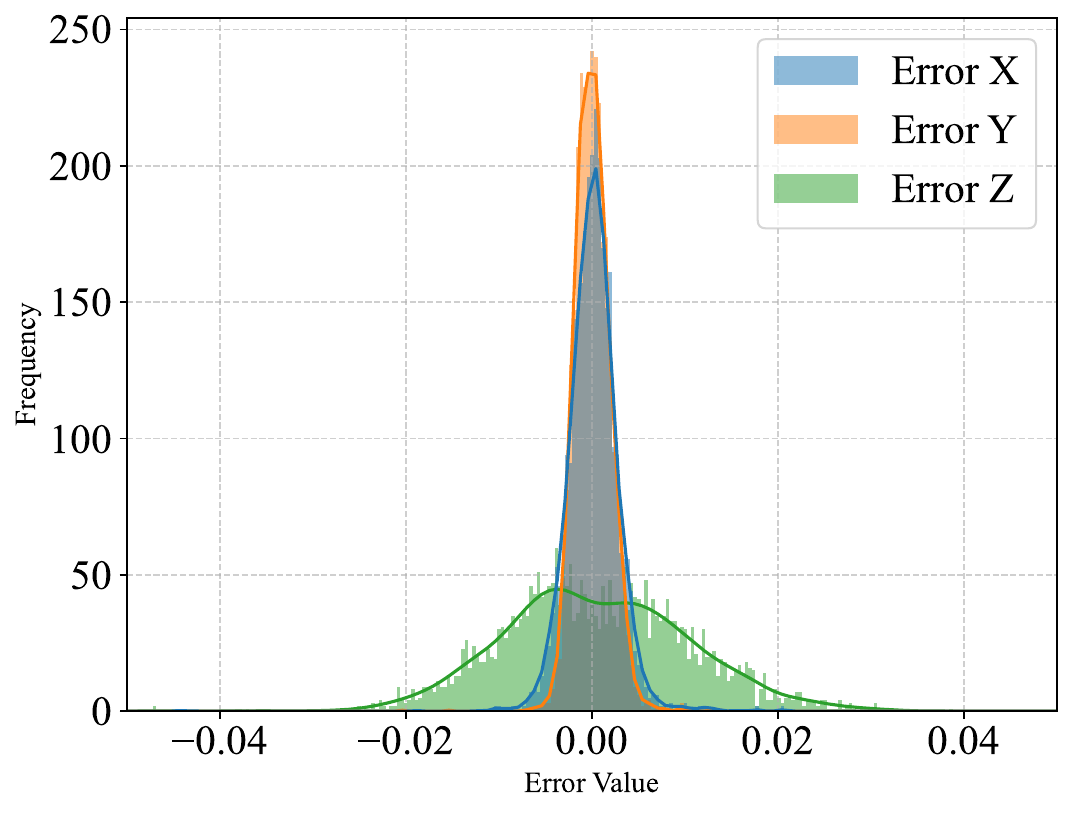}
        \caption{XRay2-s (w/o)} 
        \label{fig:sub12}
    \end{subfigure}

    \caption{The diagram shows the error histograms for the force estimation head of TransForSeg ('w' for models with segmentation heads) and TransForcer ('w/o' for models without segmentation heads) across all datasets. In each subplot, 't' indicates the tiny version and 's' denotes the small version of the model.}
    \label{fig:test}
\end{figure*}

\begin{figure*}[t]
    \centering

    \begin{subfigure}{.24\textwidth}
        \centering
        \renewcommand{\thesubfigure}{a}  
        \caption{ViT-tiny} 
        \includegraphics[width=\linewidth]{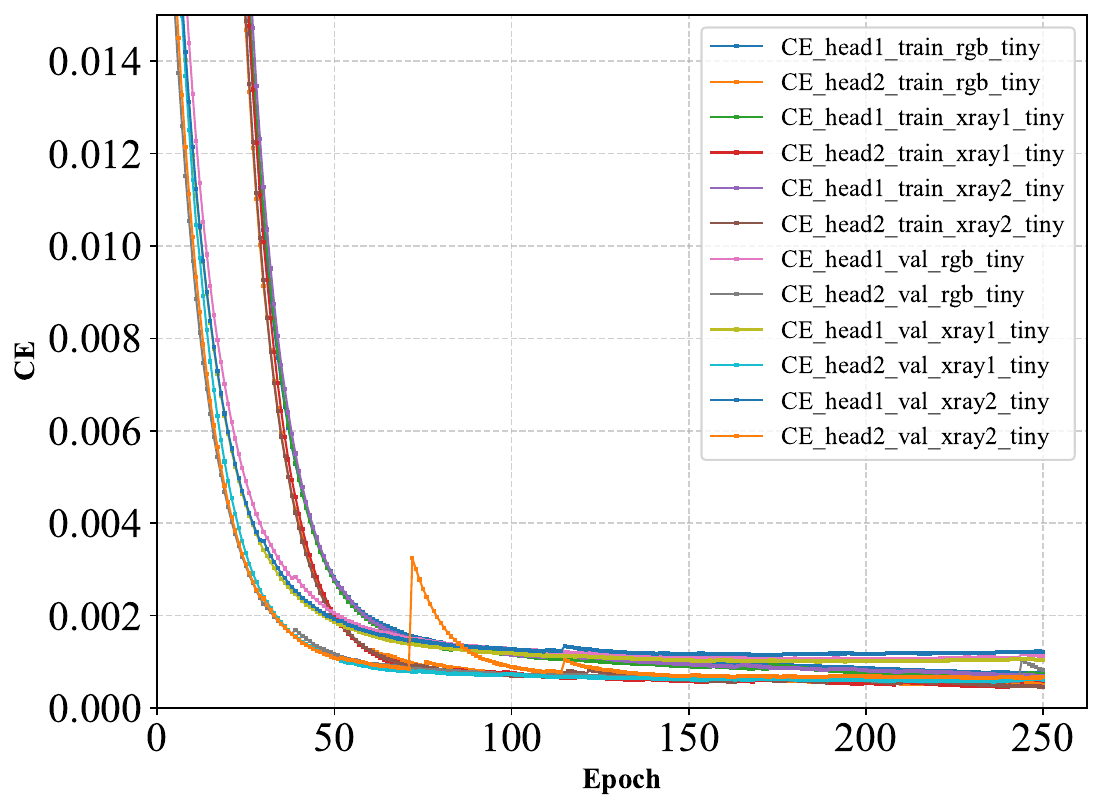}
        \label{fig:sub1}
    \end{subfigure}
    \hspace{-2mm}
    \begin{subfigure}{.24\textwidth}
        \centering
        \renewcommand{\thesubfigure}{b}
        \caption{ViT-tiny} 
        \includegraphics[width=\linewidth]{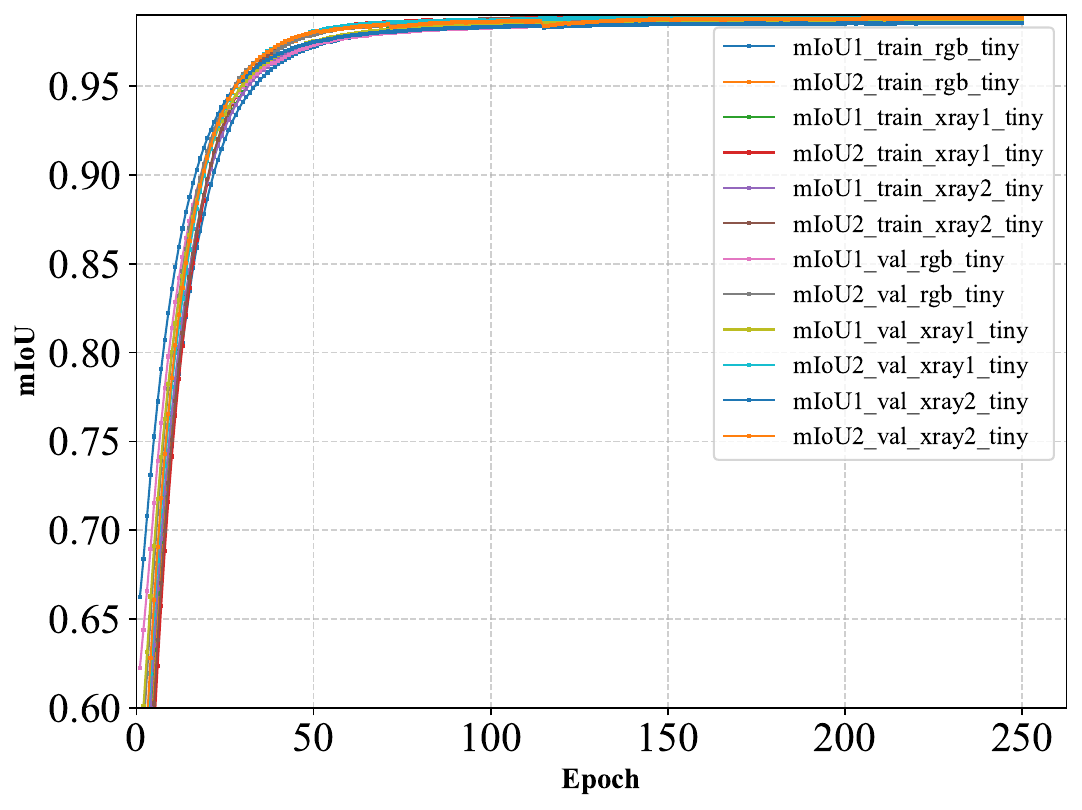}
        \label{fig:sub2}
    \end{subfigure}
    \hspace{-2mm}
    \begin{subfigure}{.24\textwidth}
        \centering
        \renewcommand{\thesubfigure}{e}
        \caption{ViT-tiny (w)} 
        \includegraphics[width=\linewidth]{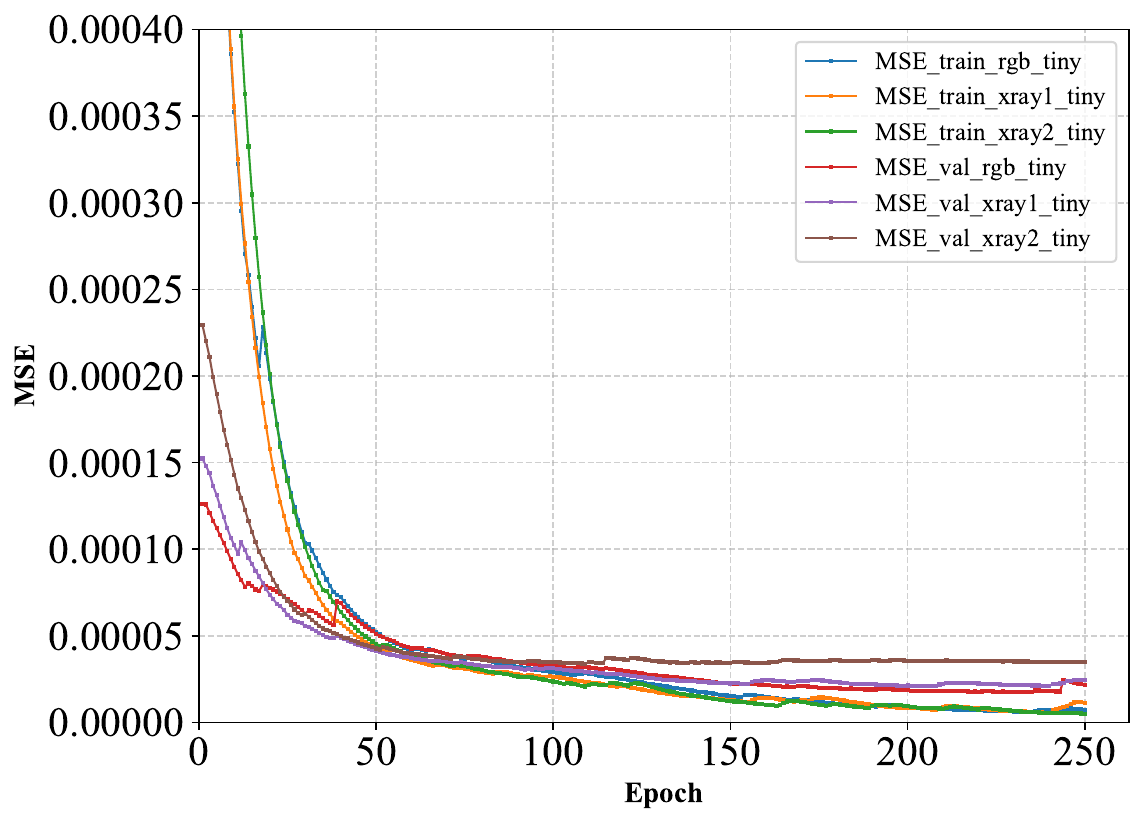}
        \label{fig:sub3}
    \end{subfigure}
    \hspace{-2mm}
    \begin{subfigure}{.24\textwidth}
        \centering
        \renewcommand{\thesubfigure}{f}
        \caption{ViT-small (w)} 
        \includegraphics[width=\linewidth]{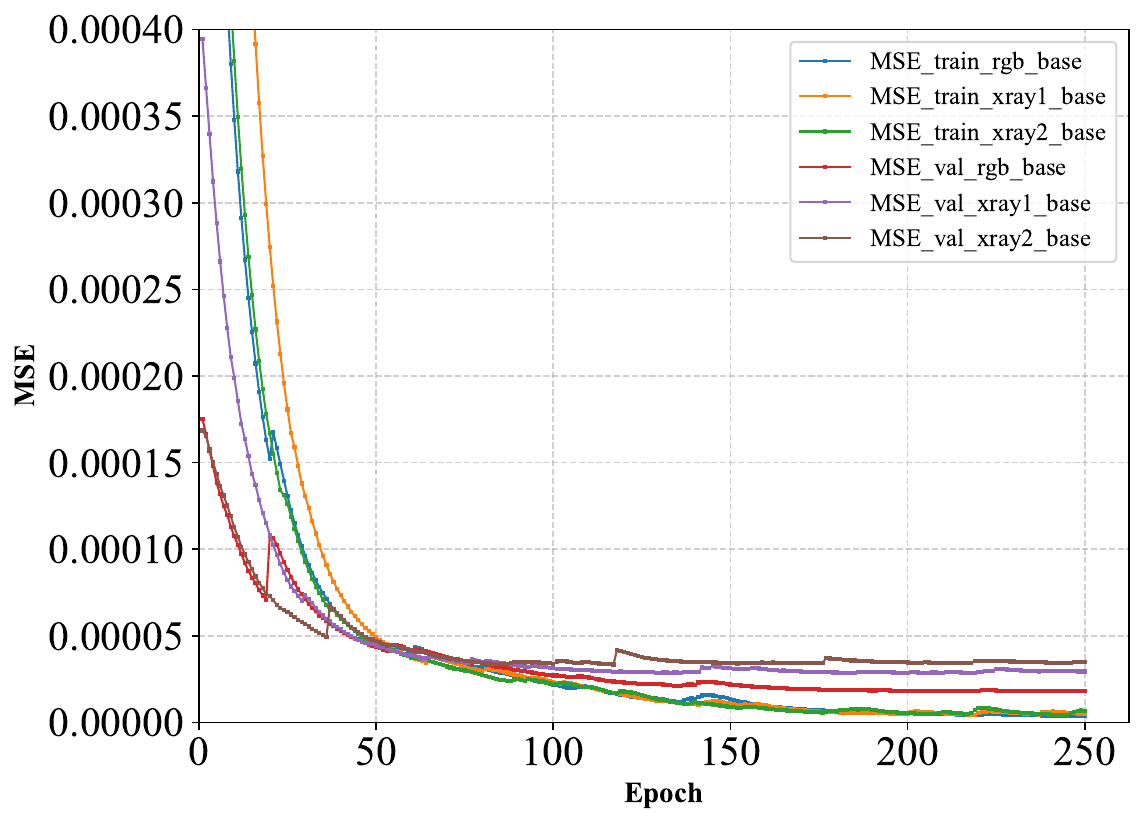}
        \label{fig:sub4}
    \end{subfigure}

    \vspace{-6mm} 

    \begin{subfigure}{.24\textwidth}
        \centering
        \renewcommand{\thesubfigure}{c}
        \includegraphics[width=\linewidth]{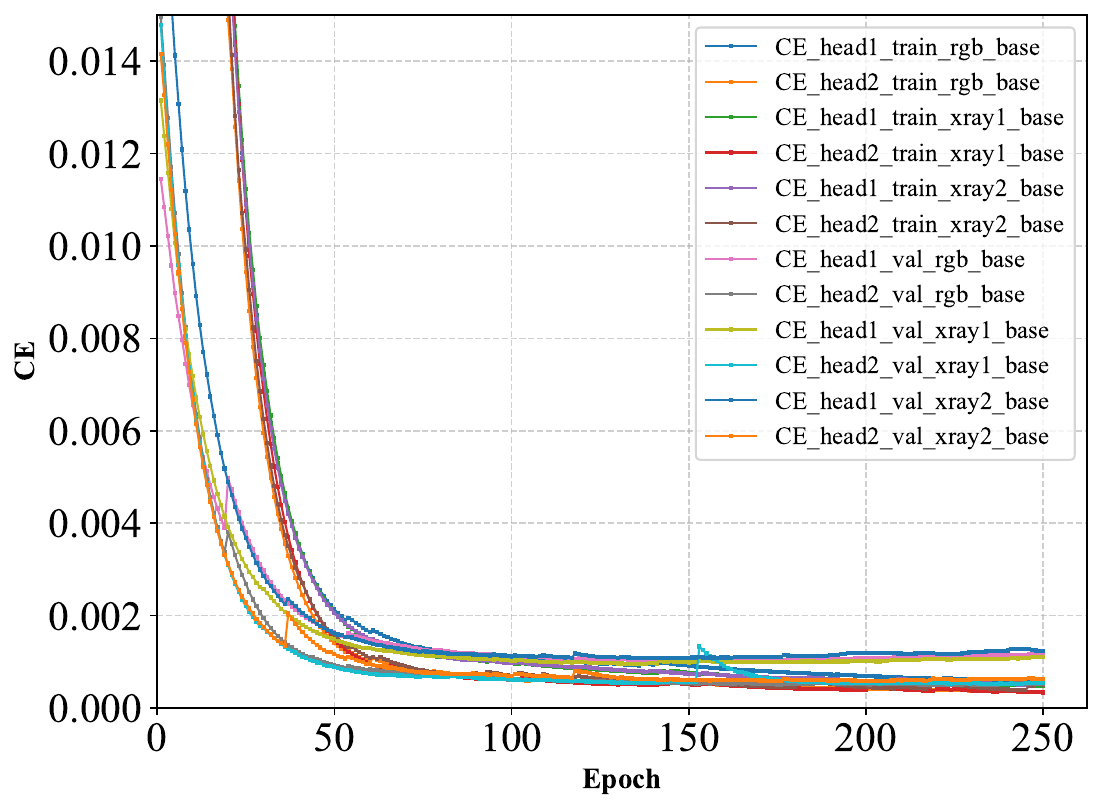}
        \caption{ViT-small} 
        \label{fig:sub5}
    \end{subfigure}
    \hspace{-2mm}
    \begin{subfigure}{.24\textwidth}
        \centering
        \renewcommand{\thesubfigure}{d}
        \includegraphics[width=\linewidth]{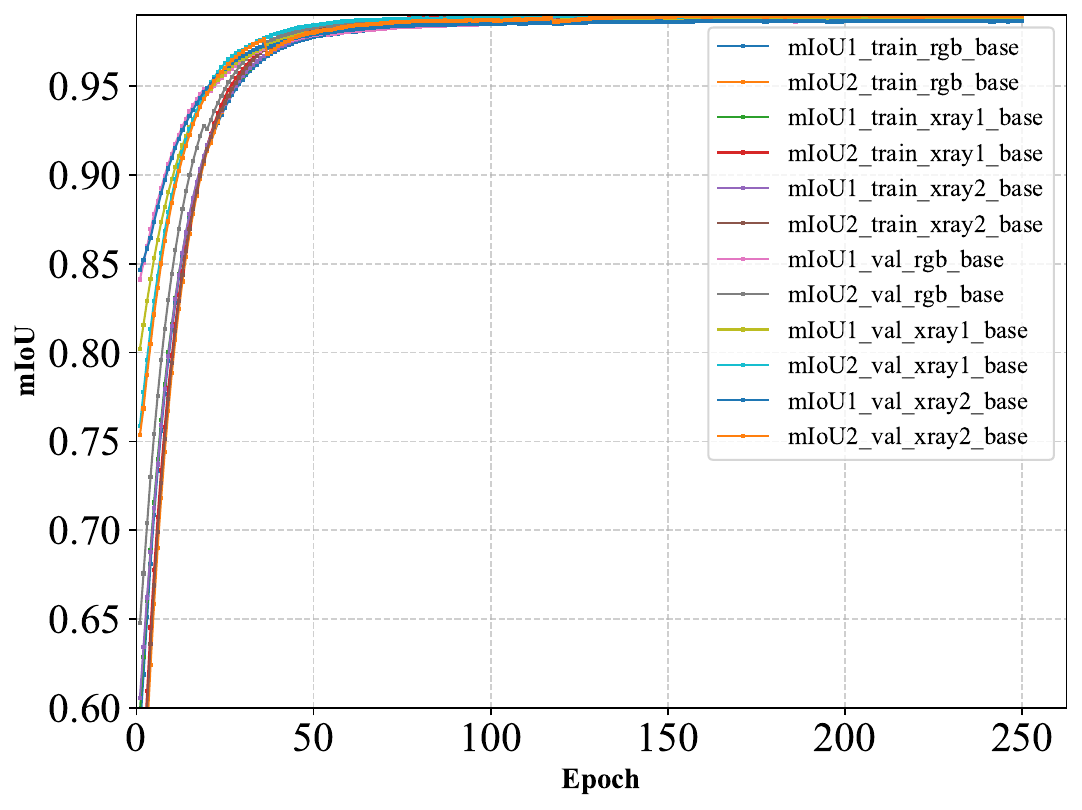}
        \caption{ViT-small} 
        \label{fig:sub6}
    \end{subfigure}
    \hspace{-2mm}
    \begin{subfigure}{.24\textwidth}
        \centering
        \renewcommand{\thesubfigure}{g}
        \includegraphics[width=\linewidth]{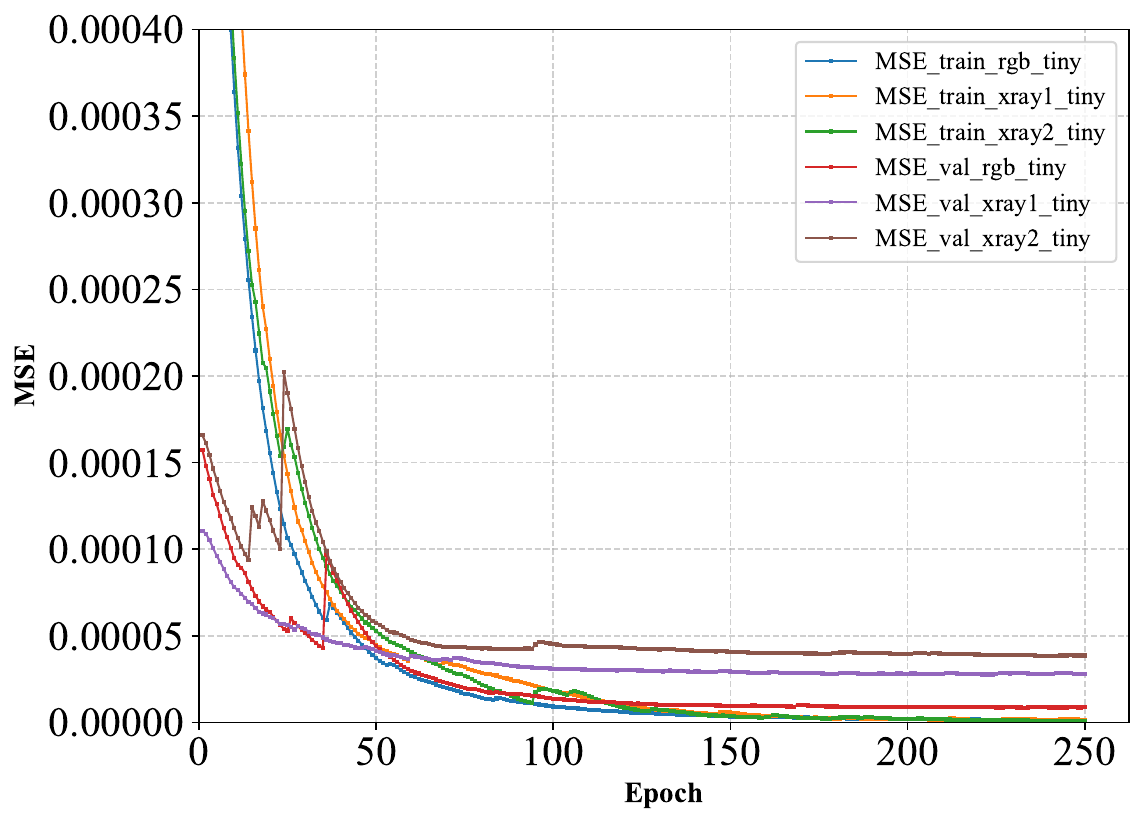}
        \caption{ViT-tiny (w/o)} 
        \label{fig:sub7}
    \end{subfigure}
    \hspace{-2mm}
    \begin{subfigure}{.24\textwidth}
        \centering
        \renewcommand{\thesubfigure}{h}
        \includegraphics[width=\linewidth]{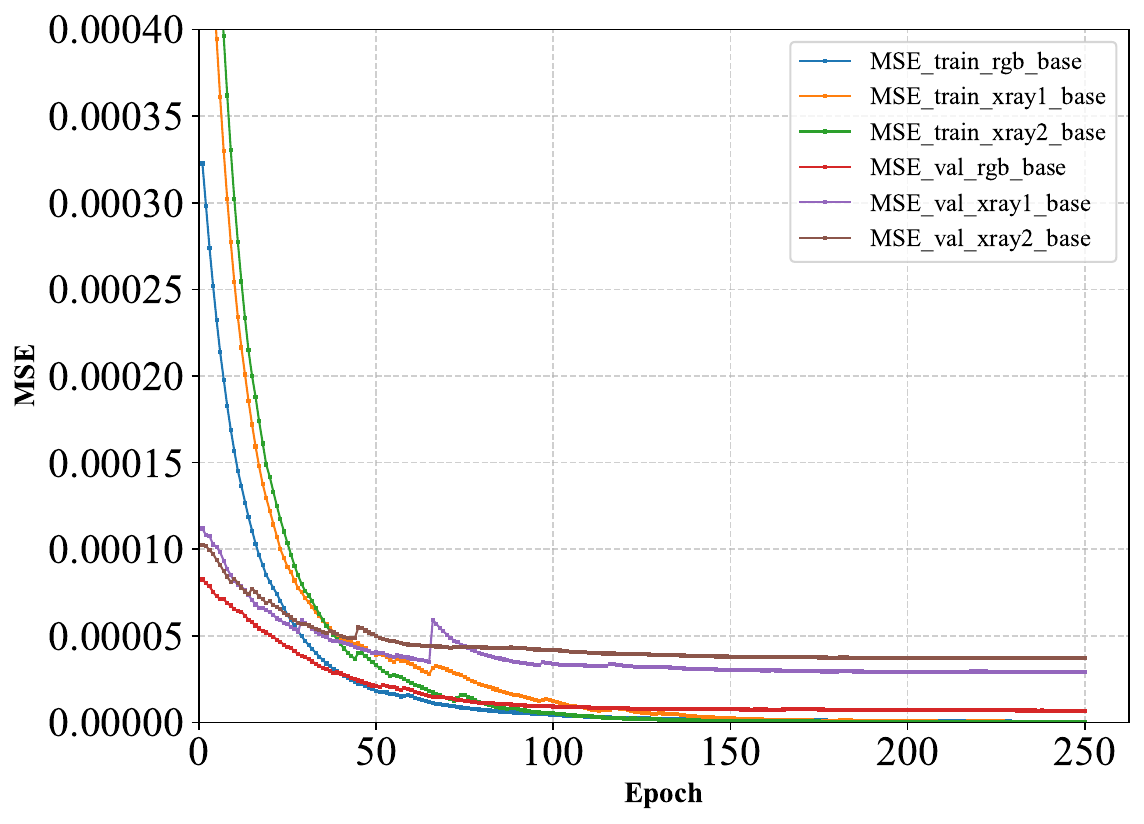}
        \caption{ViT-small (w/o)} 
        \label{fig:sub8}
    \end{subfigure}

    \caption{The diagram presents the training and validation loss curves. Subplots (a)–(d) display the segmentation loss for both tiny and small versions of TransForSeg, while subplots (e)–(h) illustrate the force estimation loss for both TransForSeg ('w' for models with segmentation heads) and TransForcer ('w/o' for models without segmentation).}
    \label{fig:test_merged}
\end{figure*}
\par TransForSeg achieved the highest mIoU and mDice scores across all three datasets, outperforming both CNN-based and Transformer-based baselines. While H-Net is also a dedicated multitask model for catheter segmentation and 3D force estimation, TransForSeg improved mIoU by $3\%$ on all datasets relative to H-Net. Despite its ViT-based encoder-decoder architecture, equipped with a fusion module, two segmentation heads, and a regression head, TransForSeg remains a lightweight model. This feature makes the model well-suited for real-time deployment. In fact, the integration of 2D segmentation with 3D force estimation not only eliminates the need for a separate segmentation model during data preprocessing for learning-based force estimation, but also serves as a perception module for both human surgeons and autonomous robotic systems. The FLOPs and parameter count highlight the capability of ViT-based models to reduce computational complexity by generating semantically and spatially rich embeddings without relying on skip connections during upsampling.
\begin{table*}[t]
\centering
\caption{Performance of Models Across Different Noises}
\label{tab:performance}
\begin{tabular}{l|p{2cm}|l|l|cccccc}
\toprule
\textbf{Noise} & \textbf{Noise Properties} & \textbf{Dataset} & \textbf{Size} & \textbf{MSE} & \textbf{MAE} & \textbf{RMSE} & \textbf{Acc(\%)} & \textbf{mIoU(\%)} & \textbf{mDice(\%)} \\
\midrule
\multirow{6}{*}{Impulse}  
    & \multirow{6}{*}{\parbox{2cm}{\raggedright Salt (255) and\\ pepper (0) \\randomly applied \\ to 20\% of \\ extreme pixels}}  
    & \multirow{2}{*}{RGB}     
    & Tiny
    & 3.82e-05 (2.1x)
    & 0.0042 (1.5x)
    & 0.0061 (1.4x)
    & 99 ($\approx$)
    & 98 (-0.5)
    & 99 (-0.2)  \\ 
    &                          &                          
    & Base  
    & 3.29e-05 (1.8x)
    & 0.0039 (1.4x)
    & 0.0057 (1.3x)
    & 99 ($\approx$)
    & 98 (-0.4)
    & 99 (-0.2)  \\ 
    \cmidrule(lr){3-10}
    &                          & \multirow{2}{*}{X-Ray1} 
    & Tiny
    & 2.19e-05 ($\approx$)
    & 0.0030 ($\approx$)
    & 0.0046 ($\approx$)
    & 99 ($\approx$)
    & 98 ($\approx$)
    & 99 ($\approx$)
    \\ 
    &                          &                          
    & Base
    & 2.84e-05 ($\approx$)
    & 0.0035 ($\approx$)
    & 0.0053 ($\approx$)
    & 99 ($\approx$)
    & 98 ($\approx$)
    & 99 ($\approx$)
    \\ 
    \cmidrule(lr){3-10}
    &                          & \multirow{2}{*}{X-Ray2} 
    & Tiny
    & 3.75e-05 (1.1x)
    & 0.0039 ($\approx$)
    & 0.0061 ($\approx$)
    & 99 ($\approx$)
    & 98 (-0.2)
    & 99 (-0.1)
    \\ 
    &                          &                          
    & Base
    & 3.56e-05 ($\approx$)
    & 0.0039 ($\approx$)
    & 0.0060 ($\approx$)
    & 99 ($\approx$)
    & 98 (-0.1)
    & 99 ($\approx$)
    \\ 
\midrule
\multirow{6}{*}{Gaussian}  
    & \multirow{6}{*}{\parbox{2cm}{\raggedright Mean: \ensuremath{\mu} = 0 \\ Std: \ensuremath{\sigma} = 0.02}}  
    & \multirow{2}{*}{RGB}     
    & Tiny
    & 6.90e-05 (3.9x)
    & 0.0059 (2.1x)
    & 0.0083 (2.0x)
    & 99 ($\approx$)
    & 96 (-2.0)
    & 98 (-1.0)
    \\ 
    &                          &                          
    & Base
    & 5.77e-05 (3.2x)
    & 0.0053 (1.9x)
    & 0.0076 (1.8x)
    & 99 ($\approx$)
    & 97 (-1.2)
    & 98 (-0.6)
    \\ 
    \cmidrule(lr){3-10}
    &                          & \multirow{2}{*}{X-Ray1} 
    & Tiny
    & 2.54e-05 (1.2x)
    & 0.0033 (1.1x)
    & 0.0050 (1.1x)
    & 99 ($\approx$)
    & 98 (-0.3)
    & 99 (-0.1)
    \\ 
    &                          &                          
    & Base
    & 3.019e-05 ($\approx$)
    & 0.0036 ($\approx$)
    & 0.0055 ($\approx$)
    & 99 ($\approx$)
    & 98 (-0.2)
    & 99 (-0.1)
    \\ 
    \cmidrule(lr){3-10}
    &                          & \multirow{2}{*}{X-Ray2} 
    & Tiny 
    & 5.42e-05 (1.6x)
    & 0.0048 (1.3x)
    & 0.0074 (1.2x) 
    & 99 ($\approx$)
    & 97 (-1.2)
    & 98 (-0.6)
    \\ 
    &                          &                          
    & Base
    & 4.78e-05 (1.4x)
    & 0.0046 (1.2x)
    & 0.0069 (1.2x)
    & 99 ($\approx$)
    & 97 (-1.1)
    & 98 (-0.5)
    \\ 
    \midrule 
\multirow{6}{*}{Poisson}  
    & \multirow{6}{*}{\parbox{2cm}{\raggedright Intensity-dependent  (\ensuremath{\lambda = I}),\\ Non-Gaussian \\noise,\\ Applied pixel-wise}}  
    & \multirow{2}{*}{RGB}     
    & Tiny
    & 4.47e-05 (2.5x)
    & 0.0046 (1.7x)
    & 0.0067 (1.6x)
    & 99 ($\approx$)
    & 98 (-1.0)
    & 99 (-0.5)
    \\ 
    &                          &                          
    & Base
    & 3.74e-05 (2.0x)
    & 0.0041 (1.2x)
    & 0.0061 (1.1x)
    & 99 ($\approx$)
    & 97 (-0.1)
    & 98 (-0.2)
    \\ 
    \cmidrule(lr){3-10}
    &                          & \multirow{2}{*}{X-Ray1} 
    & Tiny
    & 2.11e-05 ($\approx$)
    & 0.0030 ($\approx$)
    & 0.0045 ($\approx$)
    & 99 ($\approx$)
    & 98 ($\approx$)
    & 99 ($\approx$)
    \\ 
    &                          &                          
    & Base
    & 2.82e-05 ($\approx$)
    & 0.0034 ($\approx$)
    & 0.0053 ($\approx$)
    & 99 ($\approx$)
    & 98 ($\approx$)
    & 99 ($\approx$)
    \\ 
    \cmidrule(lr){3-10}
    &                          & \multirow{2}{*}{X-Ray2} 
    & Tiny
    & 3.47e-05 ($\approx$)
    & 0.0038 ($\approx$)
    & 0.0059 ($\approx$)
    & 99 ($\approx$)
    & 98 ($\approx$)
    & 99 ($\approx$)
    \\ 
    &                          &                          
    & Base
    & 3.35e-05 ($\approx$)
    & 0.0038 ($\approx$)
    & 0.0058 ($\approx$)
    & 99 ($\approx$)
    & 98 ($\approx$)
    & 99 ($\approx$)
    \\
    \midrule 
\multirow{6}{*}{Motion}  
    & \multirow{6}{*}{\parbox{2cm}{\raggedright Kernel Size (k=6), Motion Angle \\ (\ensuremath{\theta} = 20),\\ Blur Direction,\\ Linear Spread}}  
    & \multirow{2}{*}{RGB}     
    & Tiny
    & 6.01e-05 (4.6x)
    & 0.0053 (2.3x)
    & 0.0078 (x2.1x)
    & 99 (-0.1)
    & 92 (-6.4)
    & 95 (-3.5)
    \\ 
    &                          &                          
    & Base
    & 5.66e-05 (4.8x)
    & 0.0051 (2.5x)
    & 0.0075 (2.2x)
    & 99 (-0.1)
    & 92 (-6.1)
    & 95 (-3.3)
    \\ 
    \cmidrule(lr){3-10}
    &                          & \multirow{2}{*}{X-Ray1} 
    & Tiny
    & 6.26e-05 (2.9x)
    & 0.0054 (1.9x)
    & 0.0079 (1.7x)
    & 99 (-0.1)
    & 92 (-6.4)
    & 95 (-3.5)
    \\ 
    &                          &                          
    & Base
    & 6.19e-05 (2.7x)
    & 0.0054 (2.0x) 
    & 0.0079 (1.6x)
    & 99 (-0.1)
    & 92 (-6.4)
    & 95 (-3.5)
    \\ 
    \cmidrule(lr){3-10}
    &                          & \multirow{2}{*}{X-Ray2} 
    & Tiny
    & 6.31e-05 (1.7x)
    & 0.0054 (1.4x)
    & 0.0079 (1.3x)
    & 99 (-0.1)
    & 92 (-6.4)
    & 95 (-3.5)
    \\ 
    &                          &                          
    & Base
    & 5.64e-05 (1.4x)
    & 0.0052 (1.4x)
    & 0.0075 (1.2x)
    & 99 (-0.2)
    & 92 (-6.2)
    & 96 (-3.4)
    \\
    \midrule 
\multirow{6}{*}{Defocus}  
    & \multirow{6}{*}{\parbox{2cm}{\raggedright Kernel Size \\ (k=10),\\ Blur Radius \\ (\ensuremath{\sigma}=2.0), \\ Isotropic Blur}}  
    & \multirow{2}{*}{RGB}     
    & Tiny
    & 5.79e-05 (3.2x)
    & 0.0052 (1.9x)
    & 0.076 (1.8x)
    & 99 ($\approx$)
    & 97 (-1.7)
    & 98 (-0.9)
    \\ 
    &                          &                          
    & Base
    & 4.29e-05 (2.4x)
    & 0.0044 (1.7x)
    & 0.065 (1.5x)
    & 99 ($\approx$)
    & 97 (-1.0)
    & 98 (-0.5)
    \\ 
    \cmidrule(lr){3-10}
    &                          & \multirow{2}{*}{X-Ray1} 
    & Tiny
    & 6.74e-05 (3.3x)
    & 0.0053 (1.8x)
    & 0.0082 (1.8x)
    & 99 ($\approx$)
    & 96 (-2.6)
    & 97 (-1.5)
    \\ 
    &                          &                          
    & Base
    & 7.75e-05 (2.8x)
    & 0.0055 (1.6x)
    & 0.0087 (1.6x)
    & 99 ($\approx$)
    & 95 (-3.0)
    & 97 (-1.6)
    \\ 
    \cmidrule(lr){3-10}
    &                          & \multirow{2}{*}{X-Ray2} 
    & Tiny
    & 7.85e-05 (2.3x)
    & 0.0059 (1.6x)
    & 0.0089 (1.5x)
    & 99 ($\approx$)
    & 95 (-3.3)
    & 97 (-1.8)
    \\ 
    &                          &                          
    & Base
    & 7.16e-05 (2.1x)
    & 0.0057 (1.5x)
    & 0.0085 (1.4x)
    & 99 ($\approx$)
    & 96 (-2.7)
    & 97 (-1.4)
    \\
    \midrule 
\multirow{6}{*}{Stripe}  
    & \multirow{6}{*}{\parbox{2cm}{\raggedright Intensity \ensuremath{\alpha}=0.1,\\ Direction ="vertical",\\ Numbers (N=50),\\ Max Width (w=10)}}  
    & \multirow{2}{*}{RGB}& Tiny
    & 2.13e-04 (12.0x)
    & 0.0104 (3.8x)
    & 0.0143 (3.4x)
    & 99 (-0.2)
    & 92 (-6.0)
    & 96 (-3.3)
    \\ 
    & &             
    & Base
    & 2.18e-04 (12.0x)
    & 0.0101 (3.6x)
    & 0.0144 (3.4x)
    & 99 (-0.6)
    & 96 (-2.6)
    & 98 (-1.3)
    \\ 
    \cmidrule(lr){3-10}
    &                          & \multirow{2}{*}{X-Ray1} 
    & Tiny
    & 2.39e-05 (1.2x)
    & 0.0032 ($\approx$)
    & 0.0049 ($\approx$)
    & 99 (-0.9)
    & 98 (-0.1)
    & 99 ($\approx$)
    \\ 
    &                          &                          
    & Base
    & 2.98e-05 ($\approx$)
    & 0.0036 ($\approx$)
    & 0.0055 ($\approx$)
    & 99 ($\approx$)
    & 98 ($\approx$)
    & 99 ($\approx$)
    \\ 
    \cmidrule(lr){3-10}
    &                          & \multirow{2}{*}{X-Ray2} 
    & Tiny
    & 3.48e-05 ($\approx$)
    & 0.0038 ($\approx$)
    & 0.0059 ($\approx$)
    & 98 ($\approx$)
    & 98 ($\approx$)
    & 99 ($\approx$)
    \\ 
    &                          &                          
    & Base
    & 3.35e-05 ($\approx$)
    & 0.0038 ($\approx$)
    & 0.0058 ($\approx$)
    & 99 ($\approx$)
    & 98 ($\approx$)
    & 99 ($\approx$)
    \\
\bottomrule
\end{tabular}
\end{table*}

\subsection{Ablation}
As previously discussed, the segmentation heads in TransForSeg play a crucial role in enabling the model to estimate catheter contact forces directly from unsegmented X-ray images. The segmentation task helps the model focus more effectively on catheter deflections, reducing the influence of background variations during force estimation. To assess the contribution of the segmentation task, we developed a modified version of the model, named TransForcer, in which the two segmentation heads were removed, converting it into a single-task model. Like TransForSeg, TransForcer processes input images from two different angles via its encoder and decoder. The embeddings produced by the encoder are fused into the decoder, and the [CLS] token from the decoder is passed to a regression head for 3D force estimation. Unlike TransForSeg, TransForcer does not reshape the embeddings for segmentation, as it omits the segmentation heads entirely. The model architecture of TransForcer is identical to that of TransForSeg, except for the removal of the segmentation heads. This model was trained in both tiny and small configurations, using the same training settings described for TransForSeg, and repeated three times on each of the training sets: RGB, X-Ray1, and X-Ray2. Instead of the multitask loss function defined in Equation (7), TransForcer was optimized using the regression-only loss function defined in Equation (6). Table III compares the force estimation results of TransForSeg (denoted as "with" segmentation head) and TransForcer (denoted as "without" segmentation head) on the corresponding test sets. 
\par 
Considering X-Ray2, the most challenging dataset, the addition of the segmentation head improved the MSE by $15.9\%$ for the tiny version and $20.66\%$ for the small version of TransForSeg. A similar trend is observed on the X-Ray1 dataset, where the segmentation head led to MSE improvements of $28.6\%$ and $8.8\%$ for the tiny and small models, respectively. These results highlight the positive impact of integrating the segmentation task on the precision of the force estimation head when working with synthetically generated X-ray images, where the background is independent of the catheter's deflection. However, this benefit is not observed on the RGB dataset, where models without the segmentation head outperform those with the segmentation head in both model sizes. This discrepancy could be attributed to the fact that, in RGB images, background elements such as the 3D-printed surface of the force sensor and shadows are correlated with catheter movement, inadvertently aiding force estimation without the need for explicit segmentation (Please see Fig 4). Fig 2 presents the histograms of prediction errors on the test sets of RGB, X-Ray1, and X-Ray2, highlighting the positive impact of the segmentation head on force estimation accuracy. The top row displays the error distributions for TransForSeg-tiny and TransForSeg-small (both equipped with segmentation heads), while the bottom row shows the corresponding errors for TransForcer-tiny and TransForcer-small, which is not equipped with segmentation heads. Fig 3 shows the training and validation loss curves across epochs for both TransForSeg and TransForcer on all three datasets. Subplots (a)–(d) depict the cross-entropy loss and mIoU for both segmentation heads in the tiny and small configurations of TransForSeg. The curves demonstrate smooth convergence with no observable signs of overfitting, indicating stable learning behavior. Subplots (e)–(h) compare the regression loss curves of the force estimation head for both TransForSeg and TransForcer in their respective model sizes. As observed, the inclusion of segmentation heads helps mitigate overfitting; particularly on the more challenging X-Ray2 dataset by providing additional supervision and encouraging better generalization. 
\par \noindent \textbf{Domain Shift and Generalization}: Real-world deployment of segmentation and force estimation models frequently involves input data that deviate from the training distribution, highlighting the importance of evaluating robustness to domain shift and generalization. Having already assessed TransForSeg's performance across different imaging modalities; namely, RGB and synthetic X-ray datasets (X-Ray1 and X-Ray2); we further investigate its resilience under input-level perturbations. To simulate realistic deployment conditions, we apply six common noise types to the test sets: Impulse, Gaussian, Poisson, Motion blur, Defocus, and Stripe artifacts. 
\par This extended evaluation enables a comprehensive assessment of the model’s robustness and generalization across both tasks: semantic segmentation and 3D force estimation, in the presence of real-world image corruptions. Table IV reports the performance of TransForSeg-tiny and TransForSeg-small on these noisy test sets. In this setting, the source domain corresponds to the training sets of RGB, X-Ray1, and X-Ray2, while the target domains are their respective noisy test sets. It is important to note that the models were not exposed to any of the noise types during training, and no domain generalization, domain adaptation, or data augmentation techniques were applied at either training or test time. \par The statistics/information related to generation of each noise is explained in the table as well. For instance, the defocus simulation is applied by an isotropic Gaussian blur with a kernel size of $k=10$ and a blur radius (standard deviation) of $\sigma = 2.0$. The term isotropic blur indicates that the blurring effect is applied uniformly in all directions, meaning the image is smoothed equally along both horizontal and vertical axes. The kernel size 
$k$ defines the size of the convolution kernel used to perform the blur, while $\sigma$ determines the extent of the smoothing: larger values result in more intense blur. This process simulates the loss of focus that can occur due to depth-of-field limitations or optical defocusing in real-world medical imaging systems. 
The results demonstrate that TransForSeg maintains robust performance on the X-Ray1 and X-Ray2 datasets when subjected to Stripe artifacts, Poisson noise, and Impulse noise, showing no notable degradation in either segmentation accuracy or force estimation performance. \par In contrast, other noise types, such as Gaussian, Motion blur, and Defocus, negatively impacted the model, resulting in a $1.5\times$ to $3\times$increase in MSE and a $2\%$ to $6\%$ reduction in mIoU for the segmentation heads. Despite these challenges introduced by domain shift and noise perturbations, TransForSeg continues to exhibit competitive performance when compared to other methods listed in Table I and Table II, all of which were evaluated on clean, noise-free images. The noisy RGB images proved to be more challenging than the two previously discussed datasets, particularly for the force estimation task. The presence of stripe artifacts resulted in a 12$\times$ increase in MSE, while other noise types led to degradation factors ranging from $1.8\times$ to $4.8\times$. This sensitivity may be attributed to the nature of the RGB backgrounds, which contain structured elements; such as the force sensor and its surroundings that are more easily disrupted by noise. In contrast, the synthetic X-ray backgrounds were randomly generated, encouraging the model to focus more directly on the catheter itself and better tolerate image variability. Interestingly, the segmentation performance (mIoU) on the noisy RGB images remained relatively stable, following a degradation trend similar to that observed on the synthetic X-ray datasets. Fig 4 illustrates the quantitative performance of the segmentation heads on the RGB, X-Ray1, and X-Ray2 datasets, both with and without added noise. It is worth noting that test-time training and domain generalization techniques can be applied in real-world deployments to enhance performance without requiring additional domain-specific data collection or retraining \cite{domain-general_mehradad, rec-ttt, watt-david}.

\section{Conclusion}
\begin{figure*}[b]
\centering
\setlength{\tabcolsep}{1pt}
\renewcommand{\arraystretch}{0.7}

\begin{tabular}{c*{6}{c}} 
& \textbf{\scriptsize RGB-Side} & \textbf{\scriptsize RGB-Top} & \textbf{\scriptsize XRay1-Side} & 
   \textbf{\scriptsize XRay1-Top} & \textbf{\scriptsize XRay2-Side} & \textbf{\scriptsize XRay2-Top}\\ 

\rotatebox{90}{\scriptsize Normal} & 
\includegraphics[width=0.15\textwidth]{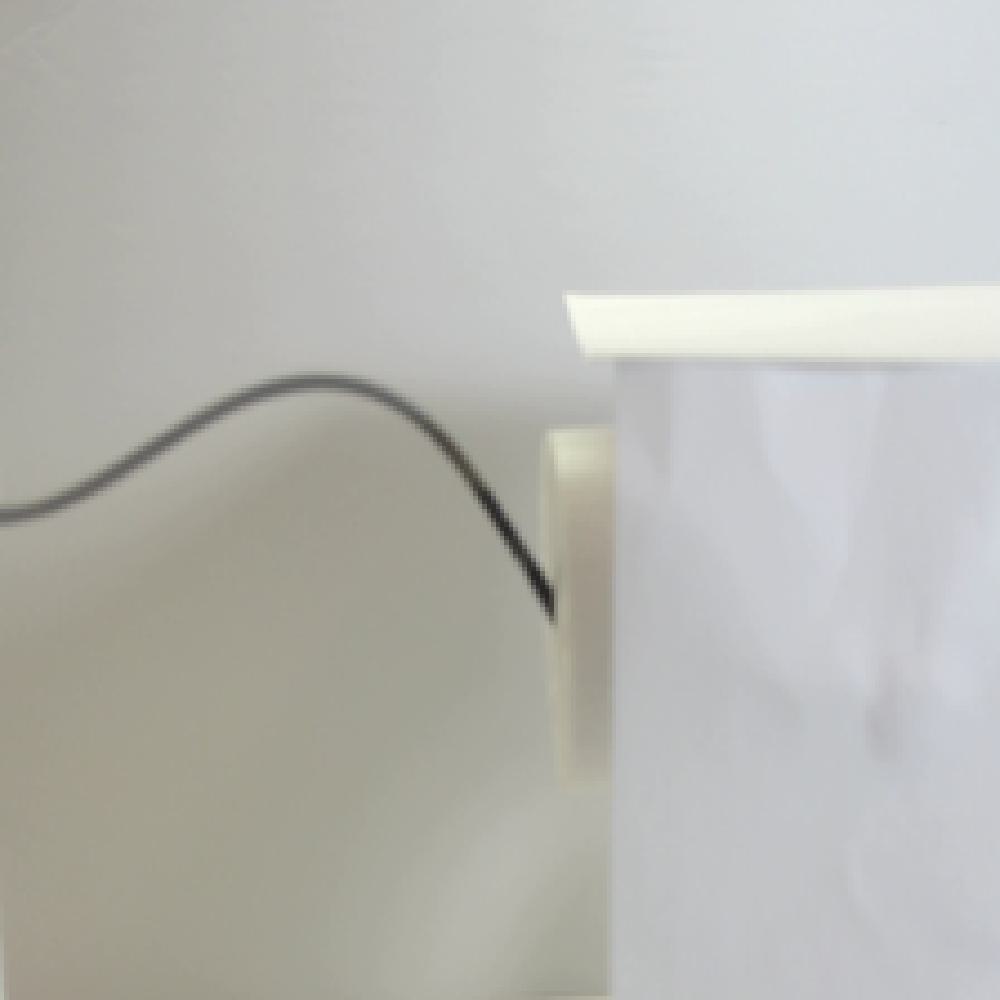} & \includegraphics[width=0.15\textwidth]{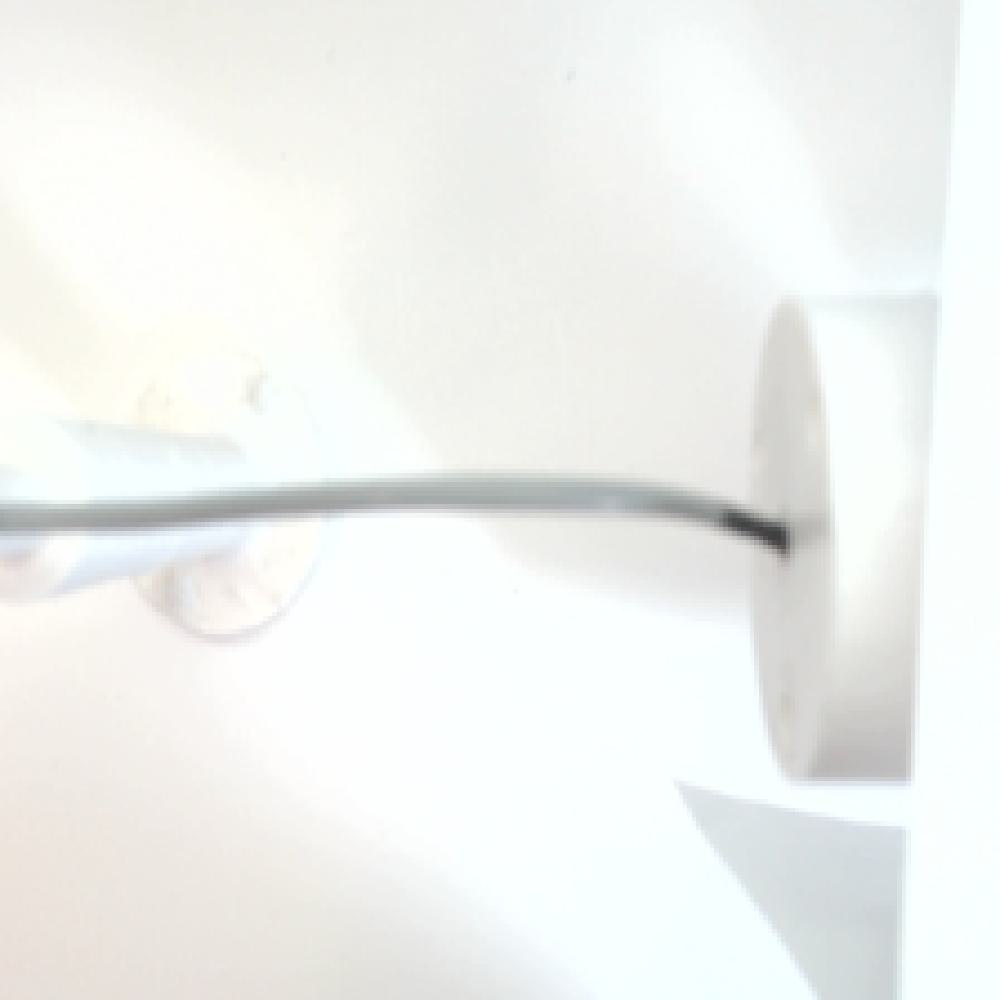} &
\includegraphics[width=0.15\textwidth]{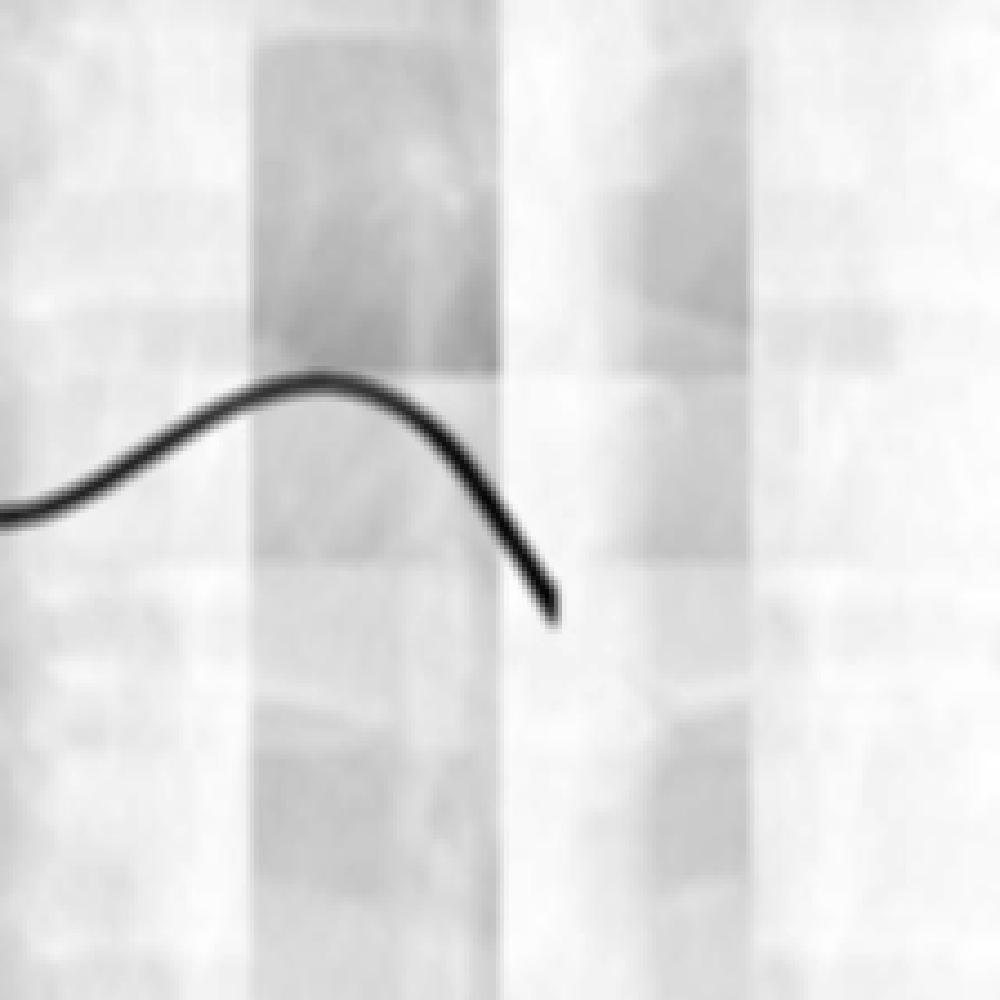} & \includegraphics[width=0.15\textwidth]{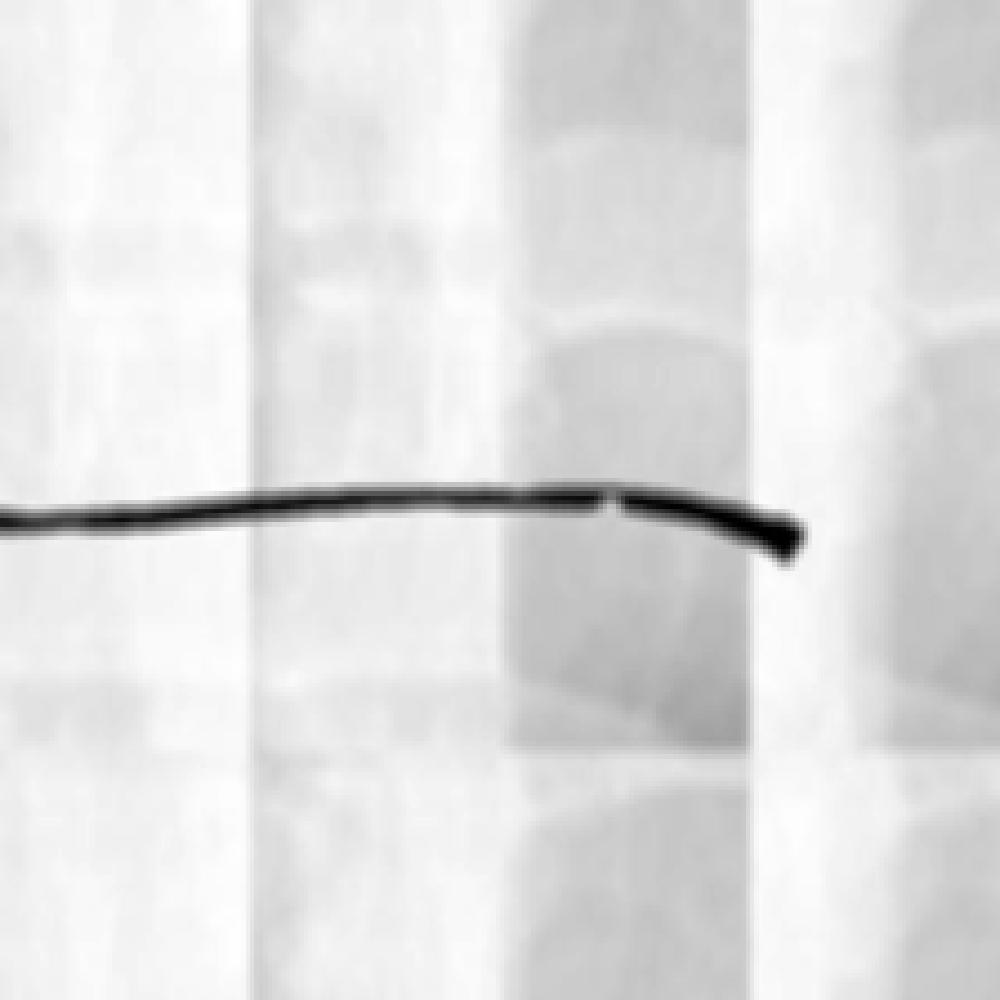} &
\includegraphics[width=0.15\textwidth]{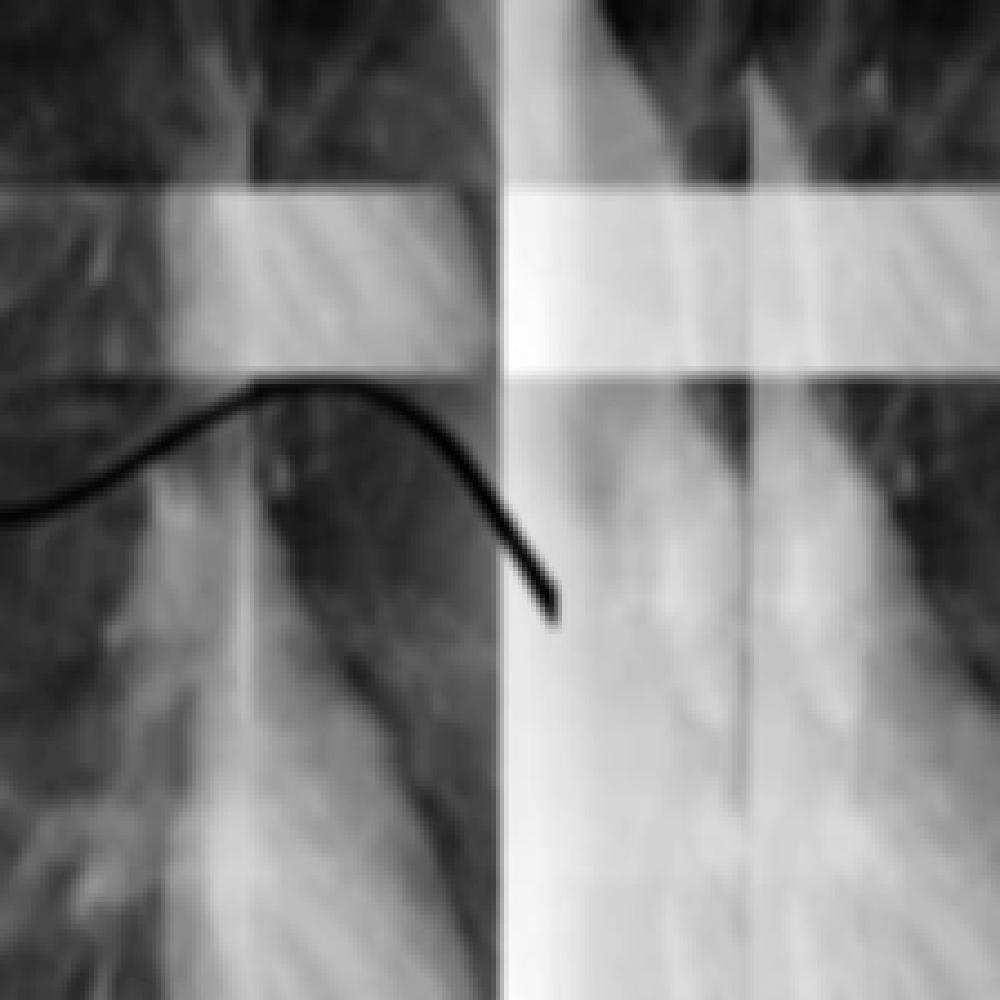} & \includegraphics[width=0.15\textwidth]{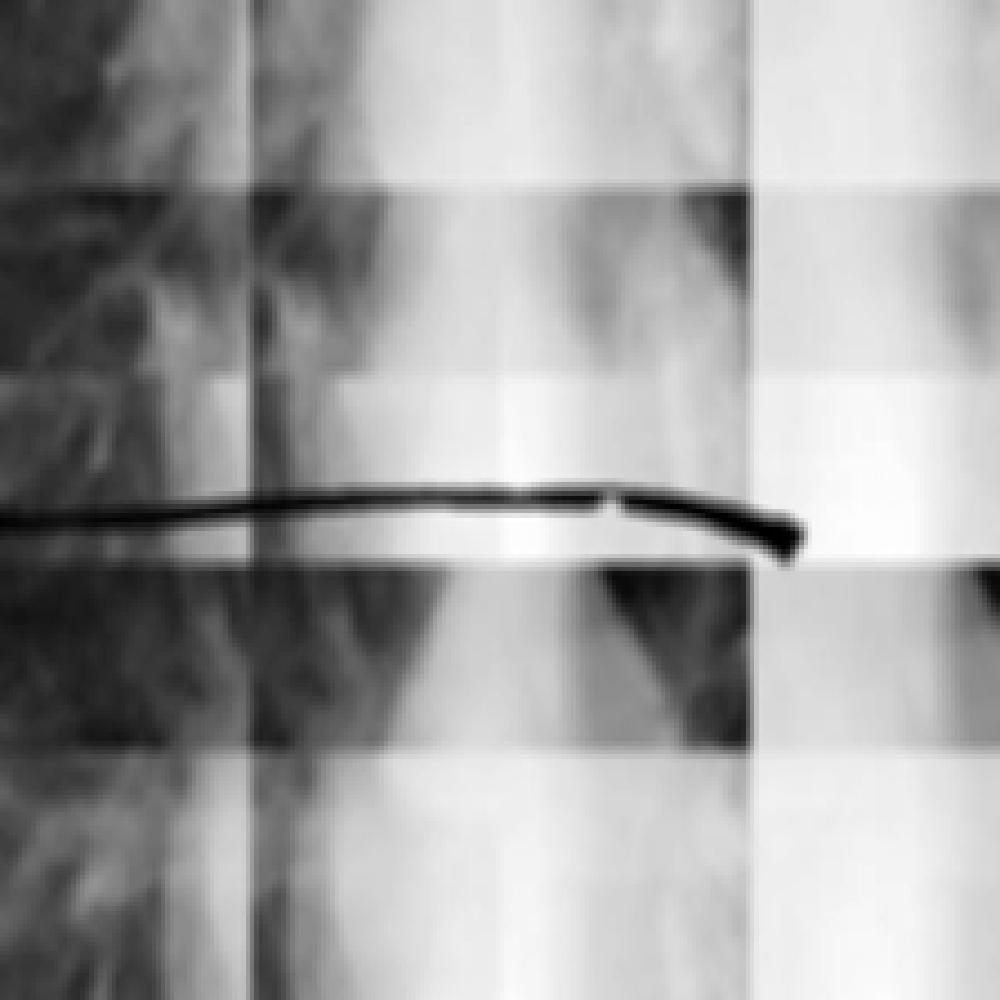} \\ 

\rotatebox{90}{\scriptsize segmentation} & 
\includegraphics[width=0.15\textwidth]{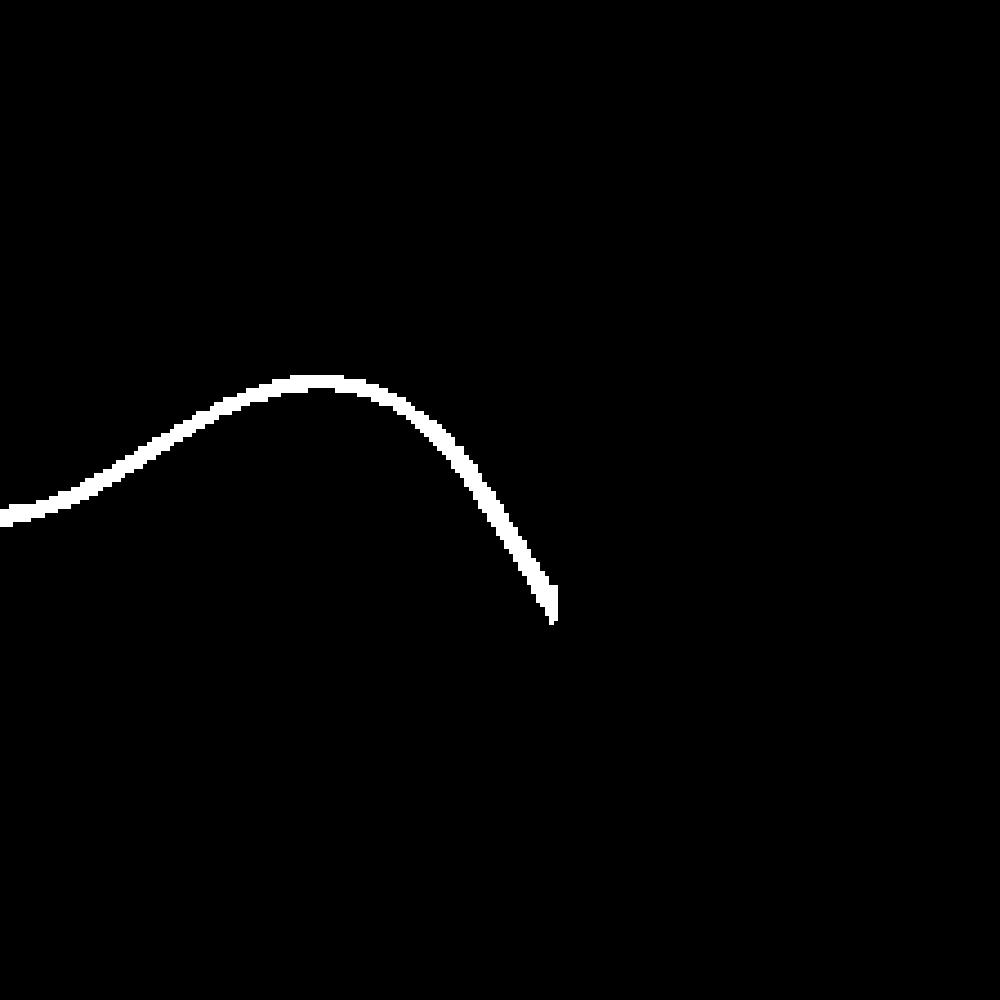} & \includegraphics[width=0.15\textwidth]{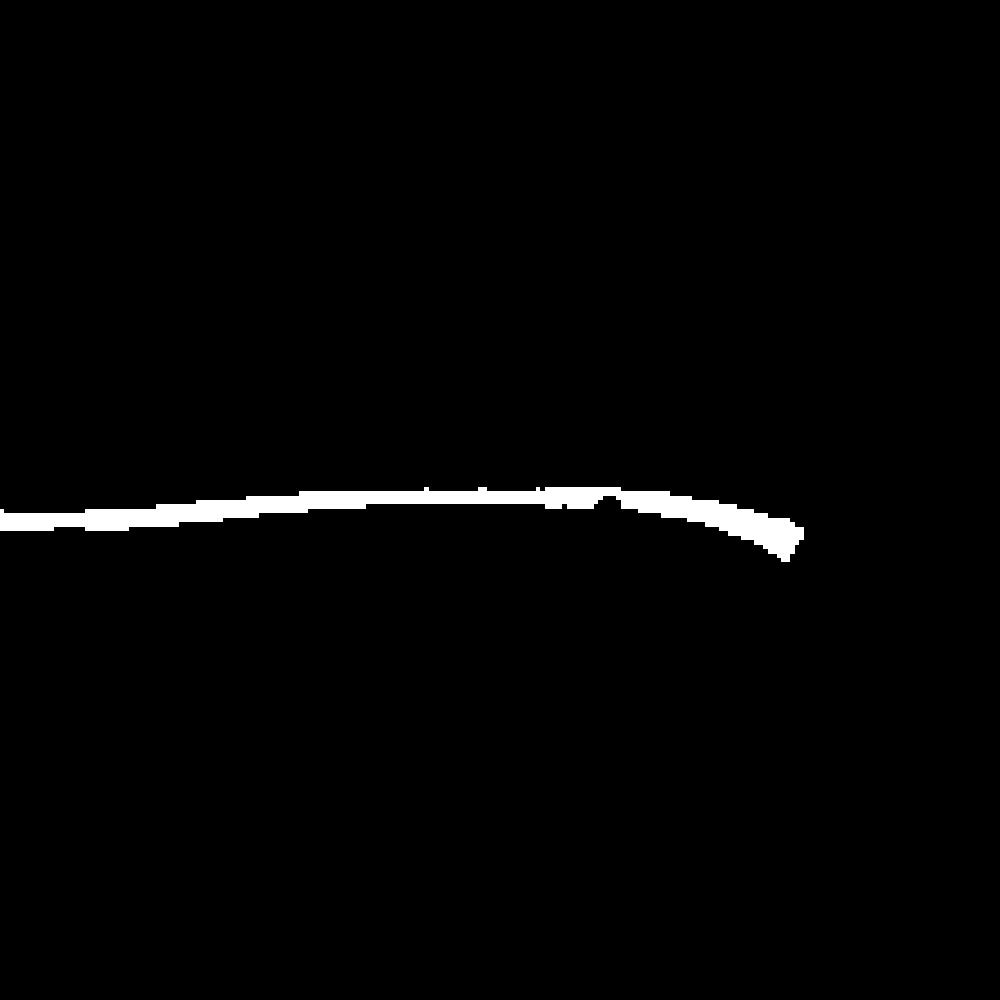} &
\includegraphics[width=0.15\textwidth]{Pictures/transforceg/images/segmentation_maps/rgb/side/image_results/254SideR2L0_rgb_im.jpg} & \includegraphics[width=0.15\textwidth]{Pictures/transforceg/images/segmentation_maps/rgb/top/image_results/254TopR2L0_rgb_im.jpg} &
\includegraphics[width=0.15\textwidth]{Pictures/transforceg/images/segmentation_maps/rgb/side/image_results/254SideR2L0_rgb_im.jpg} & \includegraphics[width=0.15\textwidth]{Pictures/transforceg/images/segmentation_maps/rgb/top/image_results/254TopR2L0_rgb_im.jpg} \\ 

\rotatebox{90}{\scriptsize Defocus} & 
\includegraphics[width=0.15\textwidth]{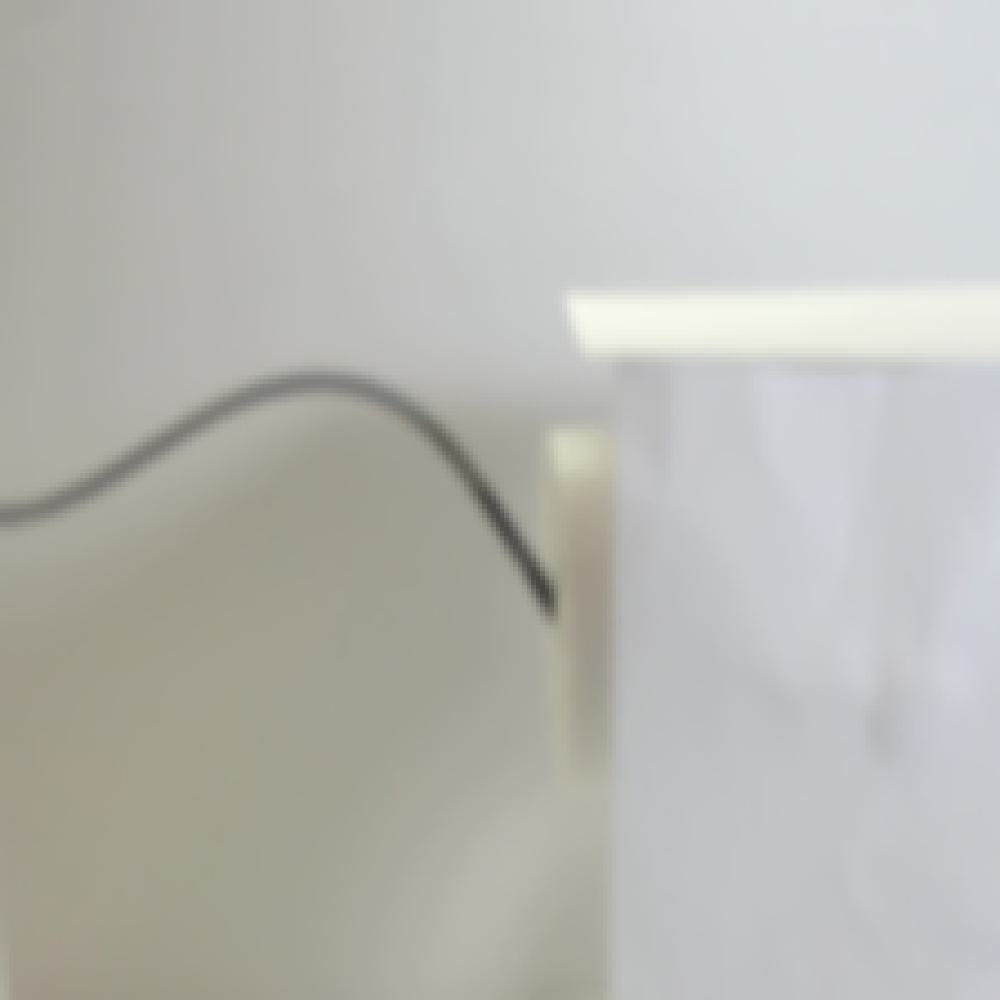} & \includegraphics[width=0.15\textwidth]{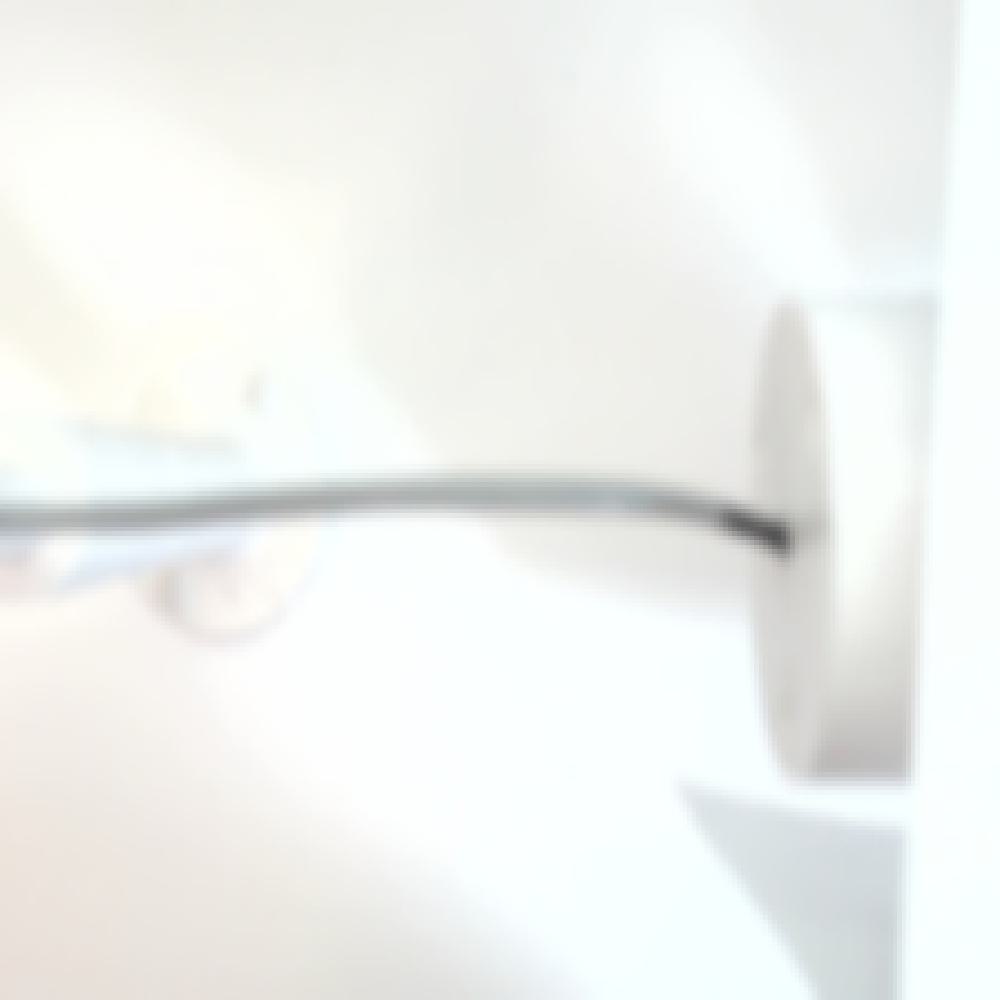} &
\includegraphics[width=0.15\textwidth]{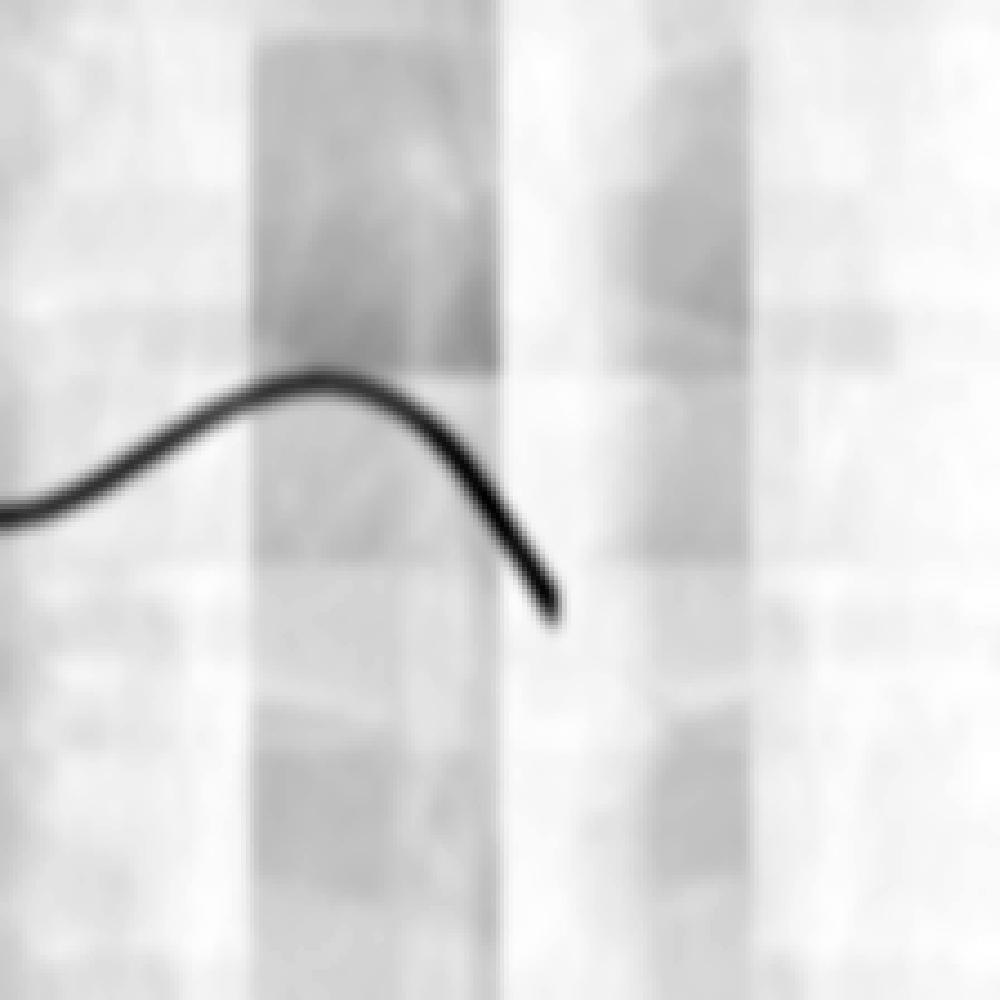} & \includegraphics[width=0.15\textwidth]{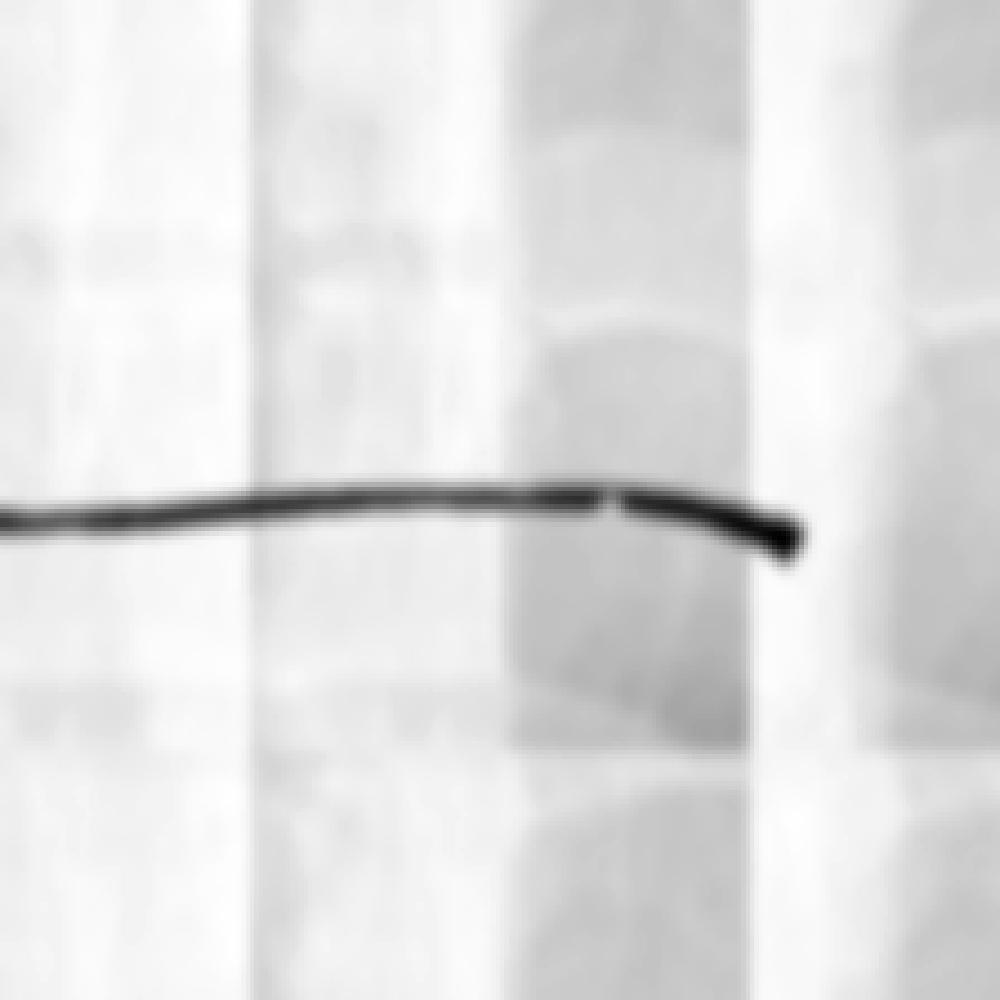} &
\includegraphics[width=0.15\textwidth]{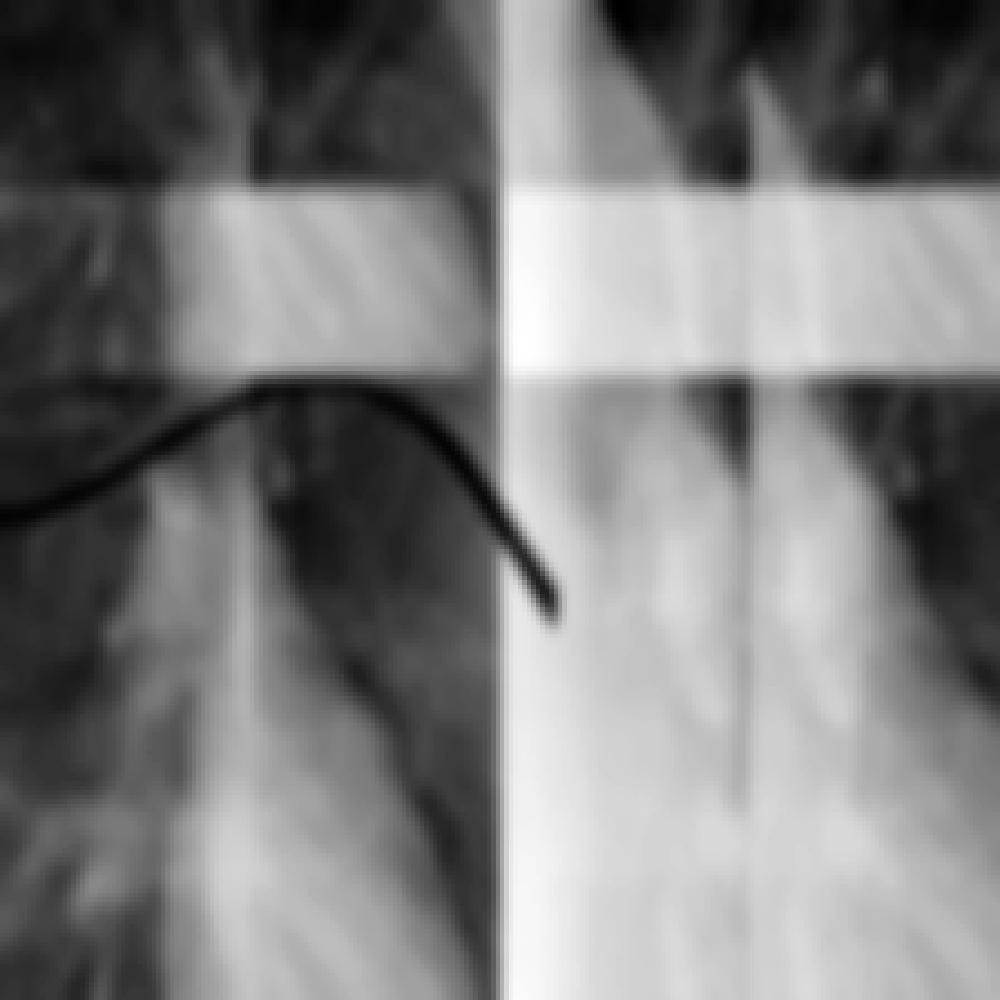} & \includegraphics[width=0.15\textwidth]{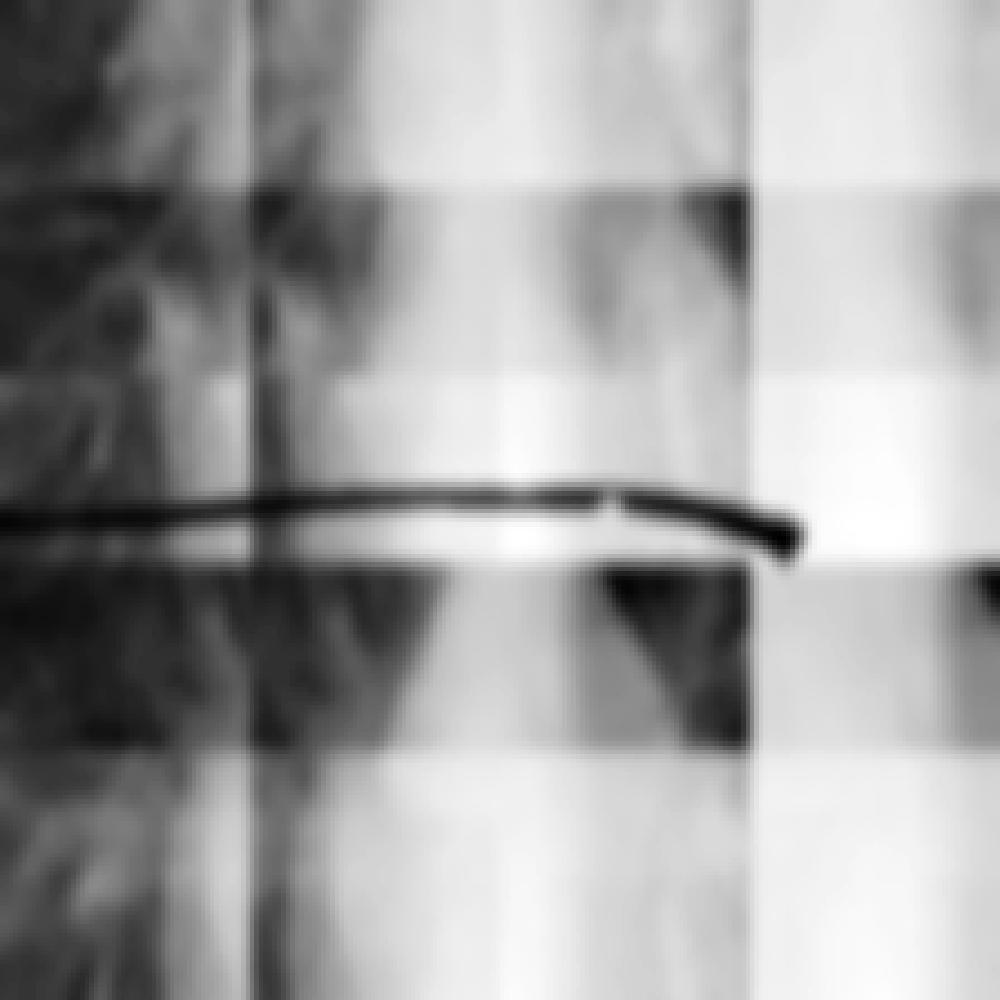} \\ 

\rotatebox{90}{\scriptsize segmentation} & 
\includegraphics[width=0.15\textwidth]{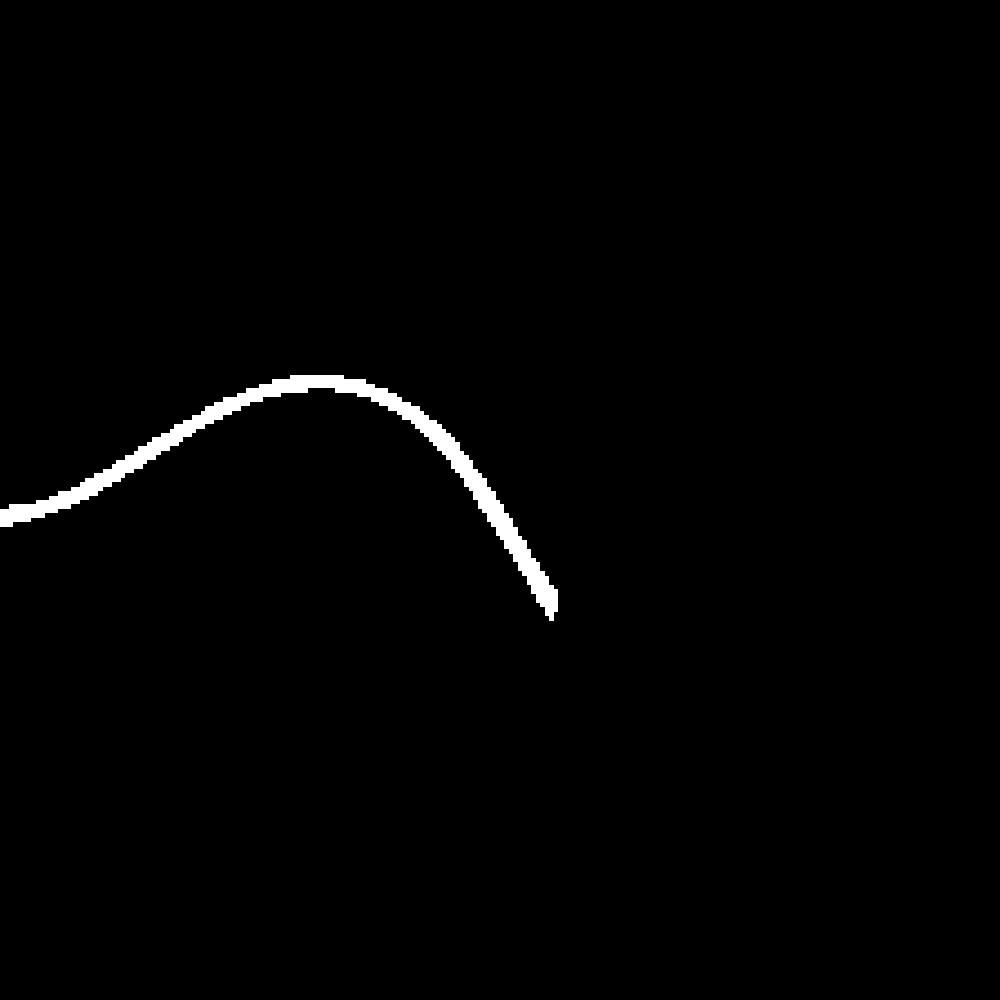} & \includegraphics[width=0.15\textwidth]{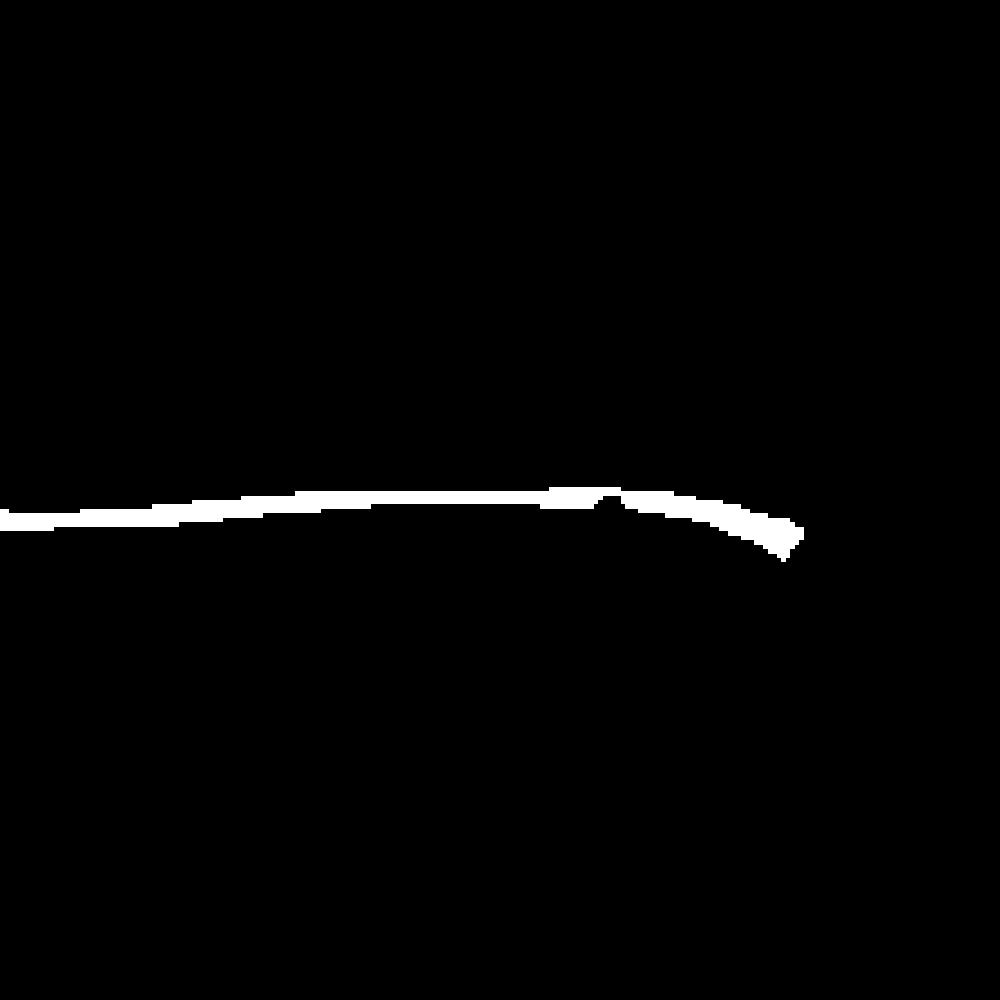} &
\includegraphics[width=0.15\textwidth]{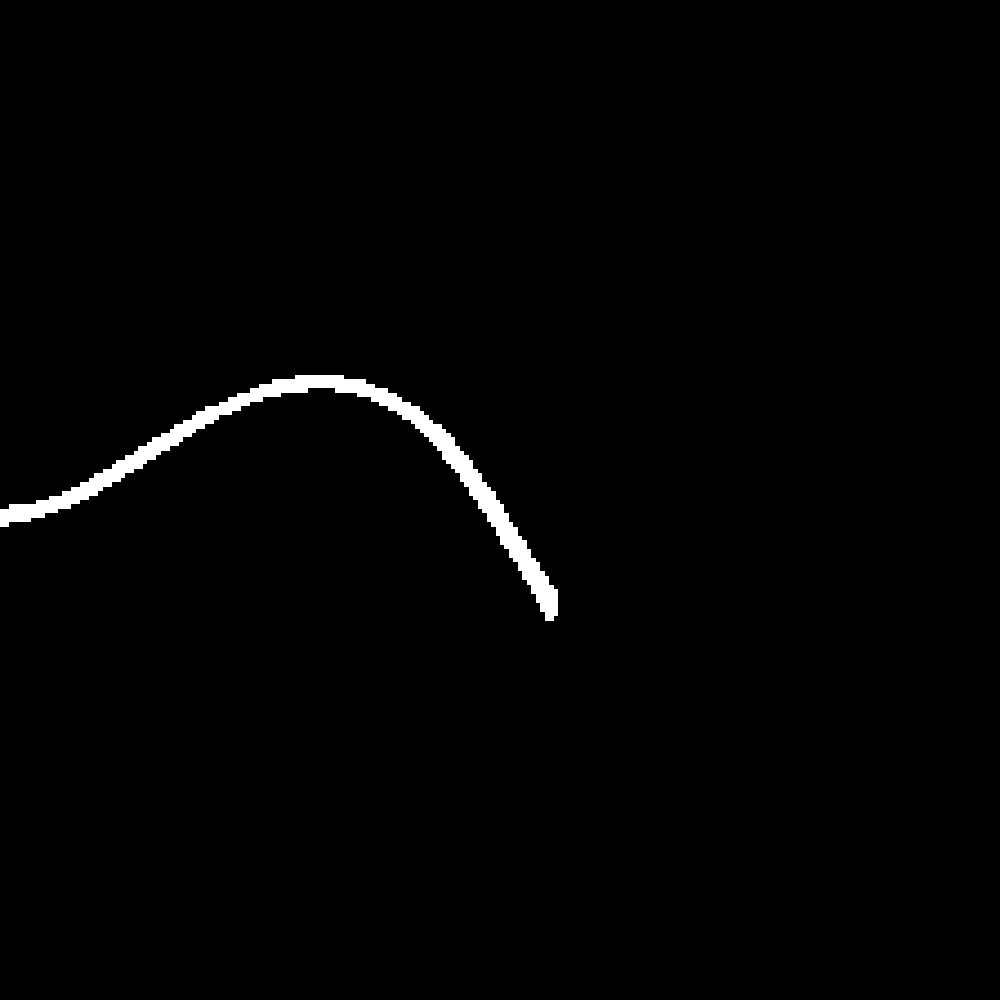} & \includegraphics[width=0.15\textwidth]{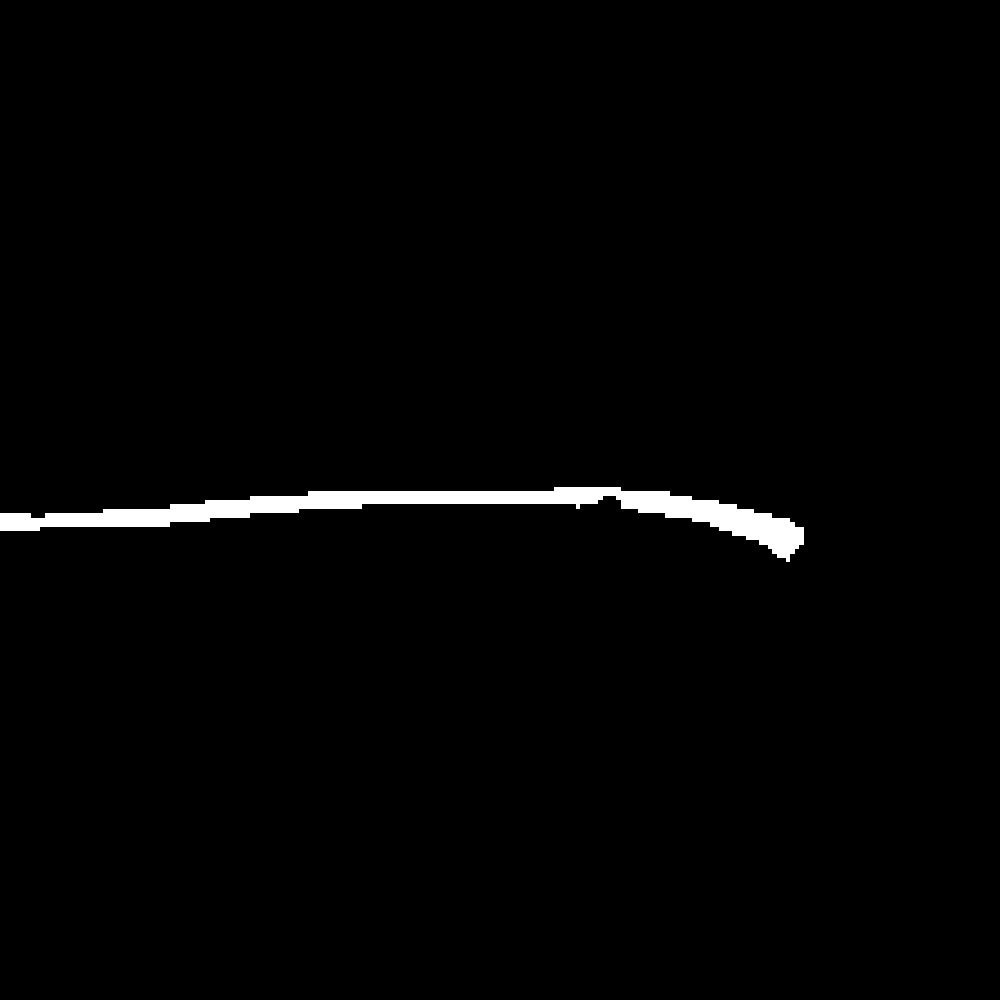} &
\includegraphics[width=0.15\textwidth]{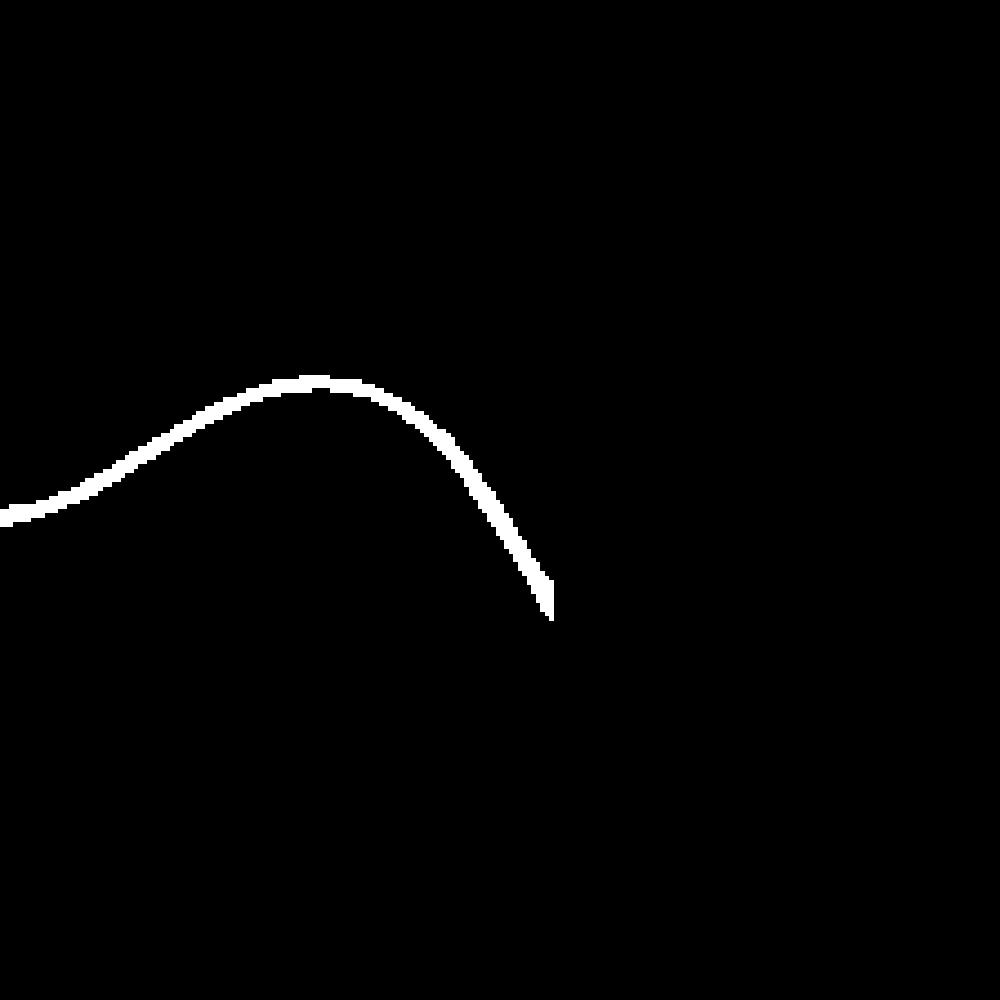} & \includegraphics[width=0.15\textwidth]{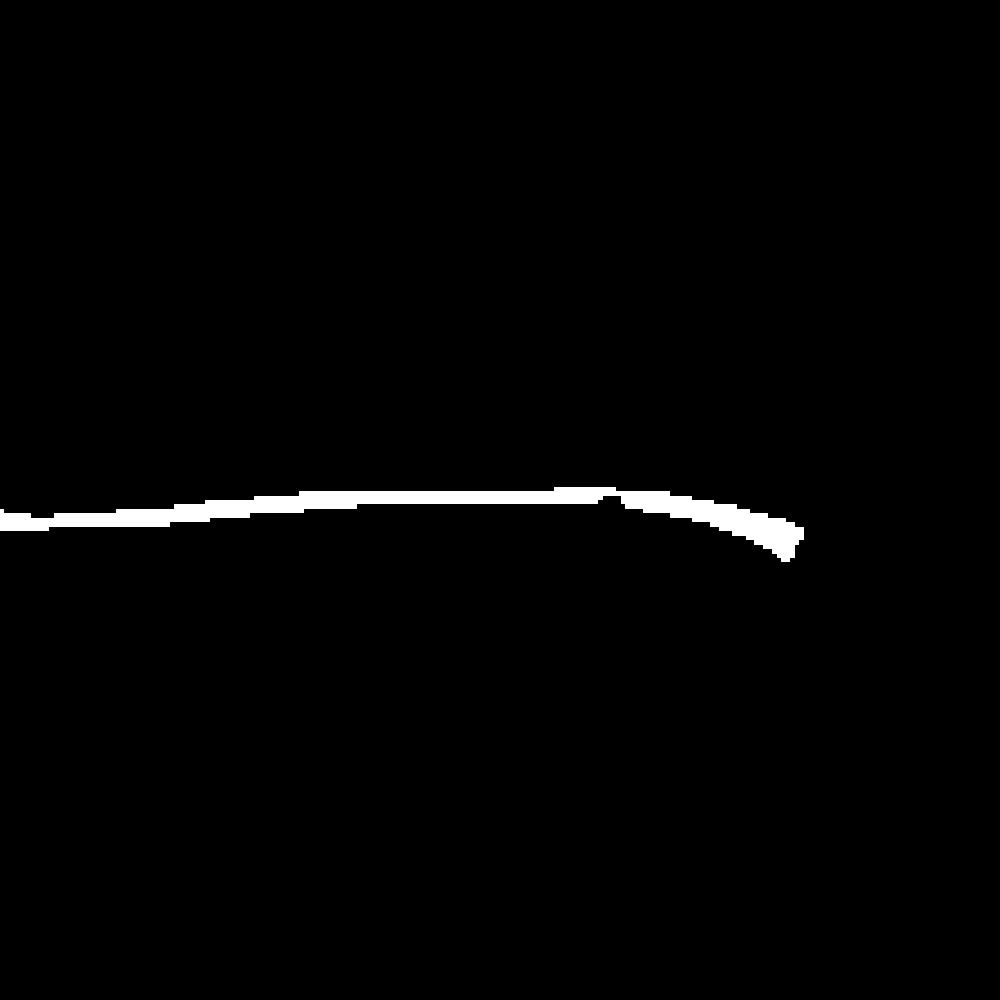} \\ 

\rotatebox{90}{\scriptsize Gaussian} & 
\includegraphics[width=0.15\textwidth]{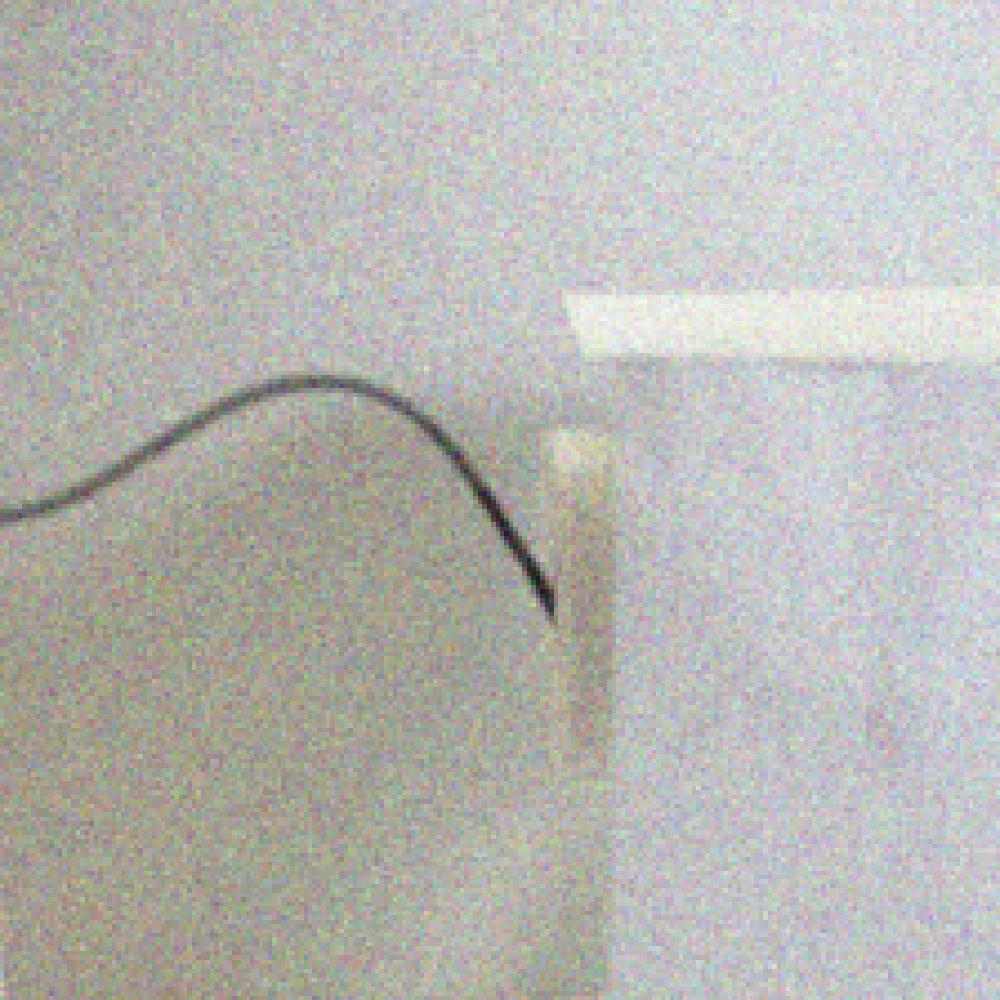} & \includegraphics[width=0.15\textwidth]{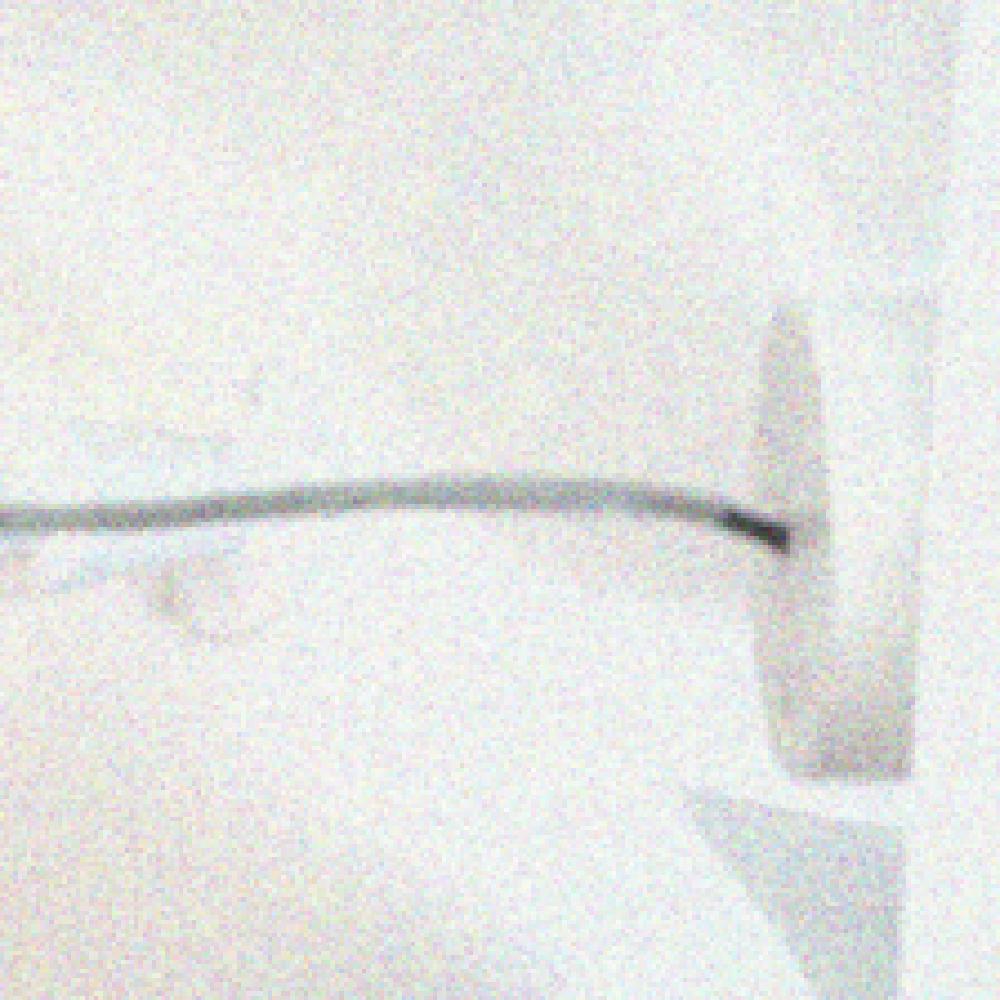} &
\includegraphics[width=0.15\textwidth]{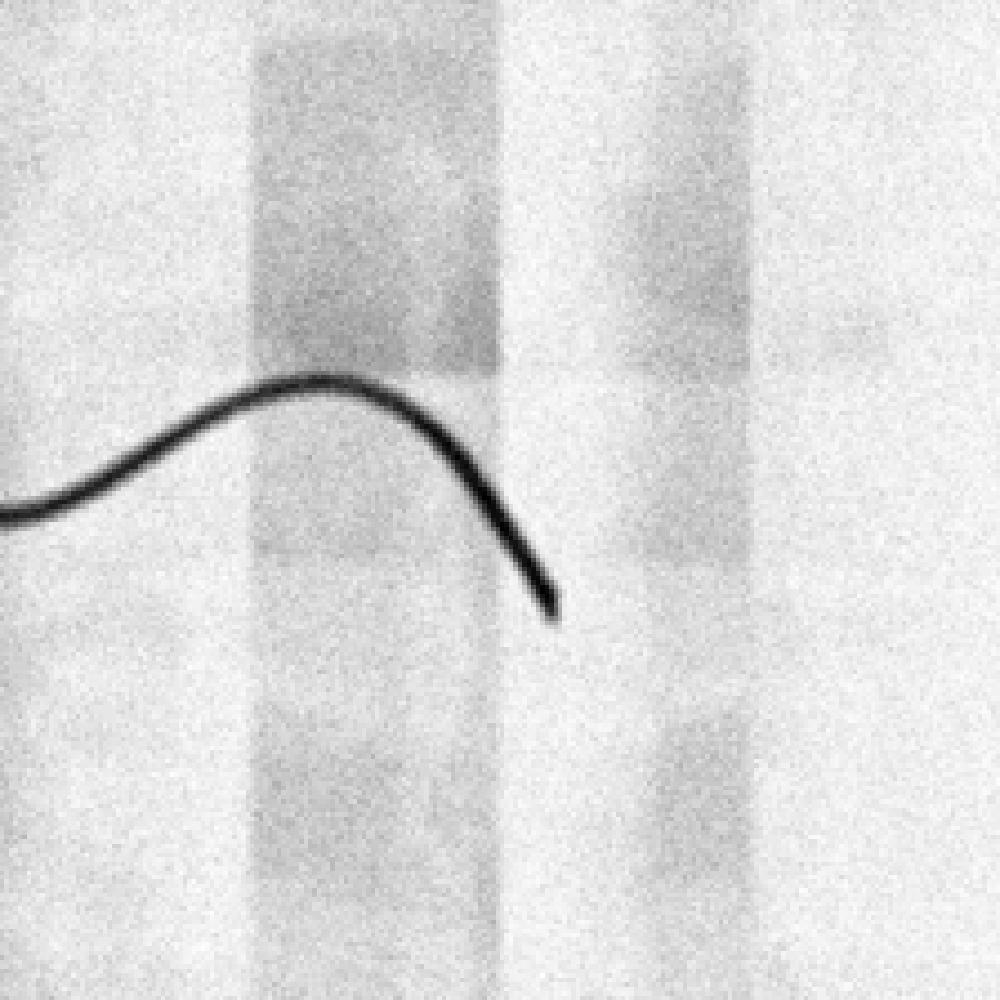} & \includegraphics[width=0.15\textwidth]{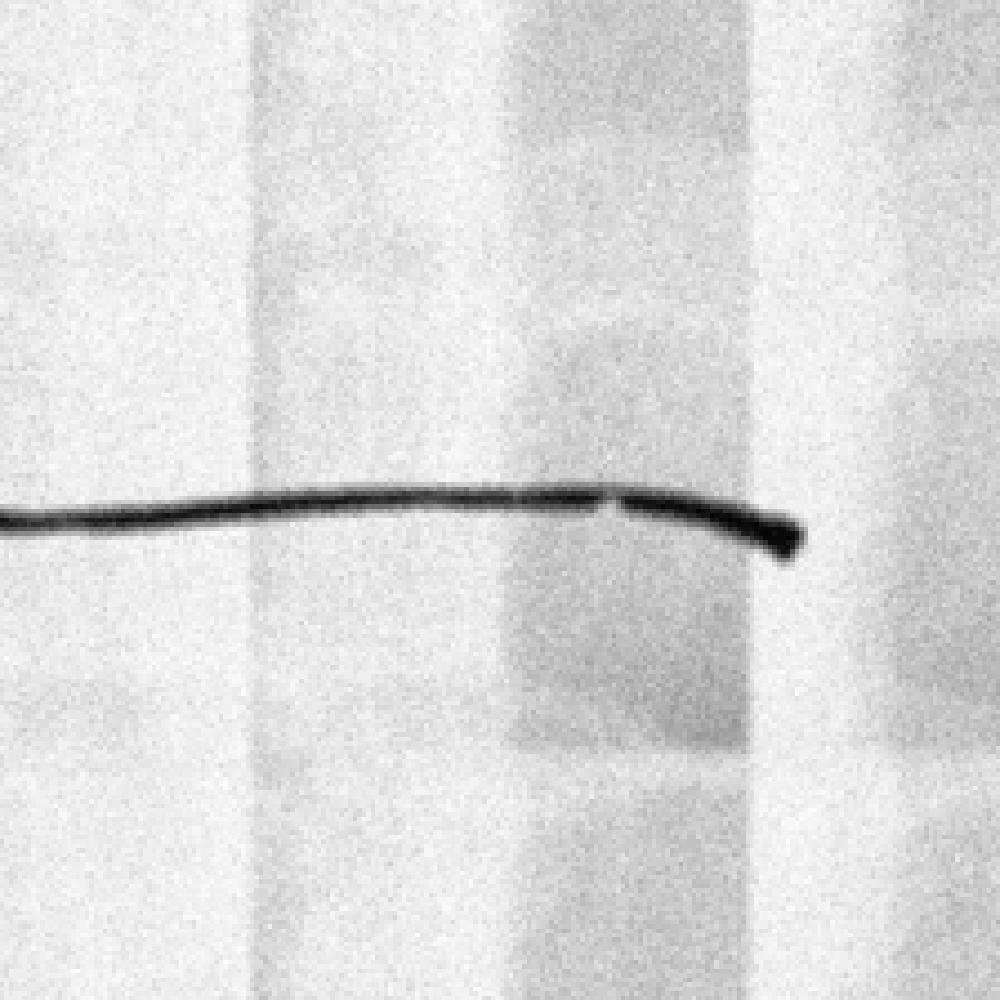} &
\includegraphics[width=0.15\textwidth]{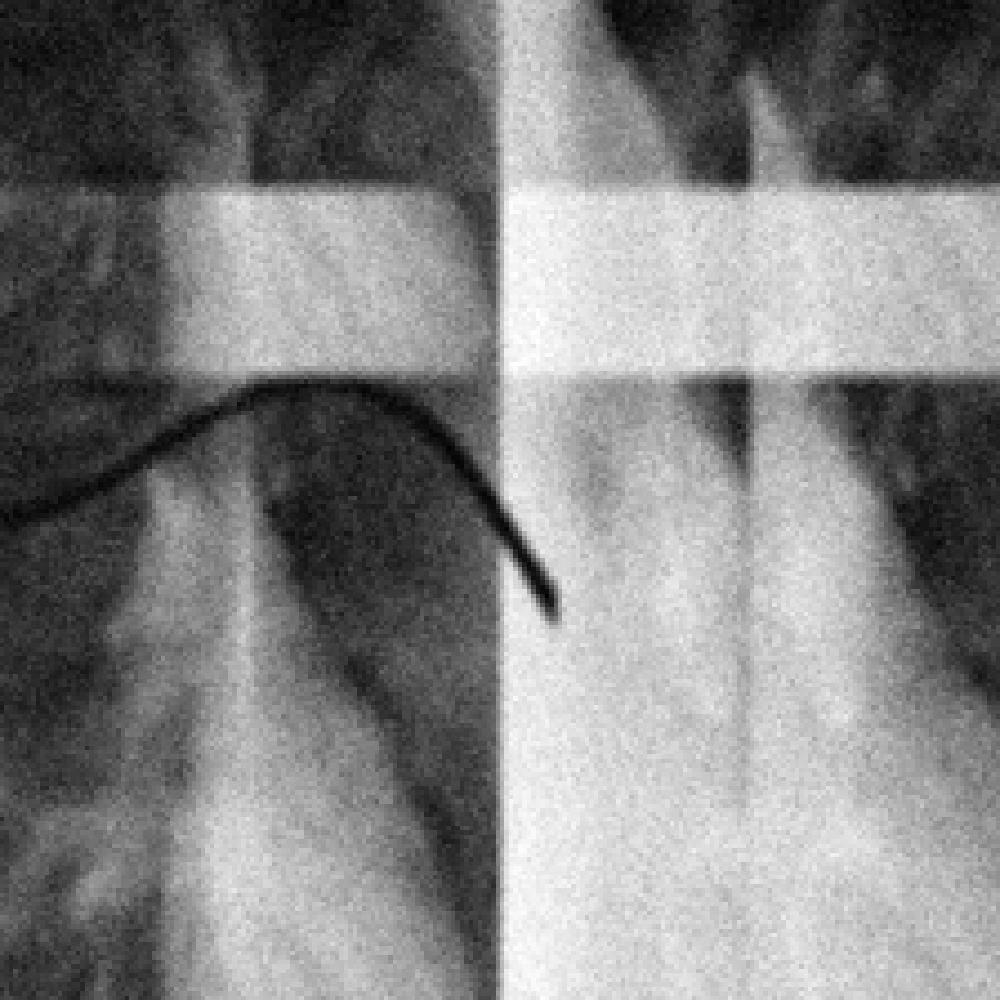} & \includegraphics[width=0.15\textwidth]{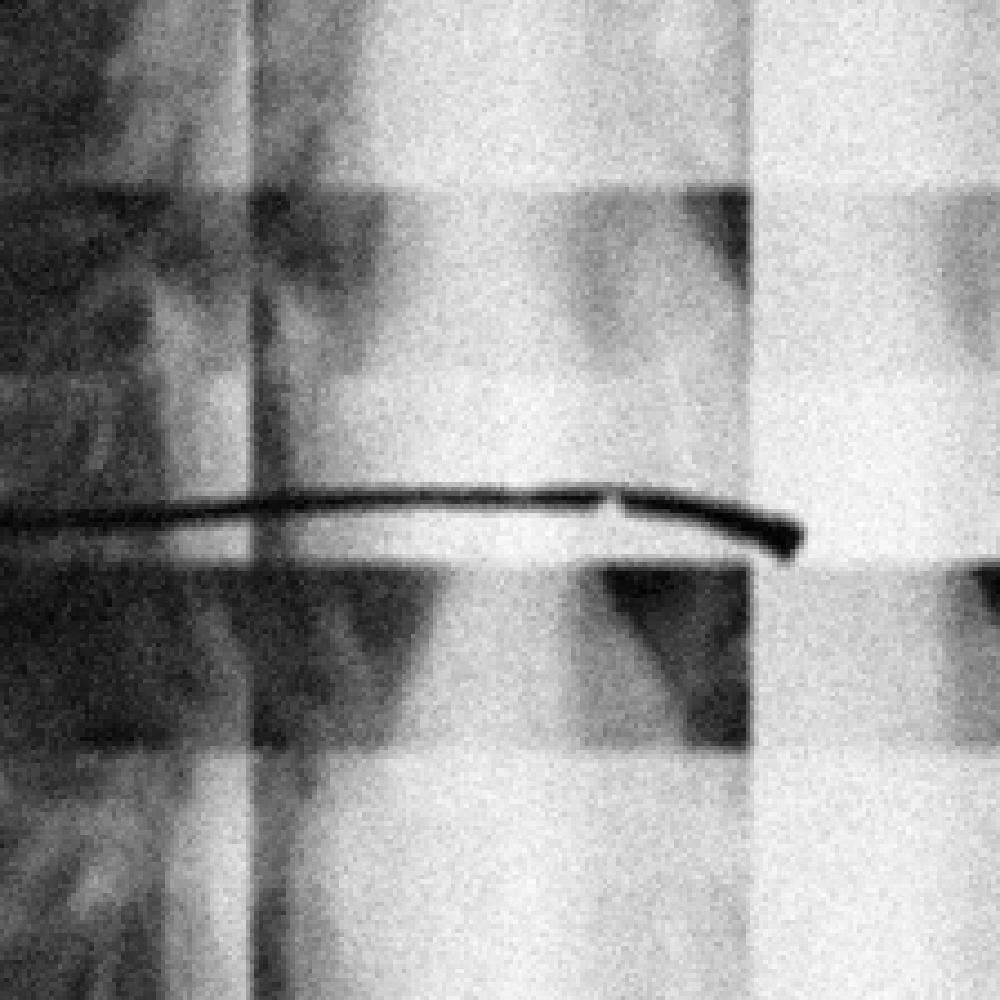} \\ 

\rotatebox{90}{\scriptsize segmentation} & 
\includegraphics[width=0.15\textwidth]{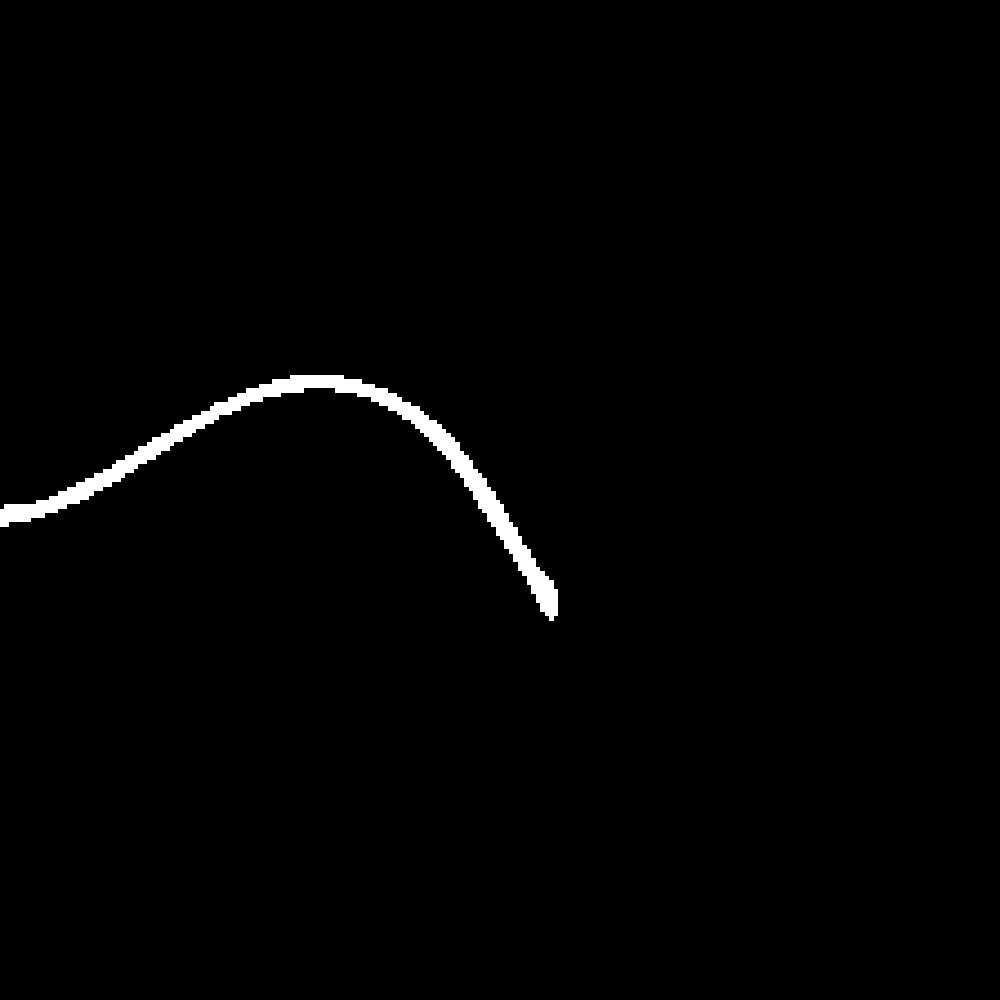} & \includegraphics[width=0.15\textwidth]{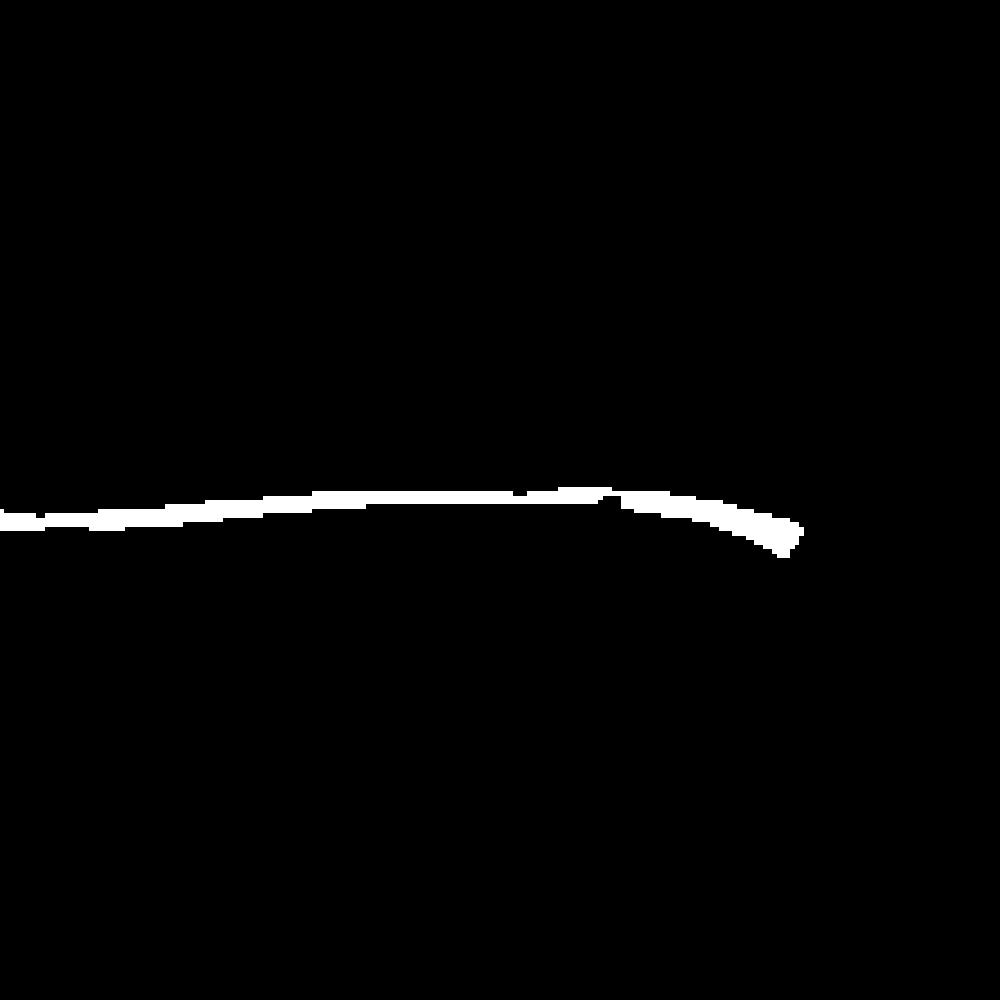} &
\includegraphics[width=0.15\textwidth]{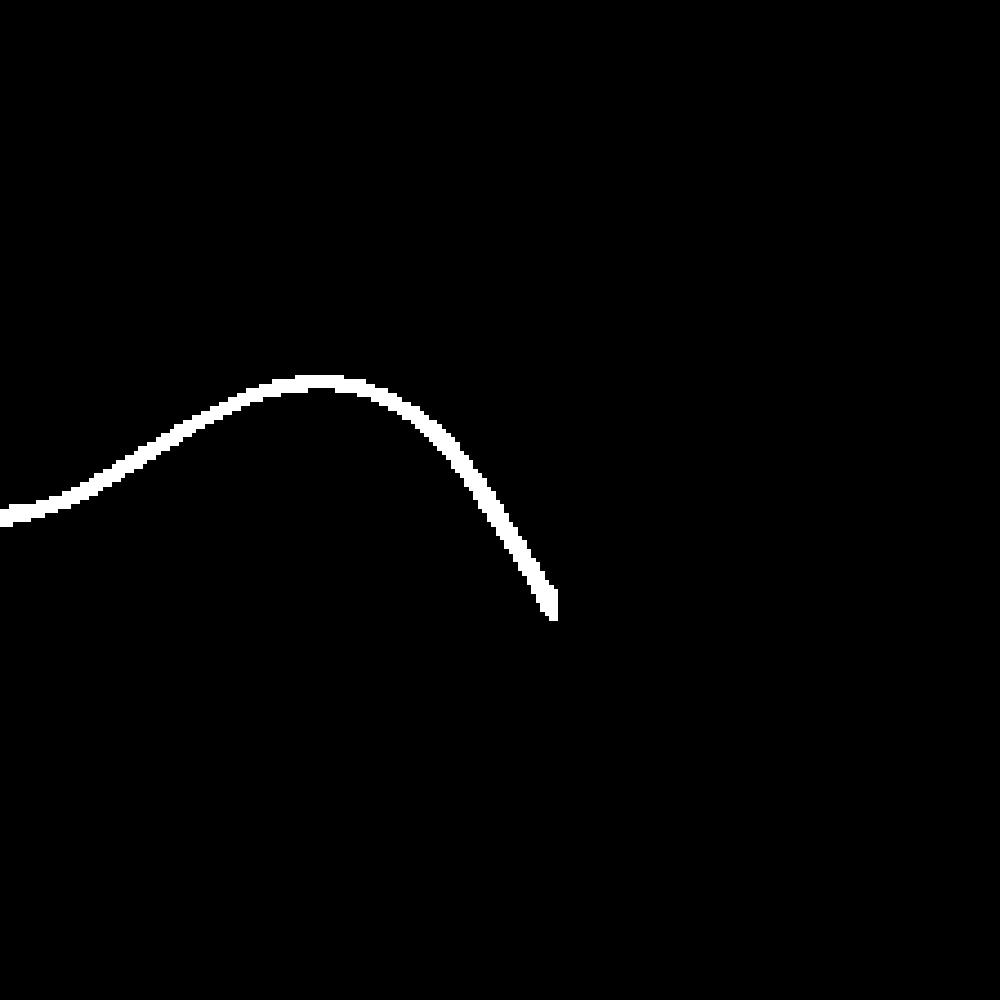} & \includegraphics[width=0.15\textwidth]{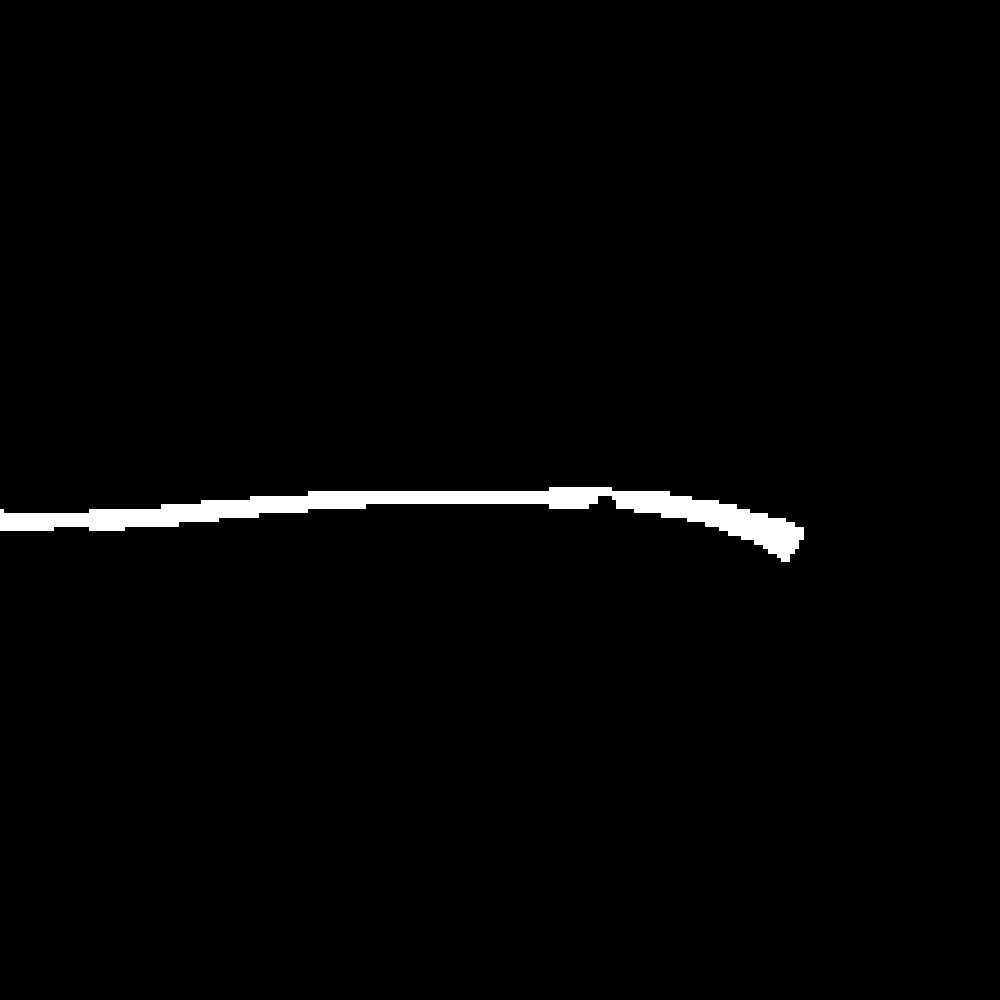} &
\includegraphics[width=0.15\textwidth]{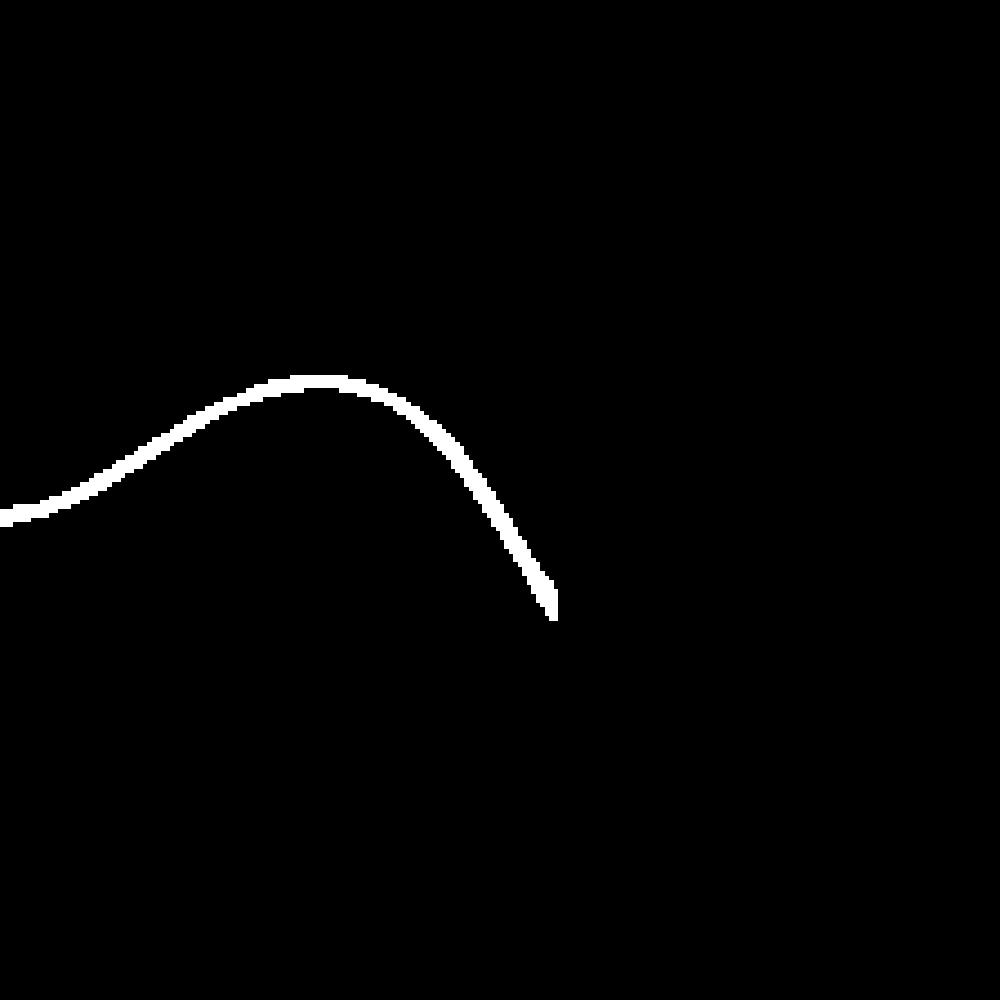} & \includegraphics[width=0.15\textwidth]{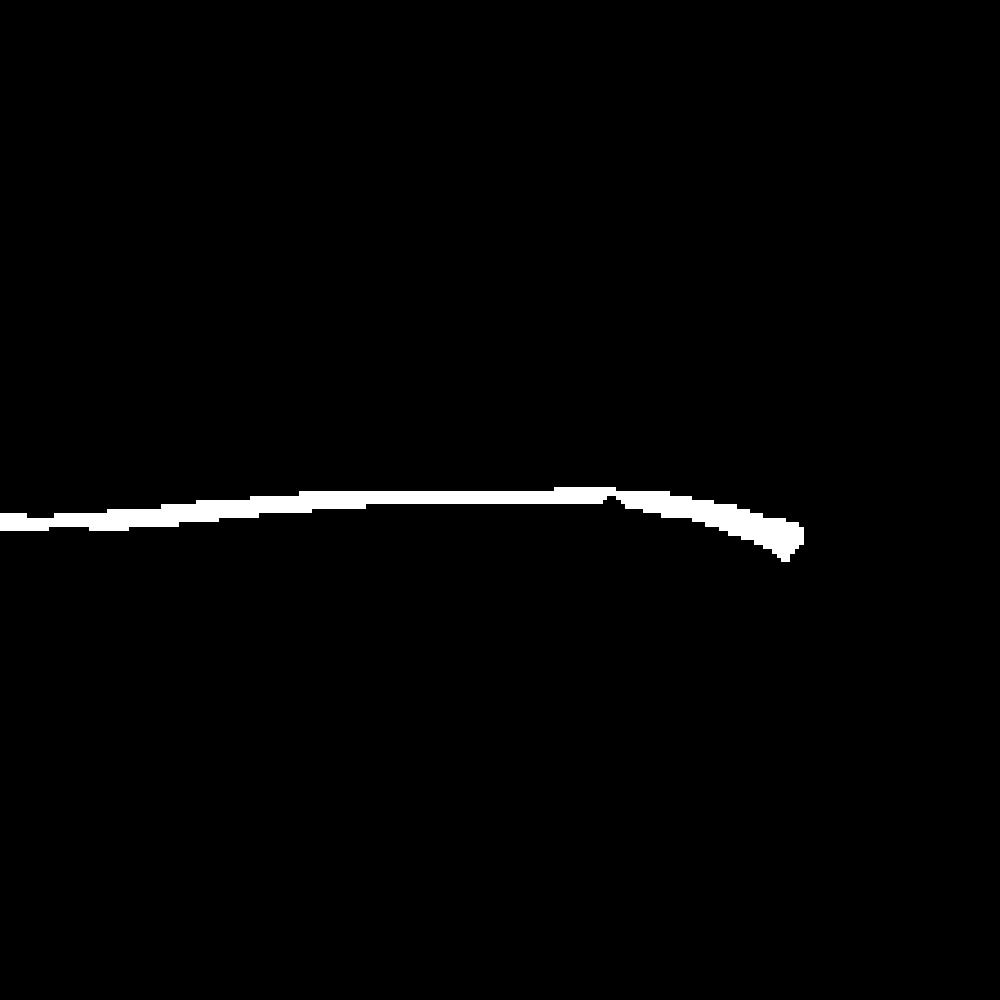} \\ 

\rotatebox{90}{\scriptsize Stripe} & 
\includegraphics[width=0.15\textwidth]{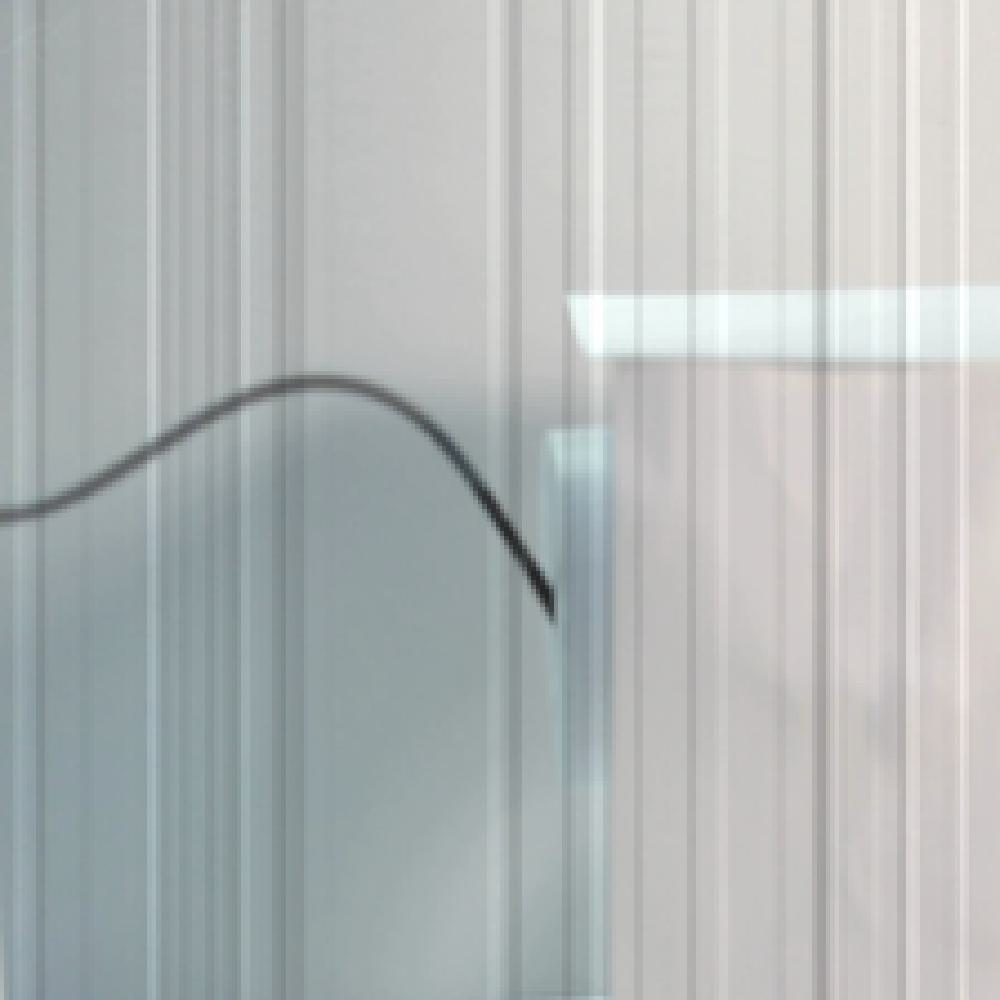} & \includegraphics[width=0.15\textwidth]{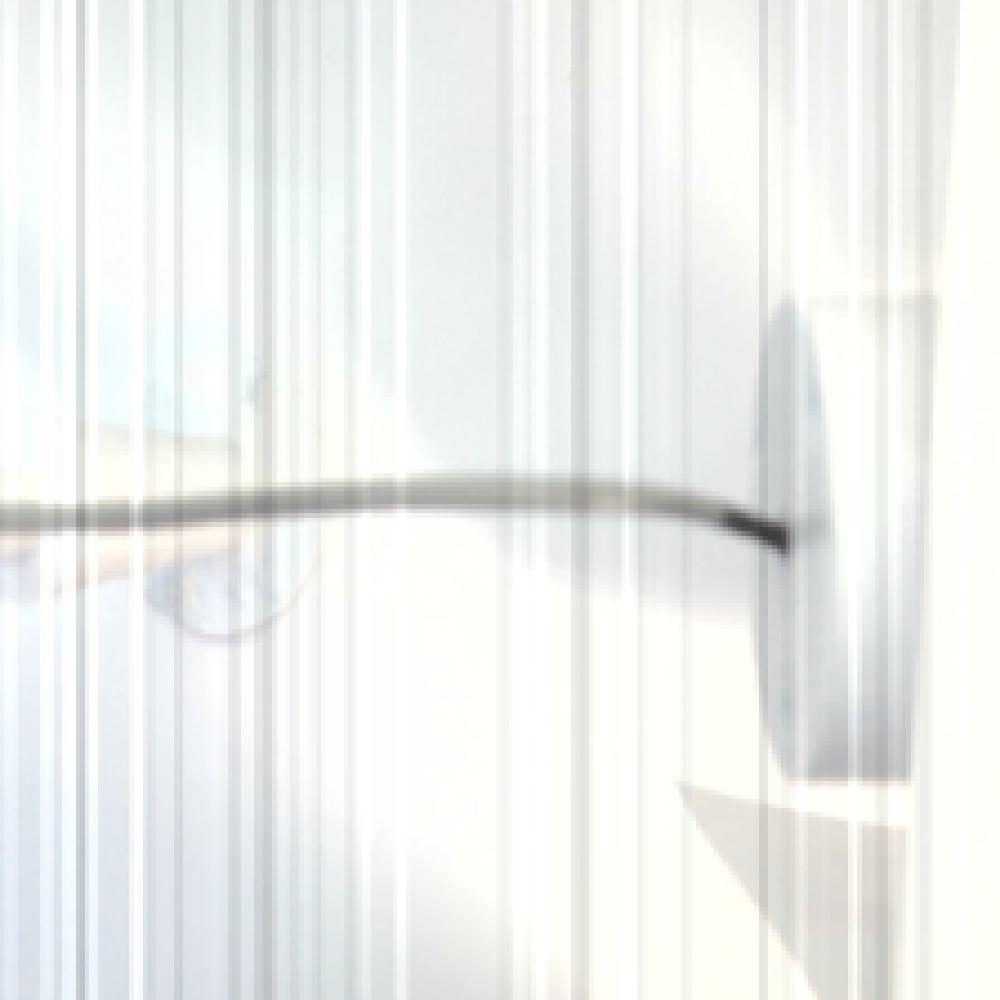} &
\includegraphics[width=0.15\textwidth]{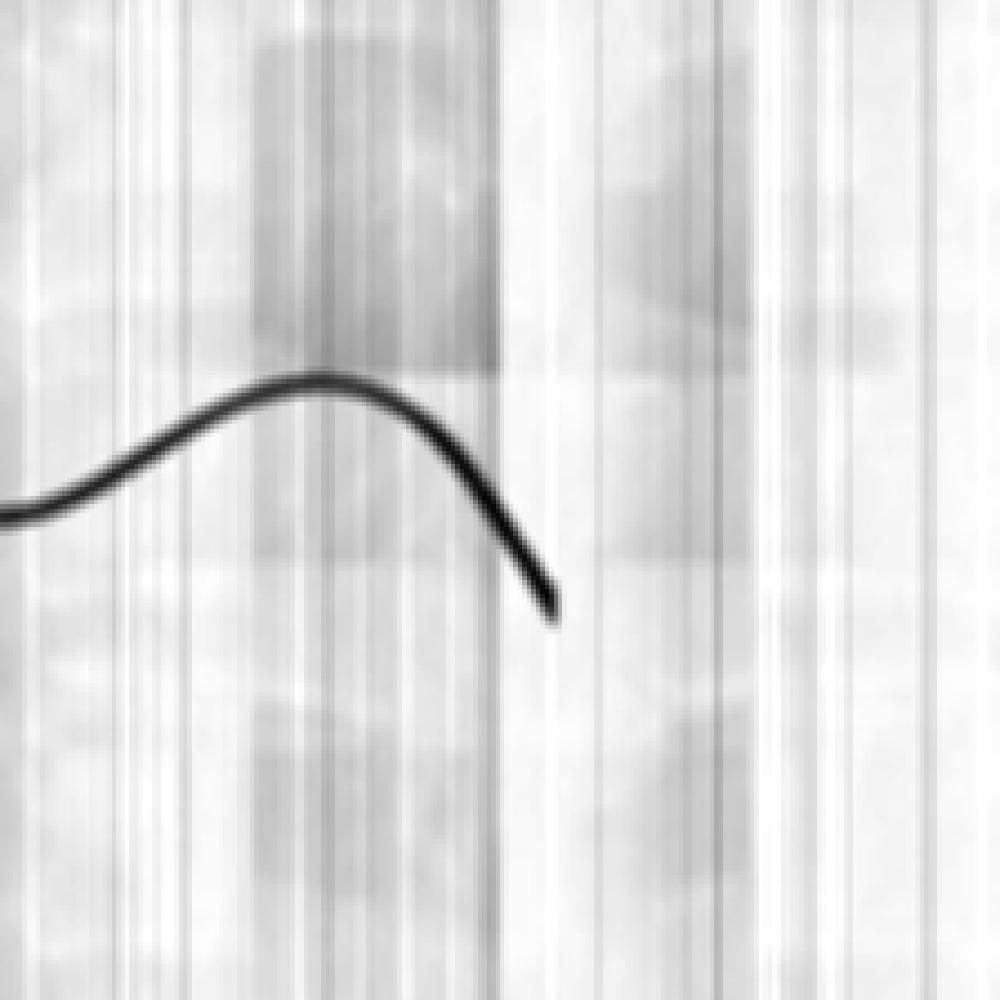} & \includegraphics[width=0.15\textwidth]{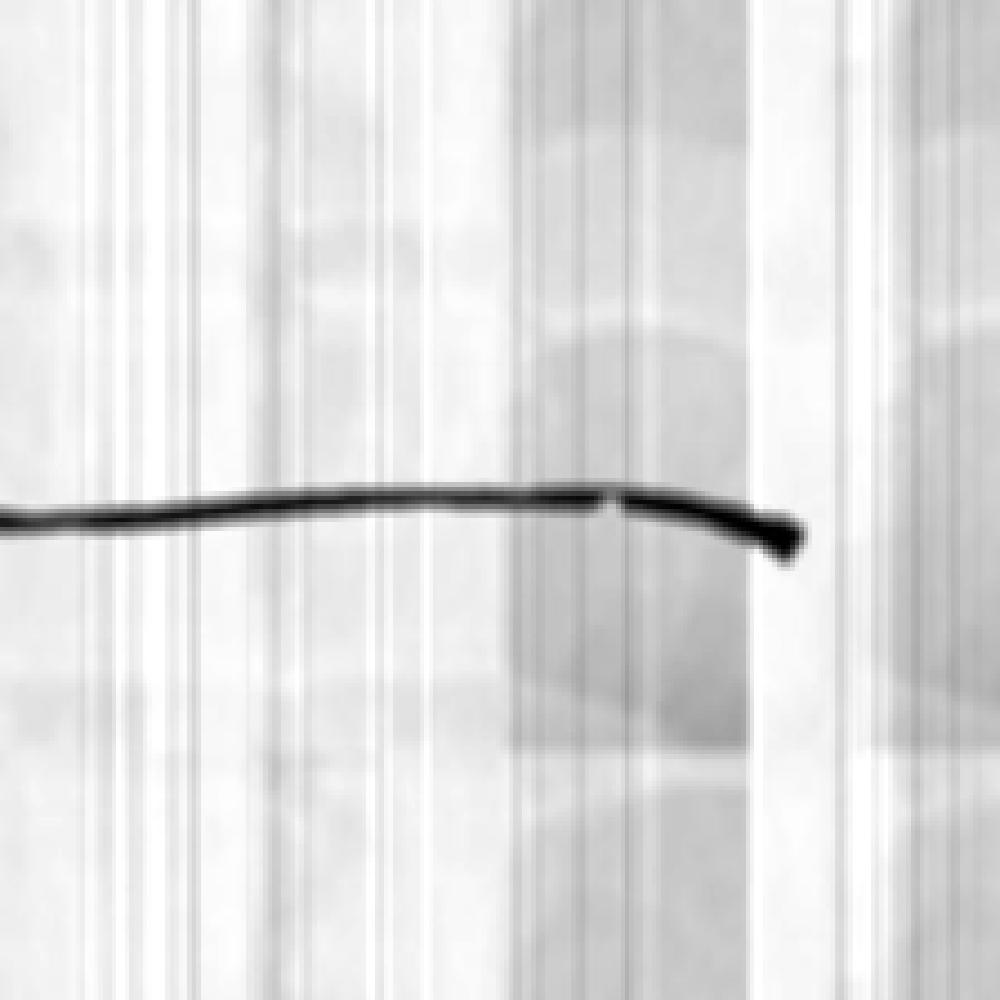} &
\includegraphics[width=0.15\textwidth]{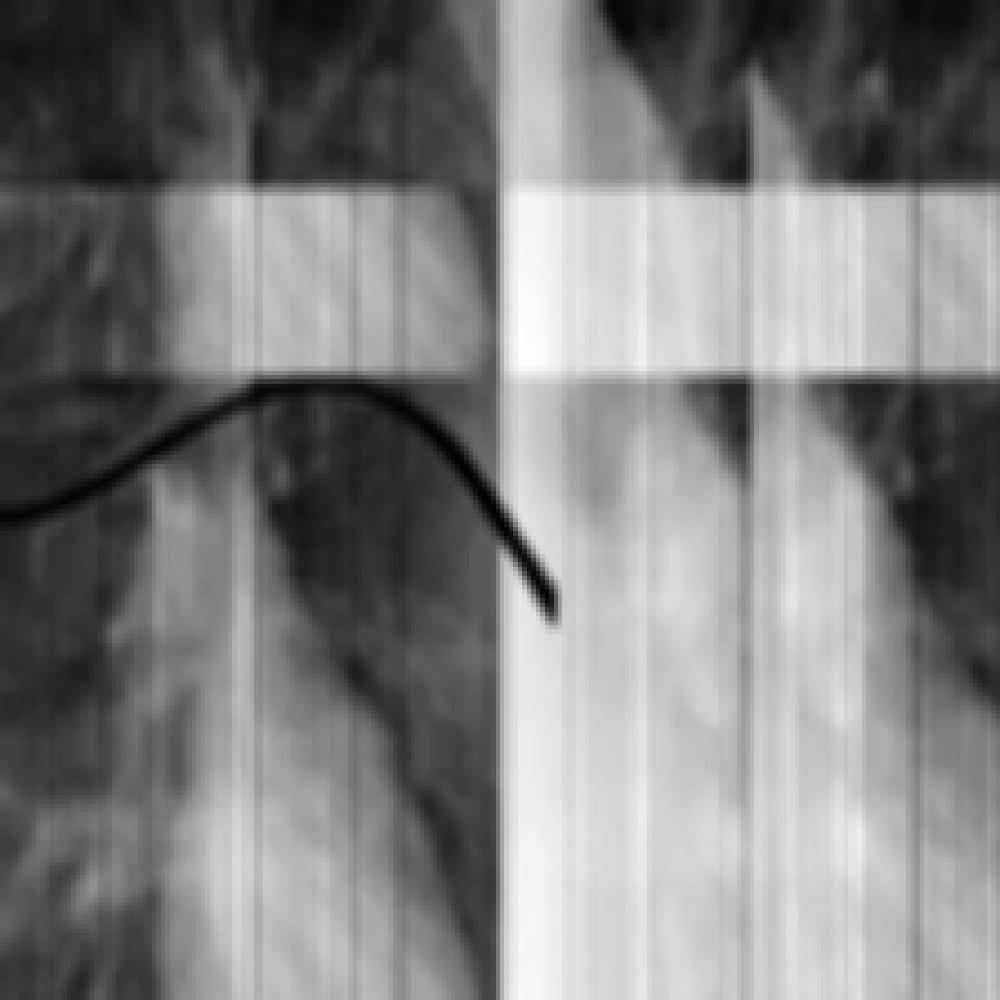} & \includegraphics[width=0.15\textwidth]{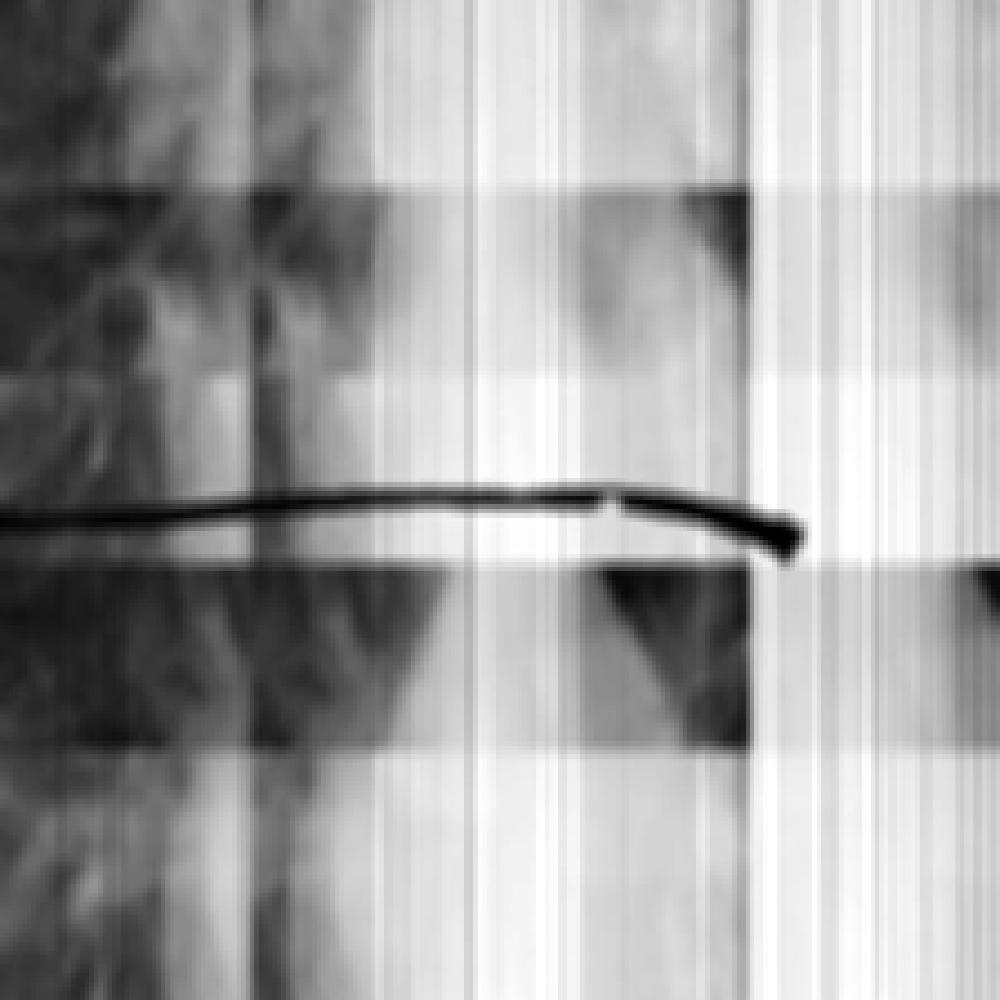} \\ 

\rotatebox{90}{\scriptsize segmentation} & 
\includegraphics[width=0.15\textwidth]{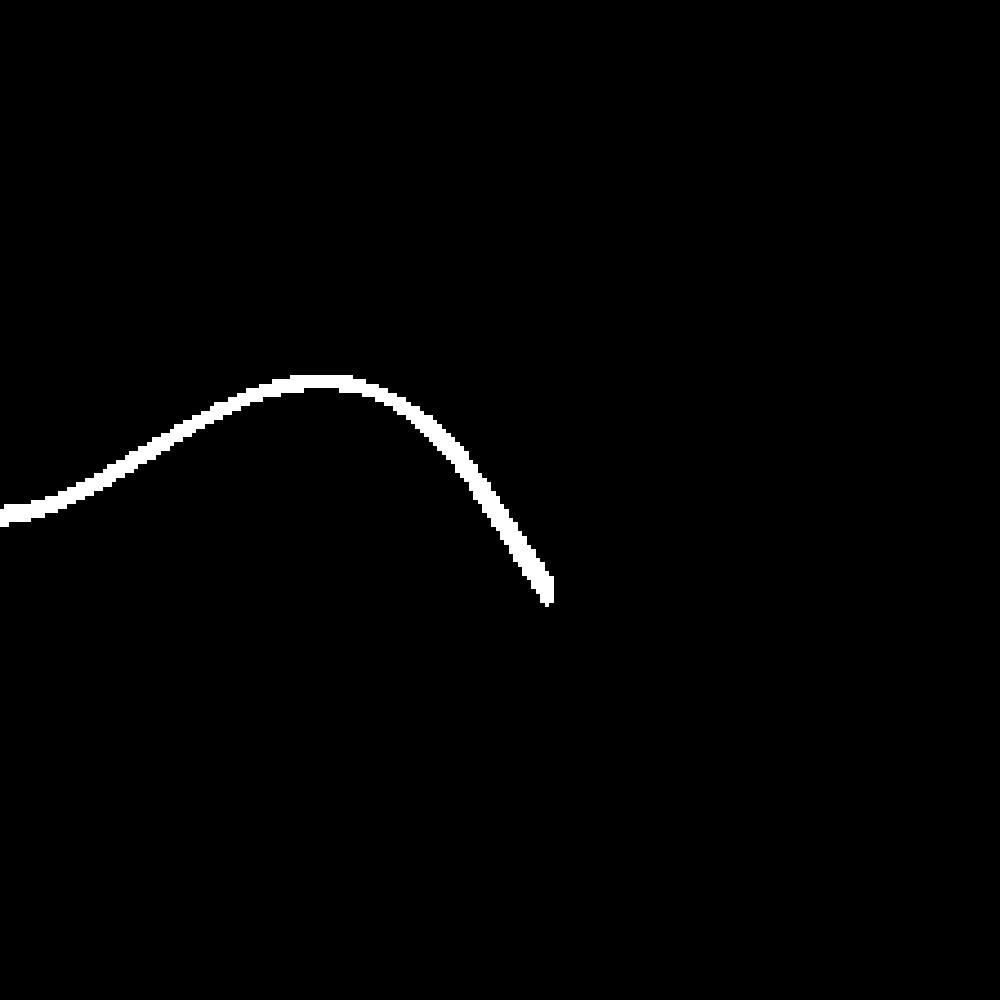} & \includegraphics[width=0.15\textwidth]{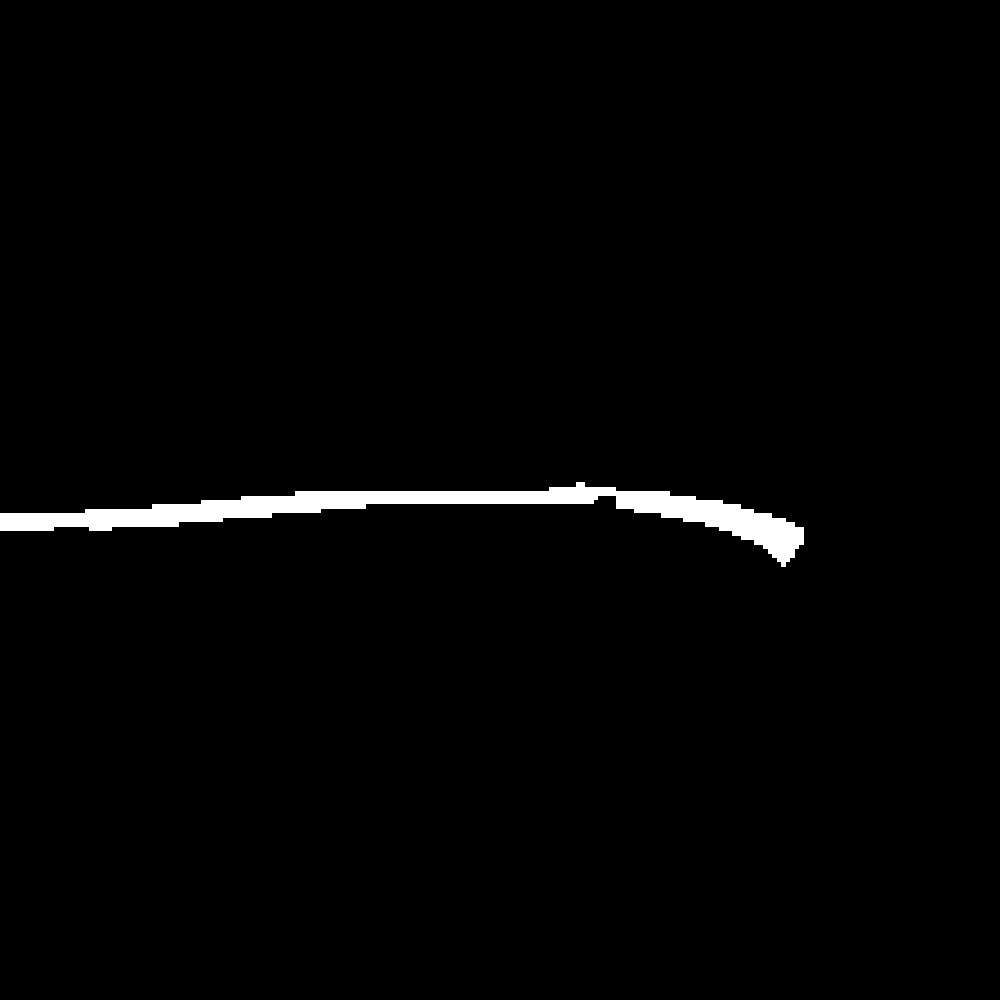} &
\includegraphics[width=0.15\textwidth]{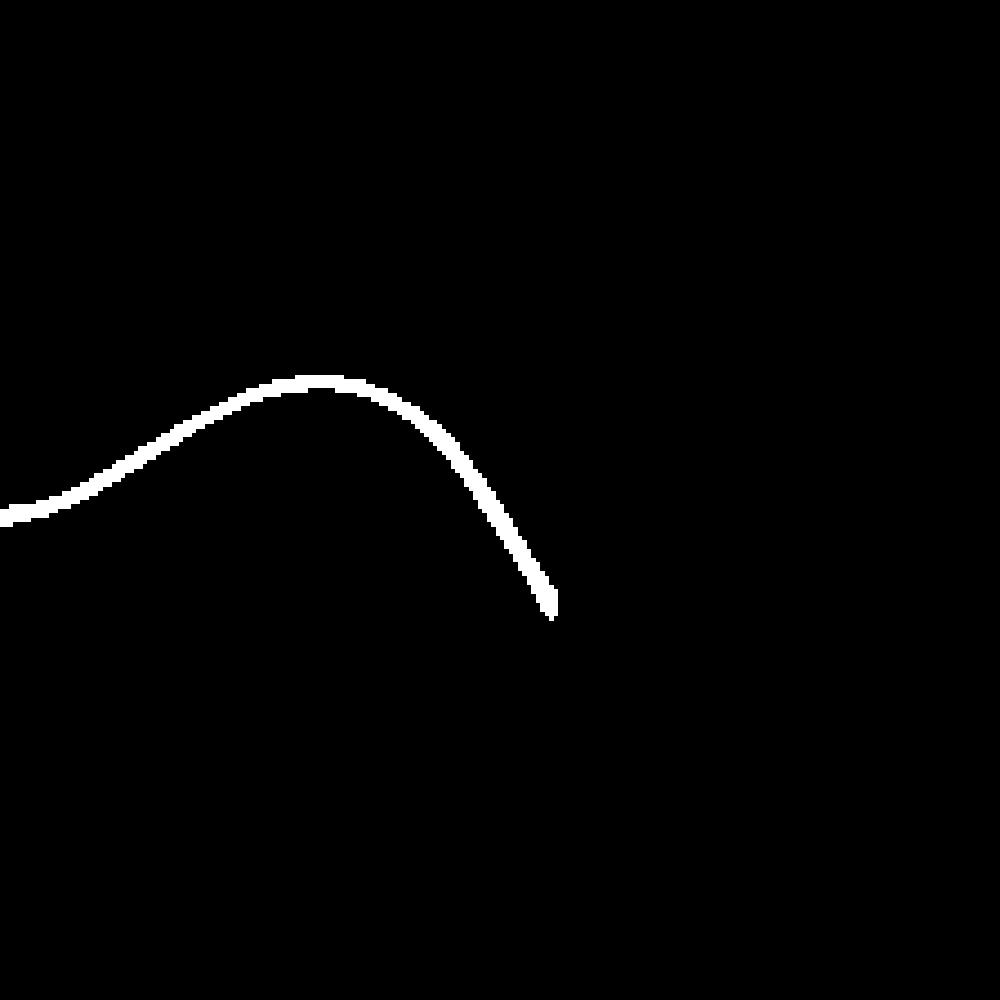} & \includegraphics[width=0.15\textwidth]{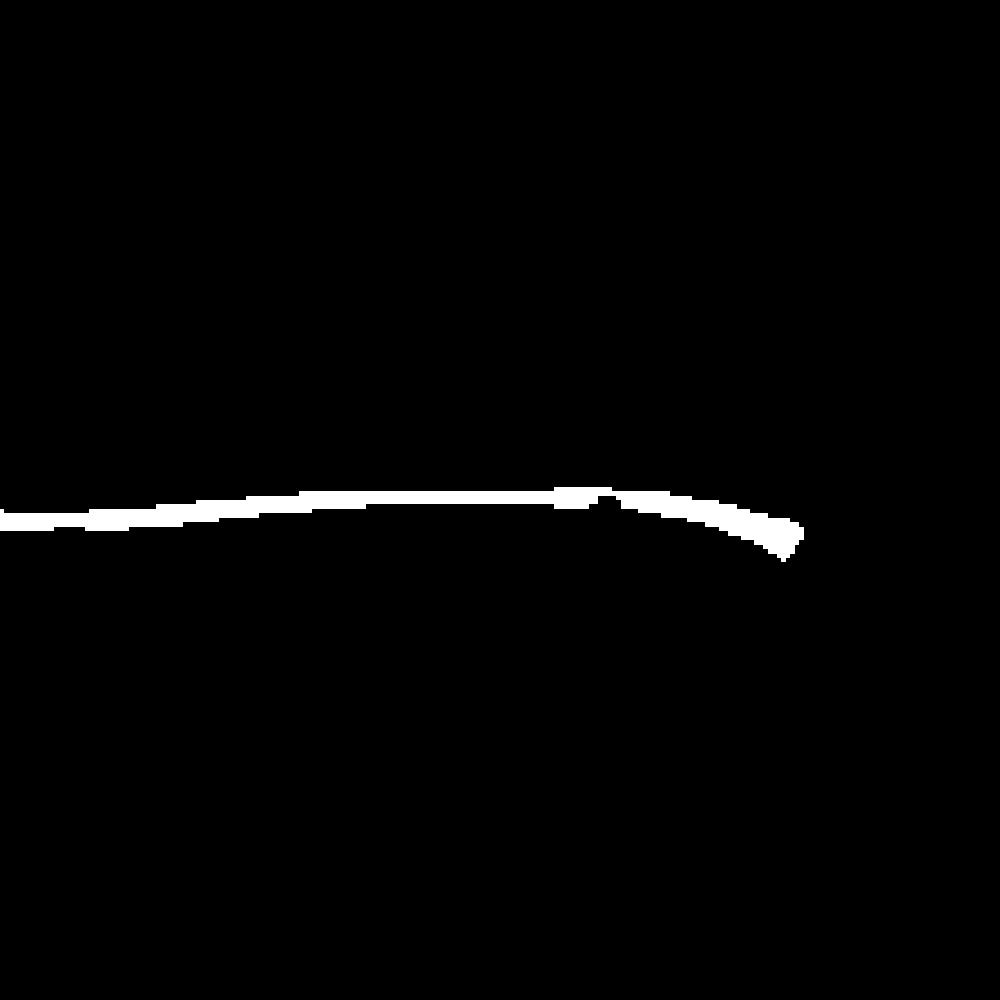} &
\includegraphics[width=0.15\textwidth]{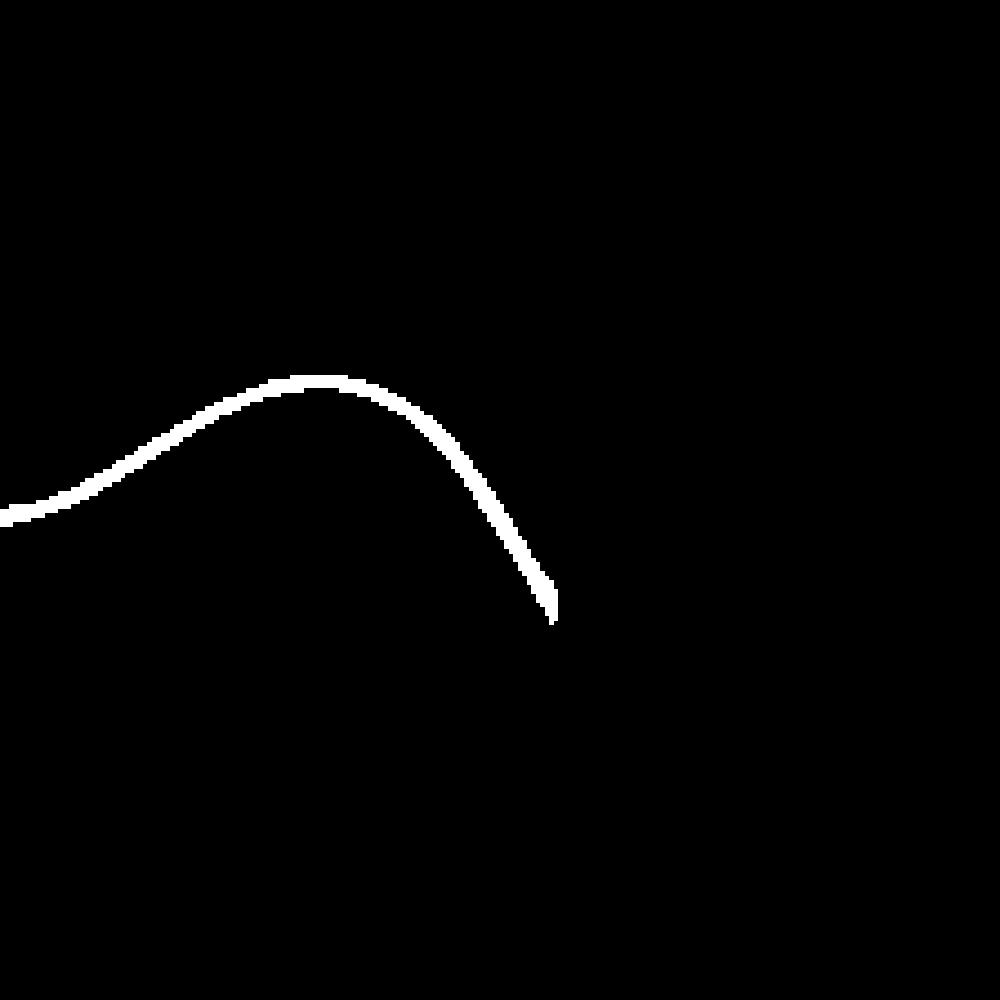} & \includegraphics[width=0.15\textwidth]{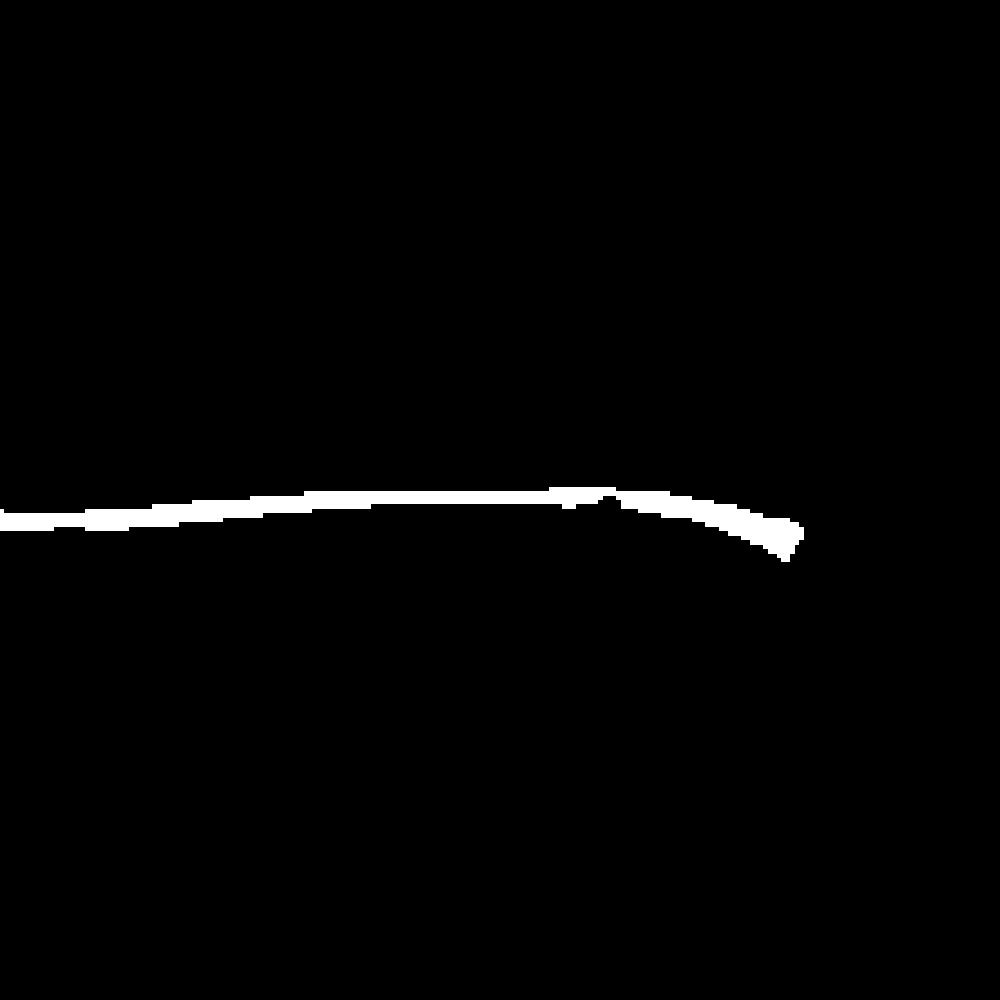} \\ 

\end{tabular}

\caption{Qualitative results across 3 out of 6 noise conditions for 3 input samples. The first row displays the original (clean) samples, followed by the corresponding segmentation outputs produced by the model in the second row. }
\label{fig:12x9_grid}
\end{figure*}

In this work, we proposed TransForSeg, a novel multitask Vision Transformer-based architecture for simultaneous catheter segmentation and 3D force estimation using stereo X-ray images. By integrating shared transformer components with task-specific heads, TransForSeg enables end-to-end learning without additional preprocessing. Extensive evaluations on RGB and synthetic X-ray datasets demonstrated that TransForSeg achieves state-of-the-art performance in force estimation while remaining computationally efficient. Ablation with a single-task variant (TransForcer) confirmed that the segmentation task significantly enhances force prediction, particularly in complex domains. Moreover, TransForSeg showed strong robustness to domain shift and input perturbations, maintaining reliable performance under various noise conditions. These findings highlight TransForSeg as a lightweight and generalizable solution for catheter-based interventions. Future work will explore its adaptation to real-world clinical settings and integration with autonomous robotic systems.

\vspace{-10pt}

\vfill


\end{document}